\RequirePackage{fix-cm}
\def\cleanbuildenabled{1}
\documentclass[10pt,letterpaper]{article}

\usepackage[margin=1in]{geometry}
\usepackage[utf8]{inputenc}
\usepackage[T1]{fontenc}
\usepackage{amsmath,amssymb,amsthm}
\usepackage{booktabs}
\usepackage{tabularx}
\usepackage[numbers,sort&compress]{natbib}
\usepackage{hyperref}

\makeatletter

\RequirePackage{iftex}
\ifPDFTeX\else
  \PackageError{llamapaper}{This style requires pdfLaTeX}
    {Build the paper with `make`.}
\fi

\RequirePackage[T1]{fontenc}
\RequirePackage{graphicx}
\RequirePackage{xcolor}
\RequirePackage{amsmath}
\RequirePackage{tcolorbox}
\tcbuselibrary{skins,breakable,listings}
\RequirePackage{microtype}
\RequirePackage{parskip}
\RequirePackage{titlesec}
\RequirePackage{caption}
\RequirePackage{enumitem}
\RequirePackage{fontawesome5}
\RequirePackage{hyperref}

\providecommand{\LIAssetPath}{assets}
\providecommand{\LILogoFile}{\LIAssetPath/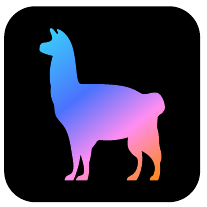}

\definecolor{LICyan}{HTML}{37D7FA}
\definecolor{LIBlue}{HTML}{4B72FE}
\definecolor{LIPink}{HTML}{FF8DF2}
\definecolor{LIOrange}{HTML}{FF8705}

\definecolor{LIInk}{HTML}{14182B}
\definecolor{LIMuted}{HTML}{5D6268}
\definecolor{LIAbstractTint}{HTML}{F1F3F5}
\colorlet{LILink}{LIBlue!78!black}
\colorlet{LIPaperTint}{LIBlue!7!white}

\pgfdeclarehorizontalshading{LIGradient}{100bp}{%
  color(0bp)=(LICyan); color(25bp)=(LICyan); color(45bp)=(LIBlue);
  color(59bp)=(LIPink); color(75bp)=(LIOrange); color(100bp)=(LIOrange)}

\AtBeginDocument{%
  \color{LIInk}%
  \global\let\default@color\current@color
}

\hypersetup{colorlinks=true, allcolors=LILink}

\renewcommand{\sfdefault}{cmss}

\titleformat*{\section}{\Large\sffamily\bfseries}
\titleformat*{\subsection}{\large\sffamily\bfseries}
\titleformat*{\subsubsection}{\normalsize\sffamily\bfseries}
\titleformat*{\paragraph}{\small\sffamily\bfseries}

\DeclareCaptionLabelSeparator{LISpace}{\enspace}
\setlist[itemize]{leftmargin=*, itemsep=0em, parsep=0.3em, topsep=0.3em}
\setlist[enumerate]{leftmargin=*, itemsep=0em, parsep=0.3em, topsep=0.3em}

\newcommand{\LIAffiliation}[1]{\gdef\li@affiliation{#1}}
\newcommand{\LIPaperType}[1]{\gdef\li@papertype{#1}}
\newcommand{\LIPaperDate}[1]{\gdef\li@paperdate{#1}}
\newcommand{\LIPDFAuthor}[1]{\gdef\li@pdfauthor{#1}}
\newcommand{\LIPDFSubject}[1]{\gdef\li@pdfsubject{#1}}
\newcommand{\LIAbstract}[1]{\long\gdef\li@abstract{#1}}
\newcommand{\LIPaperLinks}[1]{\gdef\li@paperlinks{#1}}
\newcommand{\LIHero}[1]{\long\gdef\li@hero{#1}}

\LIAffiliation{}
\LIPaperType{}
\LIPaperDate{}
\LIPDFAuthor{LlamaIndex Research}
\LIPDFSubject{LlamaIndex technical report}
\LIAbstract{}
\LIPaperLinks{}
\LIHero{}

\newcommand{\LILink}[4]{%
  \mbox{#1\hspace{0.4em}{\sffamily\bfseries #2}\hspace{0.6em}\href{#3}{\texttt{#4}}}}

\AtBeginDocument{%
  \hypersetup{
    pdftitle={\@title},
    pdfauthor={\li@pdfauthor},
    pdfsubject={\li@pdfsubject}
  }%
}

\definecolor{LIPanelGray}{HTML}{F5F6F8}
\definecolor{LIPanelBar}{HTML}{EBEDF0}
\definecolor{LIPanelLine}{HTML}{DCDFE4}

\newtcbox{\li@chip}{on line, enhanced, arc=1.5pt, boxsep=0pt, boxrule=0.4pt,
  left=1.6pt, right=1.6pt, top=1.2pt, bottom=1.2pt,
  colback=white, colframe=LIPanelLine, coltext=LIBlue!62!black,
  fontupper=\ttfamily\footnotesize}
\newcommand{\LIcode}[1]{\smash[t]{\li@chip{\detokenize{#1}}}}
\newcommand{\LIhash}{\smash[t]{\li@chip{\#}}}
\DisableLigatures{encoding=*, family=tt*}

\newtcolorbox{LIPanel}[1]{
  enhanced,
  colback=LIPanelGray, colframe=LIPanelLine, boxrule=0.5pt, arc=3pt,
  colbacktitle=LIPanelBar, coltitle=LIMuted,
  title={\textcolor{LIBlue}{\rule{5pt}{5pt}}\hspace{0.6em}\MakeUppercase{#1}},
  fonttitle=\footnotesize\sffamily\bfseries,
  toptitle=4pt, bottomtitle=4pt, lefttitle=0.45cm,
  titlerule=0.5pt, colframe=LIPanelLine,
  coltext=LIInk, fontupper=\small,
  boxsep=0pt, left=0.45cm, right=0.45cm, top=0.35cm, bottom=0.3cm,
  before skip=0.6em, after skip=0.6em
}

\newtcolorbox{LINote}{
  enhanced, frame hidden, sharp corners=west, arc=3pt,
  colback=LIPaperTint, coltext=LIInk, fontupper=\small\sffamily,
  borderline west={2.5pt}{0pt}{LIBlue},
  boxrule=0pt, boxsep=0pt, left=0.45cm, right=0.45cm, top=0.3cm, bottom=0.3cm,
  before skip=0.8em, after skip=0.8em
}

\lstdefinestyle{LIPrompt}{
  basicstyle=\ttfamily\scriptsize, columns=fullflexible, keepspaces=true,
  breaklines=true, breakatwhitespace=false, postbreak={\mbox{\hspace{1em}}},
  showstringspaces=false, upquote=true,
  literate={—}{{\textemdash}}1 {–}{{\textendash}}1}
\newtcbinputlisting{\LIPromptFile}[2]{
  listing file={#2}, listing only, listing options={style=LIPrompt},
  enhanced jigsaw, breakable,
  colback=LIPanelGray, colframe=LIPanelLine, boxrule=0.5pt, arc=3pt,
  colbacktitle=LIPanelBar, coltitle=LIMuted,
  title={\textcolor{LIBlue}{\rule{5pt}{5pt}}\hspace{0.6em}\MakeUppercase{#1}},
  fonttitle=\footnotesize\sffamily\bfseries,
  toptitle=4pt, bottomtitle=4pt, lefttitle=0.45cm, titlerule=0.5pt,
  boxsep=0pt, left=0.45cm, right=0.45cm, top=0.3cm, bottom=0.25cm,
  before skip=0.6em, after skip=0.6em}

\RequirePackage{tikz}
\RequirePackage{afterpage}   
\usetikzlibrary{arrows.meta}
\definecolor{LIFlowGray}{HTML}{7C8794}
\tikzset{
  LIFlow/.style={
    >={Stealth[length=5pt,width=4pt]},
    every node/.style={font=\scriptsize\sffamily, align=center, text width=1.75cm,
      inner xsep=0.1cm, inner ysep=0.15cm, minimum height=1.3cm, rounded corners=5pt}},
  flow source/.style={fill=LIFlowGray, text=white},
  flow step/.style={fill=LIBlue, text=white},
  flow pool/.style={fill=LIBlue!16!white, text=LIInk},
  flow arrow/.style={->, LIInk, line width=0.7pt},
}

\newcommand{\LIRowLabel}[1]{\strut\smash[t]{\sffamily\bfseries #1}}
\newcommand{\LIRowRule}{\noalign{\vspace{5pt}{\color{LIPanelLine}\hrule height 0.5pt}\vspace{5pt}}}

\newtcolorbox{liAbstractCard}{
  enhanced,
  frame hidden,
  width=\textwidth,
  colback=LIAbstractTint,
  coltext=black,
  boxrule=0pt,
  arc=5pt,
  outer arc=5pt,
  boxsep=0pt,
  left=15pt,
  right=15pt,
  top=15pt,
  bottom=13pt,
  before skip=0pt,
  after skip=0pt
}

\newcommand{\li@metasep}{\enspace\textperiodcentered\enspace}
\newcommand{\li@metaadd}[1]{%
  \ifx\li@meta\@empty\def\li@meta{#1}\else\expandafter\def\expandafter\li@meta\expandafter{\li@meta\li@metasep#1}\fi}

\newcommand{\li@maketitle}{%
  \thispagestyle{plain}%
  \begingroup
  \setlength{\parindent}{0pt}%
  \setlength{\parskip}{0pt}%
  \begin{liAbstractCard}
    {\fontsize{22}{24}\selectfont\sffamily\bfseries\raggedright\@title\par}
    \vspace{0.65em}
    {\small\raggedright\@author\par}
    \vspace{1.05em}
    {\small\bfseries Abstract\par}
    \vspace{0.3em}
    \begingroup
      \fontsize{9.25}{11.6}\selectfont
      \noindent\li@abstract\par
    \endgroup
    \vspace{0.8em}
    \def\li@meta{}%
    \ifx\li@affiliation\@empty\else\li@metaadd{\li@affiliation}\fi
    \ifx\li@papertype\@empty\else\li@metaadd{\li@papertype}\fi
    \ifx\li@paperdate\@empty\else\li@metaadd{\li@paperdate}\fi
    \noindent
    \begin{minipage}[b]{0.90\linewidth}
      \ifx\li@meta\@empty\else
        {\footnotesize\color{LIMuted}\li@meta\par}
      \fi
      \ifx\li@paperlinks\@empty\else
        \vspace{0.2em}
        {\footnotesize\li@paperlinks\par}
      \fi
    \end{minipage}%
    \begin{minipage}[b]{0.10\linewidth}
      \raggedleft
      \includegraphics[height=0.28in]{llamaindex-logo.pdf}
    \end{minipage}
  \end{liAbstractCard}
  \ifx\li@hero\@empty\else
    \vspace{0.65em}
    \captionsetup{hypcap=false}%
    \li@hero
  \fi
  \vspace{0.35em}
  \endgroup
}

\renewcommand{\maketitle}{%
  \if@twocolumn
    \twocolumn[{\li@maketitle}]%
  \else
    \li@maketitle
  \fi
}

\makeatother

\usepackage[nameinlink,capitalize]{cleveref}

\usepackage{subcaption}
\usetikzlibrary{positioning,calc,fit,shapes.misc,shapes.geometric,decorations.pathreplacing,backgrounds,patterns,matrix}
\usepackage{pgfplots}
\usepackage{pgfplotstable}
\pgfplotsset{compat=1.18}
\usepgfplotslibrary{fillbetween,groupplots}
\usepackage{colortbl}

\usepackage[normalem]{ulem}
\usepackage{multirow}
\usepackage{makecell}
\usepackage{longtable}
\usepackage{siunitx}
\usepackage{etoc}
\usepackage{pdflscape}
\usepackage{pifont}
\usepackage[section]{placeins}   %

\usepackage{algorithm}
\usepackage{algorithmic}

\definecolor{todocolor}{HTML}{D7263D}
\colorlet{prelimcolor}{LIOrange!85!black}
\colorlet{tbdcolor}{LIMuted}

\newcommand{\todo}[1]{{\color{todocolor}\textbf{[TODO:} #1\textbf{]}}}

\newcommand{\tbd}{{\color{tbdcolor}\textsc{tbd}}}

\newcommand{\prelim}[1]{{\color{prelimcolor}#1}\textsuperscript{\color{prelimcolor}\,p}}

\newcommand{\incomplete}[1]{#1\textsuperscript{*}}
\newcommand{\incompletenote}{$^{*}$Score includes pages that failed for operational reasons (provider credit, rate limit or time limit) and were scored 0.}

\ifdefined\cleanbuildenabled
\newcommand{\pierre}[1]{}
\newcommand{\pierrebot}[1]{}
\newcommand{\pierresuggest}[1]{}
\newcommand{\simon}[1]{}
\newcommand{\boyang}[1]{}
\newcommand{\antonio}[1]{}
\newcommand{\jerry}[1]{}
\newcommand{\note}[1]{}
\newcommand{\nutch}[1]{}
\else
\newcommand{\pierre}[1]{{\color{violet!75!black}\textbf{[Pierre:} #1\textbf{]}}}
\newcommand{\pierrebot}[1]{{\color{violet!45!gray}\textbf{[Pierre bot:} #1\textbf{]}}}
\newcommand{\pierresuggest}[1]{{\color{violet!75!black}\textbf{[Pierre suggest:} #1\textbf{]}}}
\newcommand{\simon}[1]{{\color{blue}\textbf{[Simon:} #1\textbf{]}}}
\newcommand{\boyang}[1]{{\color{red!70!black}\textbf{[Boyang:} #1\textbf{]}}}
\newcommand{\antonio}[1]{{\color{green!50!black}\textbf{[Antonio:} #1\textbf{]}}}
\newcommand{\jerry}[1]{{\color{orange!80!black}\textbf{[Jerry:} #1\textbf{]}}}
\newcommand{\note}[1]{{\color{violet}\textbf{[Note:} #1\textbf{]}}}
\newcommand{\nutch}[1]{{\color{violet}\textbf{[Nutch:} #1\textbf{]}}}
\fi

\newcommand{\textnext}{Text}
\newcommand{\tablenext}{Tables}
\newcommand{\chartnext}{Charts}
\newcommand{\formatnext}{Formatting}
\newcommand{\formsnext}{Forms}
\newcommand{\eprv}{Visual grounding}
\newcommand{\wordmetric}{\textsc{WordGate}}
\newcommand{\pbone}{ParseBench-1}
\newcommand{\pbtwo}{ParseBench-2}
\newcommand{\ev}[1]{\textit{#1}}  %

\newcommand{\crit}{\textsf{\textbf{C}}}
\newcommand{\maj}{\textsf{\textbf{M}}}
\newcommand{\mnr}{\textsf{m}}
\newcommand{\harm}{\textsf{0}}

\newcommand{\cmark}{$\checkmark$}
\newcommand{\xmark}{--}
\newcommand{\pmark}{$\circ$}

\newcommand{\labok}{\textcolor{sevHarm}{\ding{51}}}
\newcommand{\labweak}{\textcolor{sevMaj}{$\sim$}}
\newcommand{\labmiss}{\textcolor{sevCrit}{\ding{55}}}
\newcommand{\labna}{\textcolor{LIMuted}{$\cdot$}}

\newcommand{\capabilitydim}[2]{\par\vspace{0.45em}\noindent\textbf{#1.} #2\par}

\newcommand{\eg}{e.g.,}
\newcommand{\ie}{i.e.,}
\newcommand{\etal}{et~al.}
\newcommand{\vs}{vs.}

\colorlet{pbNext}{LIBlue}                %
\colorlet{pbLegacy}{LIPink!70!black}     %
\colorlet{pbLegacyB}{LIMuted}            %
\colorlet{pbLegacyC}{LIInk}              %
\definecolor{sevCrit}{HTML}{E5484D}      %
\colorlet{sevMaj}{LIOrange}              %
\colorlet{sevMin}{LICyan!85!black}       %
\definecolor{sevHarm}{HTML}{12A594}      %
\definecolor{gridWeak}{HTML}{D4A72C}     %
\colorlet{pbHi}{LIOrange!28!white}       %
\colorlet{pbRef}{LIPanelBar}             %
\pgfplotsset{compat=1.18,
  pbaxis/.style={font=\scriptsize\sffamily, /pgf/number format/assume math mode=true,
    tick label style={font=\scriptsize\sffamily, text=LIMuted}, label style={font=\scriptsize\sffamily, text=LIInk},
    title style={font=\footnotesize\sffamily\bfseries, text=LIInk},
    legend style={font=\scriptsize\sffamily, draw=none, fill=none, text=LIInk}, legend cell align=left,
    axis line style={LIPanelLine, line width=0.4pt}, tick style={LIPanelLine, line width=0.4pt},
    major grid style={LIPanelLine, line width=0.3pt}},
}
\tikzset{every picture/.append style={execute at begin picture={\sffamily}}}
\tikzset{
  pbbox/.style={draw=LIPanelLine, rounded corners=3pt, inner sep=3pt, font=\footnotesize\sffamily, align=center, text=LIInk},
  pbarr/.style={-{Stealth[length=5pt,width=4pt]}, line width=0.7pt, LIInk},
}
\newcommand{\sevbadge}[2]{\tikz[baseline=(b.base)]\node[fill=#1, text=white, font=\tiny\sffamily\bfseries, inner sep=1.2pt, rounded corners=1pt](b){#2};}
\pgfplotsset{keep if/.style 2 args={x filter/.append code={\edef\pbtmpa{\thisrow{#1}}\edef\pbtmpb{#2}\ifx\pbtmpa\pbtmpb\else\def\pgfmathresult{inf}\fi}}}

\newenvironment{pbnarrow}[1][0.62]{\par\centering\begin{minipage}{#1\textwidth}\centering}{\end{minipage}\par}

\newcommand{\RCsetup}{\centering\footnotesize\setlength{\tabcolsep}{5pt}%
  \setlength{\aboverulesep}{0pt}\setlength{\belowrulesep}{0pt}}
\newcommand{\RClongsetup}{\footnotesize\setlength{\tabcolsep}{5pt}%
  \setlength{\aboverulesep}{0pt}\setlength{\belowrulesep}{0pt}}
\newcommand{\RChead}[1]{\rule[-1.1ex]{0pt}{3.4ex}{\sffamily\bfseries #1}}
\newcommand{\RCgroup}[2]{\noalign{\vspace{1pt}\hbox{\setlength{\fboxsep}{5pt}\colorbox{LIPanelBar}{%
  \makebox[\dimexpr\linewidth-2\fboxsep][l]{\textcolor{LIBlue}{\rule{4pt}{4pt}}\hspace{0.5em}%
  {\scriptsize\sffamily\bfseries\color{LIMuted}\MakeUppercase{#1}}\hfill
  {\scriptsize\color{LIMuted}#2}}}}\vspace{4pt}}}
\newcommand{\RCsep}{\noalign{\vspace{3pt}{\color{LIPanelLine}\hrule height 0.4pt}\vspace{3.5pt}}}
\newcommand{\RCband}[2]{\multicolumn{#1}{>{\columncolor{LIPanelBar}}l}{\rule[-0.6ex]{0pt}{2.6ex}\hspace{3pt}\textcolor{LIBlue}{\rule{4pt}{4pt}}\hspace{0.5em}{\tiny\sffamily\bfseries\color{LIMuted}\MakeUppercase{#2}}}}
\newcommand{\RCnext}{\\\RCsep}
\newcommand{\RClast}{\\\noalign{\vspace{3.5pt}}}
\newcommand{\RCstart}{\noalign{\vspace{3.5pt}}}

\def\UrlBreaks{\do\/\do\_\do\.\do\-\do\:\do\=}
\DeclareRobustCommand{\code}[1]{\nolinkurl{#1}}
\usepackage{xfp}
\newcommand{\onedec}[1]{\num[round-mode=places,round-precision=1,round-integer-to-decimal,detect-weight=true,detect-family=true]{#1}}

\newcommand{\vonevtwotauval}{0.80}
\newcommand{\vonevtworhoval}{0.91}
\newcommand{\vtwotoptenchangedval}{6}

\newcommand{\Npartial}{Seven}
\newcommand{\bestoverall}{90.1}
\newcommand{\bestoverallname}{LlamaParse (Agentic Plus)}
\newcommand{\bestvlm}{85.7}
\newcommand{\bestvlmname}{Claude Opus 5.5}
\newcommand{\nmethods}{19}
\newcommand{\nopen}{9}
\newcommand{\npartial}{7}
\newcommand{\nvlms}{16}
\newcommand{\optAgenticChartsBase}{77.4}
\newcommand{\optAgenticChartsCostBase}{0.013}
\newcommand{\optAgenticChartsCostOpt}{0.013}
\newcommand{\optAgenticChartsLatBase}{6.5}
\newcommand{\optAgenticChartsLatOpt}{21.5}
\newcommand{\optAgenticChartsOpt}{87.1}
\newcommand{\optAgenticChartsVtwoBase}{86.0}
\newcommand{\optAgenticChartsVtwoOpt}{87.7}
\newcommand{\optAgenticFormsBase}{80.9}
\newcommand{\optAgenticFormsCoreBase}{78.9}
\newcommand{\optAgenticFormsCoreOpt}{81.0}
\newcommand{\optAgenticFormsCostBase}{0.013}
\newcommand{\optAgenticFormsCostOpt}{0.013}
\newcommand{\optAgenticFormsLatBase}{6.5}
\newcommand{\optAgenticFormsLatOpt}{25.7}
\newcommand{\optAgenticFormsOpt}{82.7}
\newcommand{\optAgenticFormsVtwoBase}{86.0}
\newcommand{\optAgenticFormsVtwoOpt}{86.3}
\newcommand{\optAzureBlank}{94.0}
\newcommand{\optAzureCore}{73.7}
\newcommand{\optAzureFormsBase}{76.3}
\newcommand{\optAzureFormsCoreBase}{73.7}
\newcommand{\optAzureFormsCoreOpt}{72.9}
\newcommand{\optAzureFormsCostBase}{0.010}
\newcommand{\optAzureFormsCostOpt}{0.010}
\newcommand{\optAzureFormsLatBase}{4.6}
\newcommand{\optAzureFormsLatOpt}{5.7}
\newcommand{\optAzureFormsOpt}{75.4}
\newcommand{\optAzureFormsVtwoBase}{66.1}
\newcommand{\optAzureFormsVtwoOpt}{66.0}
\newcommand{\optCeChartsBase}{73.1}
\newcommand{\optCeChartsCostBase}{0.004}
\newcommand{\optCeChartsCostOpt}{0.004}
\newcommand{\optCeChartsLatBase}{3.6}
\newcommand{\optCeChartsLatOpt}{20.6}
\newcommand{\optCeChartsOpt}{87.2}
\newcommand{\optCeChartsVtwoBase}{84.6}
\newcommand{\optCeChartsVtwoOpt}{87.0}
\newcommand{\optCeFormsBase}{81.2}
\newcommand{\optCeFormsCoreBase}{79.2}
\newcommand{\optCeFormsCoreOpt}{80.7}
\newcommand{\optCeFormsCostBase}{0.004}
\newcommand{\optCeFormsCostOpt}{0.004}
\newcommand{\optCeFormsLatBase}{6.5}
\newcommand{\optCeFormsLatOpt}{25.7}
\newcommand{\optCeFormsOpt}{82.5}
\newcommand{\optCeFormsVtwoBase}{84.6}
\newcommand{\optCeFormsVtwoOpt}{84.8}
\newcommand{\optDatalabChartsBase}{67.8}
\newcommand{\optDatalabChartsCostBase}{0.010}
\newcommand{\optDatalabChartsCostOpt}{0.013}
\newcommand{\optDatalabChartsLatBase}{7.1}
\newcommand{\optDatalabChartsLatOpt}{13.7}
\newcommand{\optDatalabChartsOpt}{61.8}
\newcommand{\optDatalabChartsVtwoBase}{81.3}
\newcommand{\optDatalabChartsVtwoOpt}{80.3}
\newcommand{\optPlusChartsBase}{\tbd}
\newcommand{\optPlusChartsCoreBase}{\tbd}
\newcommand{\optPlusChartsCoreOpt}{\tbd}
\newcommand{\optPlusChartsCostBase}{\tbd}
\newcommand{\optPlusChartsCostOpt}{\tbd}
\newcommand{\optPlusChartsLatBase}{\tbd}
\newcommand{\optPlusChartsLatOpt}{\tbd}
\newcommand{\optPlusChartsOpt}{\tbd}
\newcommand{\optPlusChartsVtwoBase}{\tbd}
\newcommand{\optPlusChartsVtwoOpt}{\tbd}
\newcommand{\optPlusFormsBase}{81.8}
\newcommand{\optPlusFormsCoreBase}{79.8}
\newcommand{\optPlusFormsCoreOpt}{81.2}
\newcommand{\optPlusFormsCostBase}{0.028}
\newcommand{\optPlusFormsCostOpt}{0.028}
\newcommand{\optPlusFormsLatBase}{25.7}
\newcommand{\optPlusFormsLatOpt}{35.8}
\newcommand{\optPlusFormsOpt}{82.9}
\newcommand{\optPlusFormsVtwoBase}{90.1}
\newcommand{\optPlusFormsVtwoOpt}{90.2}
\newcommand{\optReductoChartsBase}{25.2}
\newcommand{\optReductoChartsCostBase}{0.010}
\newcommand{\optReductoChartsCostOpt}{0.064}
\newcommand{\optReductoChartsLatBase}{3.0}
\newcommand{\optReductoChartsLatOpt}{238.0}
\newcommand{\optReductoChartsOpt}{\incomplete{73.6}}
\newcommand{\optReductoChartsVtwoBase}{76.8}
\newcommand{\optReductoChartsVtwoOpt}{\incomplete{84.8}}
\newcommand{\optTextractFormsBase}{59.3}
\newcommand{\optTextractFormsCoreBase}{67.8}
\newcommand{\optTextractFormsCoreOpt}{71.6}
\newcommand{\optTextractFormsCostBase}{0.015}
\newcommand{\optTextractFormsCostOpt}{0.065}
\newcommand{\optTextractFormsLatBase}{4.6}
\newcommand{\optTextractFormsLatOpt}{5.2}
\newcommand{\optTextractFormsOpt}{73.9}
\newcommand{\optTextractFormsVtwoBase}{--}
\newcommand{\optTextractFormsVtwoOpt}{64.2}
\newcommand{\resAdjParser}{18}
\newcommand{\resAdjVlm}{15}
\newcommand{\resAdjZeroParser}{7}
\newcommand{\resAdjZeroParserRanks}{3--4, 5--6, 7--8, 11--12, 13--14, 14--15 and 18--19}
\newcommand{\resAdjZeroVlm}{9}
\newcommand{\resAllMaxCharts}{92.6}
\newcommand{\resAllMaxLayout}{88.6}
\newcommand{\resAllMinCharts}{1.2}
\newcommand{\resAllMinLayout}{43.8}
\newcommand{\resCIWidthChartsParser}{19.1}
\newcommand{\resCIWidthChartsVlm}{22.9}
\newcommand{\resChartsLowCostMax}{15.2}
\newcommand{\resChartsLowCostMin}{15.0}
\newcommand{\resChartsLowCount}{Five}
\newcommand{\resChartsLowNames}{Mistral OCR 4.1, Reducto (r-1-mini), Azure Document Intelligence (layout), Reducto (non-agentic) and Extend (Parse)}
\newcommand{\resCiMedianWidth}{2.2}
\newcommand{\resCiMedianWidthPage}{1.5}
\newcommand{\resCiWidthMaxParser}{3.5}
\newcommand{\resCiWidthMaxVlm}{4.2}
\newcommand{\resCiWidthMinParser}{1.0}
\newcommand{\resCiWidthMinVlm}{2.0}
\newcommand{\resClusterPages}{4{,}320}
\newcommand{\resClustersAll}{2{,}646}
\newcommand{\resClustersCharts}{99}
\newcommand{\resClustersForms}{244}
\newcommand{\resClustersLayout}{365}
\newcommand{\resClustersMultiDataset}{50}
\newcommand{\resClustersTables}{140}
\newcommand{\resClustersText}{485}
\newcommand{\resCostBestParser}{0.0281}
\newcommand{\resCostBestVlm}{0.065}
\newcommand{\resCostBestVlmRatio}{2.3}
\newcommand{\resCostBestVlmScore}{85.7}
\newcommand{\resCostFrontier}{Reducto (r-1-mini) at \$0.0010 for 71.6, LlamaParse (Cost Effective) at \$0.0038 for 84.6, LlamaParse (Agentic) at \$0.0125 for 86.0 and LlamaParse (Agentic Plus) at \$0.0281 for 90.1}
\newcommand{\resCostFrontierN}{Four}
\newcommand{\resCostFrontierVlm}{the vision models on it are GPT-6 luna (low) at \$0.0013 for 73.5, Muse Ember 1.0 (low) at \$0.0016 for 74.5}
\newcommand{\resCostMax}{0.0300}
\newcommand{\resCostMaxName}{LandingAI and Reducto}
\newcommand{\resCostMin}{0.0010}
\newcommand{\resCostMinName}{Reducto (r-1-mini)}
\newcommand{\resCostRatio}{30}
\newcommand{\resDecorLead}{Reducto (r-1) and Reducto (r-1-mini)}
\newcommand{\resDecorLeadVerb}{led by}
\newcommand{\resDirHoldHigh}{46}
\newcommand{\resDirHoldLow}{33}
\newcommand{\resDirHoldRankHi}{15}
\newcommand{\resDirHoldRankLo}{11}
\newcommand{\resDirTauMedian}{0.95}
\newcommand{\resDirTauPfive}{0.86}
\newcommand{\resFlashLite}{62.7 and 61.4}
\newcommand{\resFmtCoreMax}{93.1}
\newcommand{\resFmtCoreMin}{85.0}
\newcommand{\resFormsBestName}{LlamaParse (Agentic Plus)}
\newcommand{\resFormsBlankMaxParser}{95.7}
\newcommand{\resFormsBlankMinParser}{91.6}
\newcommand{\resFormsCoreMaxParser}{79.8}
\newcommand{\resFormsCoreMaxVlm}{77.7}
\newcommand{\resFormsCoreMinParser}{70.9}
\newcommand{\resFormsCoreMinVlm}{68.8}
\newcommand{\resFormsMedEmpty}{81.3}
\newcommand{\resFormsMedFill}{75.1}
\newcommand{\resHeadingsFloor}{90}
\newcommand{\resLatMax}{22.2}
\newcommand{\resLatMaxName}{Pulse (Ultra 2)}
\newcommand{\resLatMin}{1.0}
\newcommand{\resLatMinName}{Mistral OCR 4.1}
\newcommand{\resLatNinetyMax}{51.0}
\newcommand{\resLatNinetyMin}{1.7}
\newcommand{\resLatRestMax}{11.8}
\newcommand{\resLatSecond}{15.9}
\newcommand{\resLatSecondName}{LlamaParse (Agentic Plus)}
\newcommand{\resLayoutBestName}{LlamaParse (Cost Effective)}
\newcommand{\resLayoutRange}{the ranked parsers range from 58.4 to 88.6}
\newcommand{\resLeadDiff}{4.0}
\newcommand{\resLeadDiffHi}{4.6}
\newcommand{\resLeadDiffLo}{3.4}
\newcommand{\resLeadGapOther}{6.2}
\newcommand{\resLeadGapSecond}{4.1}
\newcommand{\resMaxParserCharts}{92.6}
\newcommand{\resMaxParserFmt}{93.6}
\newcommand{\resMaxParserForms}{81.8}
\newcommand{\resMaxParserLayout}{88.6}
\newcommand{\resMaxParserTables}{94.6}
\newcommand{\resMaxParserText}{91.1}
\newcommand{\resMedParserCharts}{64.2}
\newcommand{\resMedParserFmt}{83.3}
\newcommand{\resMedParserForms}{77.7}
\newcommand{\resMedParserLayout}{83.8}
\newcommand{\resMedParserTables}{84.7}
\newcommand{\resMedParserText}{85.1}
\newcommand{\resMedVlmCharts}{39.6}
\newcommand{\resMedVlmFmt}{86.6}
\newcommand{\resMedVlmForms}{76.0}
\newcommand{\resMedVlmLayout}{64.4}
\newcommand{\resMedVlmTables}{89.0}
\newcommand{\resMedVlmText}{87.2}
\newcommand{\resMinParserCharts}{1.2}
\newcommand{\resMinParserFmt}{77.9}
\newcommand{\resMinParserForms}{73.5}
\newcommand{\resMinParserLayout}{58.4}
\newcommand{\resMinParserTables}{73.4}
\newcommand{\resMinParserText}{78.3}
\newcommand{\resNextThree}{Extend (Parse 2, 83.2), Pulse (Ultra 2, 82.7) and Extend (Parse Light, 81.6)}
\newcommand{\resNextThreeGap}{2.3}
\newcommand{\resNparsersAdj}{18}
\newcommand{\resNvlmWord}{16}
\newcommand{\resOpenBest}{77.1}
\newcommand{\resOpenBestName}{Infinity-Parser2 Pro}
\newcommand{\resOpenLow}{61.6}
\newcommand{\resOpenNameLow}{Qwen3.8 27B}
\newcommand{\resOpenParsersAbove}{ten}
\newcommand{\resOpenSecond}{77.0}
\newcommand{\resOpenSecondName}{LightOnOCR-3 4B}
\newcommand{\resOtherName}{Reducto}
\newcommand{\resOtherScore}{83.9}
\newcommand{\resParserLow}{66.1}
\newcommand{\resParserLowName}{Azure Document Intelligence (layout)}
\newcommand{\resParserMedian}{78.3}
\newcommand{\resParsersHeadingsMin}{90.8}
\newcommand{\resPartialCover}{Upstage Document Parse v2 covers 12, Firecrawl, olmOCR 2 7B (open weight) and Xiaomi-OCR-0 (open weight) 11, Anyformat (standard) 2 and Nutrient DWS (agentic) 1}
\newcommand{\resSecondName}{LlamaParse (Agentic)}
\newcommand{\resSecondScore}{86.0}
\newcommand{\resSensCoreSlotShiftMax}{2}
\newcommand{\resSensCoreSlotTauMin}{0.94}
\newcommand{\resSensNoChartsFrom}{12}
\newcommand{\resSensNoChartsTau}{0.52}
\newcommand{\resSensNoChartsTo}{3}
\newcommand{\resSensNoFormsTau}{0.96}
\newcommand{\resSensNoLayoutTau}{0.88}
\newcommand{\resSensNoTextTau}{0.94}
\newcommand{\resStrikeMax}{95.5}
\newcommand{\resStrikeMed}{67.9}
\newcommand{\resStrikeMin}{13.8}
\newcommand{\resTabComplexMax}{90.3}
\newcommand{\resTabComplexMed}{74.3}
\newcommand{\resTabComplexMin}{34.5}
\newcommand{\resTabCoreMax}{96.1}
\newcommand{\resTabCoreMed}{89.3}
\newcommand{\resTabCoreMin}{80.5}
\newcommand{\resTabDenseMax}{94.8}
\newcommand{\resTabDenseMed}{80.0}
\newcommand{\resTabDenseMin}{64.0}
\newcommand{\resTabFinMax}{93.7}
\newcommand{\resTabFinMed}{90.8}
\newcommand{\resTabFinMin}{81.9}
\newcommand{\resTabLead}{LlamaParse (Agentic Plus) leads every table dataset among the parsers}
\newcommand{\resTabLeadVlm}{Claude Opus 5.5 leads three of the five}
\newcommand{\resTabSerffMax}{93.5}
\newcommand{\resTabSerffMed}{84.8}
\newcommand{\resTabSerffMin}{77.7}
\newcommand{\resTightSpanHi}{94.6}
\newcommand{\resTightSpanLo}{72.2}
\newcommand{\resTightSpanMax}{22}
\newcommand{\resTightSpanName}{Tables}
\newcommand{\resTopTenLow}{78.3}
\newcommand{\resTopTenSpan}{11.8}
\newcommand{\resUnderFloor}{Five parsers score 20.0 on underline}
\newcommand{\resUnderMax}{96.8}
\newcommand{\resUnderMed}{49.5}
\newcommand{\resUnderMin}{20.0}
\newcommand{\resVlmAbove}{every parser configuration except LlamaParse (Agentic Plus), LlamaParse (Agentic)}
\newcommand{\resVlmAboveMedText}{eleven}
\newcommand{\resVlmSecondName}{Gemini 4 argon (low)}
\newcommand{\resVlmSecondScore}{84.2}
\newcommand{\resVoneCoreTau}{0.79}
\newcommand{\resVoneMoveChartsFrom}{19th}
\newcommand{\resVoneMoveChartsName}{Databricks AI Parse}
\newcommand{\resVoneMoveChartsTo}{10th}
\newcommand{\resVoneMoveLayoutFrom}{6th}
\newcommand{\resVoneMoveLayoutName}{Azure Document Intelligence (layout)}
\newcommand{\resVoneMoveLayoutTo}{11th}
\newcommand{\resVoneTauCharts}{0.80}
\newcommand{\resVoneTauFormatting}{0.82}
\newcommand{\resVoneTauLayout}{0.77}
\newcommand{\resVoneTauTables}{0.68}
\newcommand{\resVoneTauText}{0.09}

\newcommand{\ablTextPonlyMed}{6.9}
\newcommand{\ablTextPonlyTau}{0.86}
\newcommand{\ablChartsFonlyFrom}{12}
\newcommand{\ablChartsFonlyTo}{2}
\newcommand{\ablNoGroupTextMed}{0.9}
\newcommand{\ablNoGroupFmtMed}{0.4}
\newcommand{\ablNoGroupFormsMed}{0.2}
\newcommand{\ablNoGroupMaxMove}{0}
\newcommand{\ablHeadTauMin}{0.98}

\newcommand{\mconstTauMin}{0.98}
\newcommand{\mconstFlipMax}{1}
\newcommand{\mconstMoveMax}{1}
\newcommand{\mconstDeltaMax}{3.5}
\newcommand{\mconstLeadersSame}{yes}
\newcommand{\mconstWorst}{all three weights $\times1.5$}

\newcommand{\fconstMoveMax}{4}
\newcommand{\fconstTauMin}{0.92}
\newcommand{\fconstRhoMin}{0.98}
\newcommand{\fconstDeltaMax}{4.7}
\newcommand{\fconstVlmMoveMax}{2}
\newcommand{\fconstVlmTauMin}{0.93}
\newcommand{\fconstOverallTauMin}{0.99}
\newcommand{\fconstCheckDev}{0.57}

\newcommand{\falsncases}{211}
\newcommand{\falsseconds}{2.0}
\newcommand{\falsRepagShortConc}{0.900}
\newcommand{\falsRepagShortSpread}{0.900}
\newcommand{\falsRepagLongConc}{0.949}
\newcommand{\falsRepagLongSpread}{0.900}
\newcommand{\falsRepagDoc}{0.692}
\newcommand{\falsRepagTableConc}{0.936}
\newcommand{\falsRepagTableSpread}{0.900}
\newcommand{\falsRepagSplitOne}{0.800}
\newcommand{\falsRepagSplitTwo}{0.900}

\newcommand{\formsTauFillEmpty}{0.40}
\newcommand{\formsFillEmptyGapMin}{2.2}
\newcommand{\formsFillEmptyGapMax}{10.9}
\newcommand{\formsTauUkUs}{0.73}
\newcommand{\formsWithoutTau}{0.95}
\newcommand{\formsWithoutMoved}{7}
\newcommand{\formsWithoutTauVlm}{1.00}
\newcommand{\formsLaneNfilled}{445}
\newcommand{\formsLaneNempty}{514}
\newcommand{\formsLaneNcheckboxes}{458}
\newcommand{\formsLaneNsignatures}{164}
\newcommand{\formsLaneCapPagesMax}{44}
\newcommand{\formsLaneCapPagesMaxVlm}{61}
\newcommand{\formsLaneGapMax}{4.3}
\newcommand{\formsTextractFive}{62.3}
\newcommand{\formsTextractRank}{20}

\newcommand{\chcapLowN}{5}
\newcommand{\chcapLowEmitMax}{5}
\newcommand{\chcapFullN}{11}
\newcommand{\chcapCondFullMin}{64.8}
\newcommand{\chcapCondFullMax}{92.7}
\newcommand{\chcapCondMedParser}{70.6}
\newcommand{\chcapEmitMedParser}{95}
\newcommand{\chcapEmitMinVlm}{8}
\newcommand{\chcapEmitMaxVlm}{92}
\newcommand{\chcapCondMinVlm}{59.8}
\newcommand{\chcapCondMaxVlm}{89.0}
\newcommand{\chcapTauCondScore}{0.77}

\newcommand{\provNParsers}{9}
\newcommand{\provNVlms}{4}
\newcommand{\provMultiN}{5}
\newcommand{\provMedConfigs}{78.3}
\newcommand{\provMedBest}{81.3}
\newcommand{\provMedMean}{76.9}

\newcommand{\nincomplete}{3}

\newcommand{\oracleFormRefs}{500}
\newcommand{\oracleFormFail}{0}
\newcommand{\oracleFormFailScored}{0}
\newcommand{\oracleFormScored}{500}
\newcommand{\oracleBlankRefs}{100}
\newcommand{\oracleBlankFail}{0}
\newcommand{\oracleCommit}{214253ef7d}

\newcommand{\labTextNextTau}{0.56}
\newcommand{\labTextNextCrit}{96}
\newcommand{\labTextNextHarm}{0}
\newcommand{\labTextSimpleTau}{0.32}
\newcommand{\labTextSimpleCrit}{50}
\newcommand{\labTextSimpleHarm}{13}
\newcommand{\labTextIsoVoneTau}{0.46}
\newcommand{\labTextIsoVoneCrit}{15}
\newcommand{\labTextIsoVoneHarm}{27}
\newcommand{\labTextVoneTau}{0.26}
\newcommand{\labTextVoneCrit}{23}
\newcommand{\labTextVoneHarm}{27}
\newcommand{\labTableNextTau}{0.49}
\newcommand{\labTableNextCrit}{94}
\newcommand{\labTableNextHarm}{0}
\newcommand{\labTableSimpleTau}{0.20}
\newcommand{\labTableSimpleCrit}{64}
\newcommand{\labTableSimpleHarm}{50}
\newcommand{\labTableIsoVoneTau}{0.06}
\newcommand{\labTableIsoVoneCrit}{15}
\newcommand{\labTableIsoVoneHarm}{100}
\newcommand{\labTableVoneTau}{0.12}
\newcommand{\labTableVoneCrit}{57}
\newcommand{\labTableVoneHarm}{75}
\newcommand{\labChartNextTau}{0.68}
\newcommand{\labChartNextCrit}{91}
\newcommand{\labChartNextHarm}{0}
\newcommand{\labChartSimpleTau}{0.11}
\newcommand{\labChartSimpleCrit}{75}
\newcommand{\labChartSimpleHarm}{87}
\newcommand{\labChartIsoVoneTau}{0.45}
\newcommand{\labChartIsoVoneCrit}{0}
\newcommand{\labChartIsoVoneHarm}{23}
\newcommand{\labChartVoneTau}{0.36}
\newcommand{\labChartVoneCrit}{78}
\newcommand{\labChartVoneHarm}{23}
\newcommand{\labFormatNextTau}{0.86}
\newcommand{\labFormatNextCrit}{100}
\newcommand{\labFormatNextHarm}{0}
\newcommand{\labFormatSimpleTau}{-0.05}
\newcommand{\labFormatSimpleCrit}{33}
\newcommand{\labFormatSimpleHarm}{15}
\newcommand{\labFormatIsoVoneTau}{0.10}
\newcommand{\labFormatIsoVoneCrit}{0}
\newcommand{\labFormatIsoVoneHarm}{100}
\newcommand{\labFormatVoneTau}{0.15}
\newcommand{\labFormatVoneCrit}{0}
\newcommand{\labFormatVoneHarm}{43}

\newcommand{\docsText}{485}
\newcommand{\docsTables}{140}
\newcommand{\docsCharts}{99}
\newcommand{\docsLayout}{364}
\newcommand{\docsForms}{244}
\newcommand{\docsComplexHeaders}{149}
\newcommand{\docsSerff}{159}
\newcommand{\docsFinancial}{234}
\newcommand{\docsDenseTables}{373}
\newcommand{\docsHeadings}{100}
\newcommand{\docsStrikeout}{126}
\newcommand{\docsUnderlined}{125}
\newcommand{\docsBlankForms}{98}
\newcommand{\docsFinancialRotated}{234}
\newcommand{\docsFinancialOriented}{234}
\newcommand{\docsWatermarks}{60}
\newcommand{\docsSplitTables}{96}
\newcommand{\docsIconTables}{98}
\newcommand{\docsDenseText}{98}
\newcommand{\docsChartPoints}{185}
\newcommand{\docsDiagrams}{114}
\newcommand{\docsFormulas}{92}
\newcommand{\docsRevisions}{98}
\newcommand{\docsLineNumbers}{98}
\newcommand{\docsInlineImages}{205}
\newcommand{\docsDecoration}{125}
\newcommand{\docsHyperlinks}{117}
\newcommand{\docsCore}{1{,}304}
\newcommand{\docsCasesCore}{2{,}611}
\newcommand{\docsFeature}{2{,}644}
\newcommand{\docsCasesFeature}{3{,}747}
\newcommand{\docsHeadline}{2{,}645}
\newcommand{\docsCasesHeadline}{4{,}317}
\newcommand{\docsAll}{3{,}896}
\newcommand{\docsCasesAll}{6{,}358}

\newcommand{\fmtPages}{480}
\newcommand{\fmtPagesHeadings}{319}
\newcommand{\fmtPagesBold}{316}
\newcommand{\fmtPagesLists}{221}
\newcommand{\fmtPagesItalic}{158}
\newcommand{\fmtPagesUnderline}{130}
\newcommand{\fmtPagesSup}{58}
\newcommand{\fmtPagesStrike}{14}
\newcommand{\fmtPagesHighlight}{13}
\newcommand{\fmtPagesSub}{7}
\newcommand{\fmtPagesCode}{6}
\newcommand{\fmtMedianItems}{10}
\newcommand{\hdrPages}{149}
\newcommand{\hdrRulesChain}{596}
\newcommand{\hdrRulesColspan}{532}
\newcommand{\hdrRulesRowspan}{97}
\newcommand{\hdrRules}{1{,}225}

\newcommand{\labFormatCapOnly}{91}
\newcommand{\labFormatPropOnly}{52}
\newcommand{\labSweepChartCritFifty}{91}
\newcommand{\labSweepChartCritHundred}{91}
\newcommand{\labSweepChartCritMax}{91}
\newcommand{\labSweepChartCritMin}{90}
\newcommand{\labSweepChartCritTen}{90}
\newcommand{\labSweepChartCritThirty}{90}
\newcommand{\labSweepChartOvrFifty}{79}
\newcommand{\labSweepChartOvrHundred}{79}
\newcommand{\labSweepChartOvrMax}{79}
\newcommand{\labSweepChartOvrMin}{77}
\newcommand{\labSweepChartOvrTen}{77}
\newcommand{\labSweepChartOvrThirty}{78}
\newcommand{\labSweepChartRankFifty}{1}
\newcommand{\labSweepChartRankHundred}{1}
\newcommand{\labSweepChartRankTen}{1}
\newcommand{\labSweepChartRankThirty}{1}
\newcommand{\labSweepFormatCritFifty}{100}
\newcommand{\labSweepFormatCritHundred}{100}
\newcommand{\labSweepFormatCritMax}{100}
\newcommand{\labSweepFormatCritMin}{100}
\newcommand{\labSweepFormatCritTen}{100}
\newcommand{\labSweepFormatCritThirty}{100}
\newcommand{\labSweepFormatOvrFifty}{93}
\newcommand{\labSweepFormatOvrHundred}{92}
\newcommand{\labSweepFormatOvrMax}{94}
\newcommand{\labSweepFormatOvrMin}{92}
\newcommand{\labSweepFormatOvrTen}{94}
\newcommand{\labSweepFormatOvrThirty}{94}
\newcommand{\labSweepFormatRankFifty}{1}
\newcommand{\labSweepFormatRankHundred}{1}
\newcommand{\labSweepFormatRankTen}{1}
\newcommand{\labSweepFormatRankThirty}{1}
\newcommand{\labSweepTableCritFifty}{94}
\newcommand{\labSweepTableCritHundred}{94}
\newcommand{\labSweepTableCritMax}{94}
\newcommand{\labSweepTableCritMin}{94}
\newcommand{\labSweepTableCritTen}{94}
\newcommand{\labSweepTableCritThirty}{94}
\newcommand{\labSweepTableOvrFifty}{73}
\newcommand{\labSweepTableOvrHundred}{72}
\newcommand{\labSweepTableOvrMax}{73}
\newcommand{\labSweepTableOvrMin}{72}
\newcommand{\labSweepTableOvrTen}{73}
\newcommand{\labSweepTableOvrThirty}{73}
\newcommand{\labSweepTableRankFifty}{1}
\newcommand{\labSweepTableRankHundred}{1}
\newcommand{\labSweepTableRankTen}{1}
\newcommand{\labSweepTableRankThirty}{1}
\newcommand{\labSweepTextCritFifty}{96}
\newcommand{\labSweepTextCritHundred}{96}
\newcommand{\labSweepTextCritMax}{96}
\newcommand{\labSweepTextCritMin}{96}
\newcommand{\labSweepTextCritTen}{96}
\newcommand{\labSweepTextCritThirty}{96}
\newcommand{\labSweepTextOvrFifty}{76}
\newcommand{\labSweepTextOvrHundred}{75}
\newcommand{\labSweepTextOvrMax}{76}
\newcommand{\labSweepTextOvrMin}{75}
\newcommand{\labSweepTextOvrTen}{75}
\newcommand{\labSweepTextOvrThirty}{76}
\newcommand{\labSweepTextRankFifty}{1}
\newcommand{\labSweepTextRankHundred}{1}
\newcommand{\labSweepTextRankTen}{1}
\newcommand{\labSweepTextRankThirty}{1}

\newcommand{\fmtVoneHarmlessMax}{67}

\newcommand{\vonevtwotau}{\vonevtwotauval}  %
\newcommand{\vonevtworho}{\vonevtworhoval}  %

\newcommand{\vonepages}{${\sim}2{,}000$}
\newcommand{\vonerules}{169{,}011}
\newcommand{\vonemethods}{14}

\newcommand{\vtwotoptenchanged}{\vtwotoptenchangedval}  %
\newcommand{\critlossshareval}{13}
\newcommand{\critlossshareHigh}{15}
\newcommand{\critlossshareMedLow}{10}
\newcommand{\critlossshareMedHigh}{13}
\newcommand{\critpageslossbest}{60}

\newcommand{\capBindBestAll}{42}
\newcommand{\capBindMaxCharts}{60}
\newcommand{\capBindMaxFmt}{40}
\newcommand{\capBindMaxForms}{84}
\newcommand{\capBindMaxTables}{49}
\newcommand{\capBindMaxText}{63}
\newcommand{\capBindMedAll}{50}
\newcommand{\capBindMedCharts}{43}
\newcommand{\capBindMedChartsWith}{45}
\newcommand{\capBindMedFmt}{33}
\newcommand{\capBindMedForms}{79}
\newcommand{\capBindMedTables}{39}
\newcommand{\capBindMedText}{54}
\newcommand{\capBindMedVlmCharts}{12}
\newcommand{\capBindMedVlmFmt}{30}
\newcommand{\capBindMedVlmForms}{80}
\newcommand{\capBindMedVlmTables}{32}
\newcommand{\capBindMedVlmText}{47}
\newcommand{\capBindMinCharts}{0}
\newcommand{\capBindMinFmt}{20}
\newcommand{\capBindMinForms}{75}
\newcommand{\capBindMinTables}{21}
\newcommand{\capBindMinText}{47}
\newcommand{\capCritBest}{0.63}
\newcommand{\capCritBestRank}{1}
\newcommand{\capCritMax}{7.13}
\newcommand{\capCritMed}{1.64}
\newcommand{\capCritMedCharts}{2.19}
\newcommand{\capCritMedFmt}{0.17}
\newcommand{\capCritMedForms}{1.77}
\newcommand{\capCritMedTables}{2.75}
\newcommand{\capCritMedText}{0.98}
\newcommand{\capCritMin}{0.63}
\newcommand{\capCritVlmBest}{0.75}
\newcommand{\capCritVlmMin}{0.75}
\newcommand{\capEvBest}{6.3}
\newcommand{\capEvBestRank}{1}
\newcommand{\capEvMax}{17.3}
\newcommand{\capEvMed}{9.7}
\newcommand{\capGapMedCharts}{5.0}
\newcommand{\capGapMedFmt}{1.1}
\newcommand{\capGapMedForms}{8.2}
\newcommand{\capGapMedTables}{4.1}
\newcommand{\capGapMedText}{6.9}
\newcommand{\capNoChartParsers}{5}

\newcommand{\critlossshare}{\critlossshareval{} to \critlossshareHigh}
\newcommand{\headlineresultbody}{\nmethods{} document-parser configurations and \nvlms{} direct vision-model configurations, \bestoverallname{} leads with \bestoverall{} of 100, the best vision model scores \bestvlm{}, and Charts separates the field most, from \resMinParserCharts{} to \resMaxParserCharts{} among the parser configurations}
\newcommand{\headlineresult}{across \headlineresultbody}   %
\newcommand{\Headlineresult}{Across \headlineresultbody}   %

\theoremstyle{definition}

\title{\pbtwo{}: Severity-Aware Evaluation of Document Parsing}
\author{
  Pierre-Lo\"{i}c Doulcet \quad
  Boyang Zhang \quad
  Sebasti\'{a}n G. Acosta \quad
  Sacha Bron \\[2pt]
  Daniel B. Ospina \quad
  Antonio Jimeno Yepes \quad
  Nutchapol Dendumrongsup \quad
  Simon Suo \\[4pt]
  {\mdseries\footnotesize\color{LIMuted}\texttt{\{pierre, boyang, sebas, sacha, daniel, antonio, nutchapol, simon\}@runllama.ai}\quad LlamaIndex}
}
\date{}

\LIPaperDate{November 2026}

\LIPDFAuthor{Pierre-Loic Doulcet, Boyang Zhang, Sebastian G. Acosta, Sacha Bron, Daniel B. Ospina, Antonio Jimeno Yepes, Nutchapol Dendumrongsup, Simon Suo}
\LIPDFSubject{Document-parsing benchmark}
\LIAbstract{Document-parsing metrics can give high scores to outputs that change important facts and penalize changes that preserve meaning.
People can often judge these consequences more easily than they can design a metric to measure them.
We introduce a method for developing parsing metrics from severity policies that specify how each error type should affect the score.
We express these policies through test scenarios that specify how a metric should behave.
We then optimize the metric implementations against these fixed expectations, using coding agents to refine comparison rules, error detectors and weights under human review.
This makes stated human priorities an explicit objective of metric design.
We apply this method in \textbf{\pbtwo{}}, a comprehensive document-parsing benchmark that combines broad capability coverage with metrics calibrated to error severity.
It covers \docsCasesHeadline{} pages across six dimensions of parsing quality: text, formatting, tables, charts, visual grounding, and forms.
We use detailed annotations and targeted difficult cases to check for changed facts, broken structural relationships, and missing content.
We take the lower of a parsing score that measures content recovery and a severity cap that penalizes detected errors, so recovering most of a page cannot hide a serious error.
Across six severity test suites, our metrics more closely match the expert judgments expressed by the scenario targets than existing metrics and penalize at most 3.4\% of the tested benign variations.
We conduct a comprehensive evaluation spanning frontier vision-language models, specialized document parsers and open-weight systems, comparing their overall rankings, performance across parsing capabilities and tradeoffs between accuracy and cost.

}
\LIHero{%
  {\centering\begingroup
\begin{tikzpicture}[font=\sffamily]
  \node[rounded corners=4pt, draw=LIPanelLine, fill=LIPanelGray, inner sep=4pt, anchor=north west, font=\scriptsize] (src) at (0,0) {%
    \setlength{\tabcolsep}{3pt}\renewcommand{\arraystretch}{1.05}%
    \begin{tabular}{@{}lcrrr@{}}
      \multicolumn{5}{@{}l}{\textbf{Revenue by region} (USD m)}\\[1pt]
      \hline
      Region & $\cdots$ & 2022 & 2023 & 2024\\ \hline
      North America & $\cdots$ & 1,153 & 1,220 & {\setlength{\fboxsep}{1pt}\colorbox{pbHi}{\strut\textbf{\textcolor{sevCrit}{1,822}}}}\\
      Latin America & $\cdots$ & 229 & 237 & 255\\
      Western Europe & $\cdots$ & 805 & 850 & 908\\
      Eastern Europe & $\cdots$ & 157 & 171 & 177\\
      \multicolumn{5}{c}{\textcolor{LIMuted}{$\cdots$ 20 more regions $\cdots$}}\\ \hline
    \end{tabular}};
  \node[anchor=south west, inner sep=1pt, font=\footnotesize\bfseries, text=LIInk] (ta) at ([yshift=2pt]src.north west) {a\enspace Revenue overstated by USD 540 million\vphantom{g}};
  \node[anchor=north west, text width=6.6cm, inner sep=1pt, font=\scriptsize, text=LIInk, align=left] (note) at ([yshift=-5pt]src.south west)
    {A parser writes \textbf{1,282} as \textbf{\textcolor{sevCrit}{1,822}}. \textcolor{LIMuted}{The other 274 cells are exact.}};
  \begin{scope}[shift={([yshift=-1.55cm]note.south west)}]
    \fill[pbNext!80!white] (0.02,-0.62) .. controls (0.02,0.12) and (1.08,0.12) .. (1.08,-0.62) -- cycle;
    \fill[LIOrange!22!white, draw=LIInk, line width=0.5pt] (0.55,0.32) circle (0.30);
    \begin{scope}\clip (0.2,0.45) rectangle (0.9,0.7); \fill[LIInk] (0.55,0.32) circle (0.30);\end{scope}
    \draw[LIInk, line width=0.5pt] (0.44,0.30) circle (0.085) (0.66,0.30) circle (0.085) (0.525,0.30) -- (0.575,0.30);
    \draw[LIInk, line width=0.5pt, line cap=round] (0.49,0.14) -- (0.61,0.14);
    \node[anchor=north, font=\tiny, text=LIMuted, inner sep=1pt] at (0.55,-0.66) {expert reader};
    \fill[white, draw=LIFlowGray, line width=0.5pt] (1.00,0.62) circle (0.045) (1.20,0.72) circle (0.07) (1.44,0.80) circle (0.095);
    \node[rounded corners=9pt, draw=LIFlowGray, line width=0.5pt, fill=white, anchor=west, text width=4.55cm, inner sep=5pt, font=\scriptsize, text=LIInk, align=left] at (1.62,0.42)
      {North America's 2024 revenue is now off by USD 540 m. Anyone who uses this table gets a wrong fact. \textbf{\textcolor{sevCrit}{The score has to show it.}}};
  \end{scope}
  \node[anchor=south west, inner sep=1pt, font=\footnotesize\bfseries, text=LIInk] at (ta.south west -| 6.8,0) {b\enspace Score of this output\vphantom{g}};
  \begin{scope}[shift={(9.9,-0.38)}]
  \fill[sevCrit, opacity=0.07] (0,0.27) rectangle (1.03,-4.28);
  \draw[LIPanelLine] (4.48,0.27) -- (4.48,-4.28);
  \draw[LIPanelLine] (0,-4.28) -- (4.48,-4.28);
  \draw[LIPanelLine] (1.03,-4.28) -- ++(0,-0.06) node[anchor=north, font=\tiny, text=LIMuted, inner sep=1.5pt] {90};
  \draw[LIPanelLine] (2.76,-4.28) -- ++(0,-0.06) node[anchor=north, font=\tiny, text=LIMuted, inner sep=1.5pt] {95};
  \draw[LIPanelLine] (4.48,-4.28) -- ++(0,-0.06) node[anchor=north, font=\tiny, text=LIMuted, inner sep=1.5pt] {100};
  \draw[LIMuted, line width=0.4pt] (0.10,-4.35) -- (0.18,-4.21) (0.18,-4.35) -- (0.26,-4.21);
  \begin{scope}[shift={(0.02,-4.90)}, scale=0.36]
    \fill[pbNext!80!white] (0.02,-0.62) .. controls (0.02,0.12) and (1.08,0.12) .. (1.08,-0.62) -- cycle;
    \fill[LIOrange!22!white, draw=LIInk, line width=0.5pt] (0.55,0.32) circle (0.30);
    \begin{scope}\clip (0.2,0.45) rectangle (0.9,0.7); \fill[LIInk] (0.55,0.32) circle (0.30);\end{scope}
    \draw[LIInk, line width=0.5pt] (0.44,0.30) circle (0.085) (0.66,0.30) circle (0.085) (0.525,0.30) -- (0.575,0.30);
    \draw[LIInk, line width=0.5pt, line cap=round] (0.49,0.14) -- (0.61,0.14);
  \end{scope}
  \node[anchor=west, font=\tiny, text=sevCrit, align=left, inner sep=1pt] at (0.46,-4.90) {the expert:\\a serious error};
  \node[anchor=east, font=\tiny, text=LIMuted, align=right, inner sep=1pt] at (4.48,-4.90) {reads as\\nearly perfect};
  \node[anchor=south, font=\tiny, text=LIMuted, align=center, inner sep=1pt] at (5.68,0.30) {agrees with\\the expert?};
  \node[anchor=east, font=\scriptsize, text=LIInk, inner sep=0pt] at (-0.12,0.00) {\textbf{\pbtwo{}}};
  \fill[pbNext] (0,-0.150) rectangle (1.035,0.150);
  \node[font=\small, text=sevHarm, inner sep=0pt] at (5.68,0.00) {\ding{51}};
  \node[anchor=west, font=\small\bfseries, text=pbNext, inner sep=0pt] at (1.135,0.00) {90};
  \node[anchor=east, font=\scriptsize, text=LIInk, inner sep=0pt] at (-0.12,-0.66) {\pbone{}};
  \fill[LIFlowGray!55!white] (0,-0.810) rectangle (4.407,-0.510);
  \node[font=\small, text=sevCrit, inner sep=0pt] at (5.68,-0.66) {\ding{55}};
  \node[anchor=west, font=\scriptsize, text=LIInk, inner sep=0pt] at (4.487,-0.66) {99.8};
  \node[anchor=east, font=\scriptsize, text=LIInk, inner sep=0pt] at (-0.12,-1.32) {OmniDocBench};
  \fill[LIFlowGray!55!white] (0,-1.470) rectangle (4.439,-1.170);
  \node[font=\small, text=sevCrit, inner sep=0pt] at (5.68,-1.32) {\ding{55}};
  \node[anchor=west, font=\scriptsize, text=LIInk, inner sep=0pt] at (4.519,-1.32) {99.9};
  \node[anchor=east, font=\scriptsize, text=LIInk, inner sep=0pt] at (-0.12,-1.98) {PubTables-1M};
  \fill[LIFlowGray!55!white] (0,-2.130) rectangle (4.460,-1.830);
  \node[font=\small, text=sevCrit, inner sep=0pt] at (5.68,-1.98) {\ding{55}};
  \node[anchor=west, font=\scriptsize, text=LIInk, inner sep=0pt] at (4.540,-1.98) {99.9};
  \node[anchor=east, font=\scriptsize, text=LIInk, inner sep=0pt] at (-0.12,-2.64) {PulseBench-Tab};
  \fill[LIFlowGray!55!white] (0,-2.790) rectangle (4.289,-2.490);
  \node[font=\small, text=sevCrit, inner sep=0pt] at (5.68,-2.64) {\ding{55}};
  \node[anchor=west, font=\scriptsize, text=LIInk, inner sep=0pt] at (4.369,-2.64) {99.4};
  \node[anchor=east, font=\scriptsize, text=LIInk, inner sep=0pt] at (-0.12,-3.30) {olmOCR-Bench};
  \fill[LIFlowGray!55!white] (0,-3.450) rectangle (4.335,-3.150);
  \node[font=\small, text=sevCrit, inner sep=0pt] at (5.68,-3.30) {\ding{55}};
  \node[anchor=west, font=\scriptsize, text=LIInk, inner sep=0pt] at (4.415,-3.30) {99.6};
  \node[anchor=east, font=\scriptsize, text=LIInk, inner sep=0pt] at (-0.12,-3.96) {RD-TableBench};
  \fill[LIFlowGray!55!white] (0,-4.110) rectangle (4.416,-3.810);
  \node[font=\small, text=sevCrit, inner sep=0pt] at (5.68,-3.96) {\ding{55}};
  \node[anchor=west, font=\scriptsize, text=LIInk, inner sep=0pt] at (4.496,-3.96) {99.8};
  \end{scope}
\end{tikzpicture}
\endgroup
\par}
  \captionof{figure}{\pbtwo{} aligns parsing scores with human judgments of error severity.}
  \label{fig:teaser}
}

\begin{document}

\maketitle
\etocdepthtag.toc{mainmatter}

\section{Introduction}
\label{sec:intro}

Document parsers supply the text, tables, and structure that retrieval systems and language-model agents use.
Errors in this output can propagate: a changed number in a financial table can become a wrong fact in a generated answer.
A useful evaluation should reflect the consequences of these errors while accepting different representations that preserve the information needed by the application.

Existing metrics do not consistently capture this distinction.
Similarity metrics average differences across an output, so a wrong digit can barely affect the score of a long page (\Cref{fig:teaser}).
Yet these metrics can penalize a table encoded with different HTML tags even when its values and header relationships are unchanged~\citep{zhong2020imagebasedtablerecognitiondata}.
Rule-based checks make individual requirements explicit.
But a pass rate gives every failed check equal weight, so a cosmetic mistake can count as much as a wrong number~\citep{poznanski2025olmocr,zhang2026parsebench}.
These limitations make it difficult to tell whether a high-scoring output preserves the information that matters.

People can often judge the consequence of a parsing error more easily than they can design a metric that measures it.
They may agree that writing 1{,}250 as 1{,}350 deserves a substantial penalty and that writing it as 1250 should cost nothing.
Implementing these expectations consistently across text, tables, charts, and other content requires many interacting rules.
We therefore start by defining a policy for how errors should affect the score, then develop metrics that implement it.

We call this a \emph{severity policy}.
It specifies how each \emph{error type}, such as changed numbers or missing fields, should affect the score.
We use test scenarios to check whether a metric follows this policy~\citep{ribeiro2020checklist,sai2021perturbation}.
Each test scenario applies one change to a correct output on a particular page.
The policy specifies how much lower the changed output should score than the correct output.
With these targets fixed, coding agents revise each metric's comparison rules, error detectors and weights.
We review the changes and keep those that bring score drops closer to the targets.
We then investigate why the metric still fails to meet the targets on other test scenarios.
This makes stated priorities an explicit objective of metric design, with the targets fixed and the implementation free to change (\Cref{sec:method}).

We apply this method in \textbf{\pbtwo{}}, a comprehensive benchmark with scores designed to reflect the impact of parsing errors.
Building on \pbone{}'s enterprise document coverage~\citep{zhang2026parsebench}, it combines detailed annotations and targeted difficult cases across \docsCasesHeadline{} pages.
We evaluate six dimensions of parsing quality: text, formatting, tables, charts, visual grounding, and forms.
Our severity policies require a single critical error to lower an otherwise correct output's score by at least ten points out of 100, regardless of page length.
Benign variations must leave the score unchanged.
A high score should require both recovering the page's content and avoiding serious errors.
We therefore take the lower of a parsing score, which measures content recovery, and a severity cap, which limits the score according to the severity of detected errors.
We report detected errors alongside scores, so users can adjust error penalties for their own applications without rerunning the parsers (\Cref{sec:dataset}).

We first compare parsing metrics on six test suites to measure how closely they follow our severity policies.
Our metrics achieve the highest agreement with the severity policies in all six test suites (\Cref{sec:val-summary}).
For example, our Formatting metric meets the required score drop in 96\% of scenarios with critical errors, compared with 5\% for \pbone{}'s formatting metric.
We then evaluate \fpeval{\nmethods+\nvlms+\nopen} configurations of document parsers, direct vision models, and open-weight systems across the six dimensions.
The best document parser scores \bestoverall{} overall, compared with \bestvlm{} for the best direct vision model.
Our severity-aware evaluation distinguishes parsers that recover similar amounts of content but differ in how often they change facts or lose important information.

Our contributions are:
\begin{itemize}[leftmargin=*,itemsep=2pt]
  \item A method for constructing parsing metrics from severity policies and test scenarios through iterative refinement by coding agents, with human review (\Cref{sec:method}).
  \item \pbtwo{}, a comprehensive document parsing benchmark that combines broad document coverage and targeted difficult cases with metrics calibrated to the impact of errors (\Cref{sec:dataset}).
  \item An evaluation of metric behavior and parsing systems, with error diagnostics, capability comparisons, and cost measurements (\Cref{sec:experiments}).
\end{itemize}

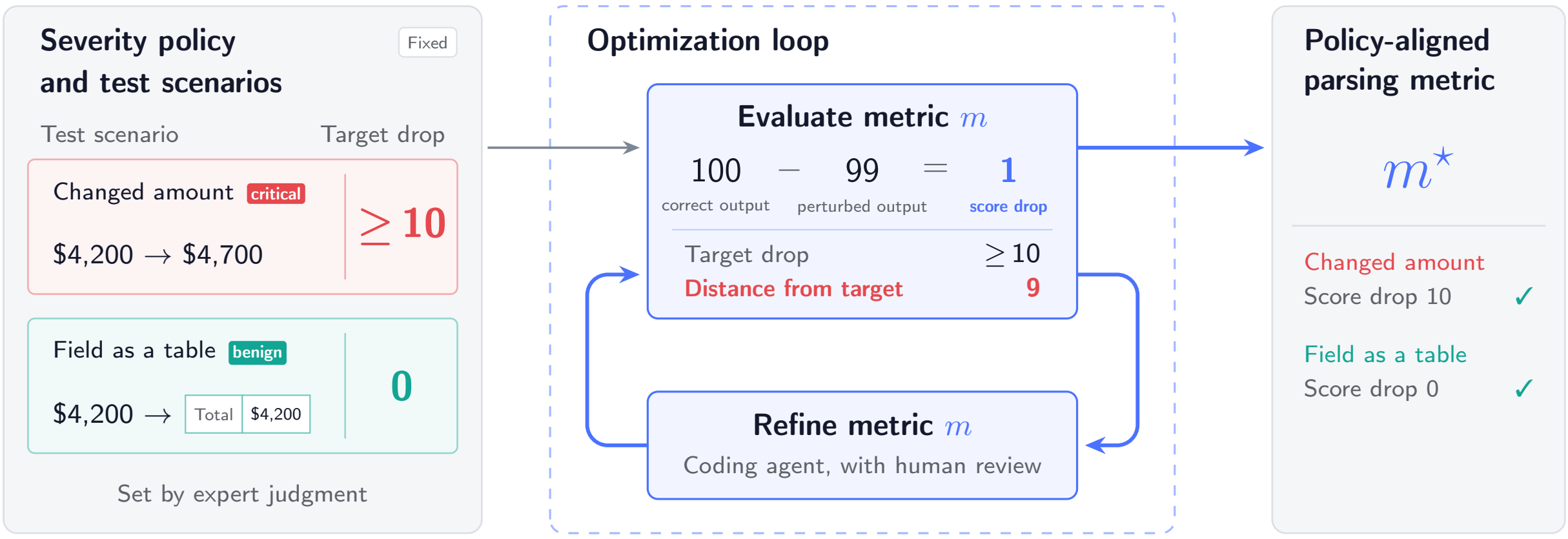
\begin{figure}[!t]
\centering
\begin{tikzpicture}[x=1cm, y=1cm,
  ttl/.style={font=\small\sffamily\bfseries, text=LIInk},
  lab/.style={font=\scriptsize\sffamily, text=LIMuted},
  pstep/.style={draw=LIBlue, fill=LIBlue!7, rounded corners=3pt, line width=0.6pt},
  card/.style={draw=#1!45, fill=#1!6, rounded corners=2pt, line width=0.5pt},
  ploop/.style={-{Stealth[length=6pt,width=5pt]}, line width=1.1pt, LIBlue},
  feed/.style={-{Stealth[length=5pt,width=4pt]}, line width=0.7pt, LIFlowGray}]

  \draw[draw=LIPanelLine, fill=LIPanelGray, rounded corners=4pt, line width=0.5pt] (0,0.35) rectangle (4.9,5.75);
  \node[ttl, anchor=west] at (0.25,5.38) {Severity policy};
  \node[ttl, anchor=west] at (0.25,4.98) {and test scenarios};
  \node[draw=LIPanelLine, fill=white, rounded corners=1.5pt, font=\tiny\sffamily, text=LIMuted, inner sep=2.5pt, anchor=east] at (4.65,5.38) {Fixed};
  \node[lab, anchor=base west] at (0.25,4.36) {Test scenario};
  \node[lab, anchor=base east] at (4.65,4.36) {Target drop};

  \draw[card=sevCrit] (0.25,2.8) rectangle (4.65,4.18);
  \node[font=\scriptsize\sffamily, text=LIInk, anchor=west] at (0.38,3.83) {Changed amount\enspace\sevbadge{sevCrit}{critical}};
  \node[font=\footnotesize\sffamily, text=LIInk, anchor=west] at (0.38,3.2) {\$4,200 $\rightarrow$ \$4,700};
  \draw[sevCrit!45, line width=0.5pt] (3.5,2.95) -- (3.5,4.03);
  \node[font=\large\sffamily\bfseries, text=sevCrit] at (4.08,3.49) {$\boldsymbol{\geq}$\,10};

  \draw[card=sevHarm] (0.25,1.17) rectangle (4.65,2.55);
  \node[font=\scriptsize\sffamily, text=LIInk, anchor=west] at (0.38,2.2) {Field as a table\enspace\sevbadge{sevHarm}{benign}};
  \node[font=\footnotesize\sffamily, text=LIInk, anchor=west] (plain) at (0.38,1.57) {\$4,200 $\rightarrow$};
  \node[draw=sevHarm!60, fill=white, font=\tiny\sffamily, text=LIMuted, inner sep=2.5pt, minimum height=11pt, anchor=west, line width=0.4pt] (key) at ([xshift=0pt]plain.east) {Total};
  \node[draw=sevHarm!60, fill=white, font=\tiny\sffamily, text=LIInk, inner sep=2.5pt, minimum height=11pt, anchor=west, line width=0.4pt] at ([xshift=-0.4pt]key.east) {\$4,200};
  \draw[sevHarm!45, line width=0.5pt] (3.5,1.32) -- (3.5,2.4);
  \node[font=\large\sffamily\bfseries, text=sevHarm] at (4.08,1.86) {0};

  \node[lab] at (2.45,0.74) {Set by expert judgment};

  \draw[LIBlue!50, dashed, rounded corners=4pt, line width=0.6pt] (5.6,0.35) rectangle (12.0,5.75);
  \node[ttl, anchor=west] at (5.85,5.38) {Optimization loop};

  \draw[pstep] (6.6,2.55) rectangle (11.0,4.95);
  \node[ttl] at (8.8,4.63) {Evaluate metric \textcolor{LIBlue}{$m$}};
  \node[font=\normalsize\sffamily, text=LIInk] at (7.3,4.07) {100};
  \node[font=\normalsize\sffamily, text=LIMuted] at (8.05,4.07) {$-$};
  \node[font=\normalsize\sffamily, text=LIInk] at (8.8,4.07) {99};
  \node[font=\normalsize\sffamily, text=LIMuted] at (9.55,4.07) {$=$};
  \node[font=\normalsize\sffamily\bfseries, text=LIBlue] at (10.3,4.07) {1};
  \node[font=\tiny\sffamily, text=LIMuted] at (7.3,3.69) {correct output};
  \node[font=\tiny\sffamily, text=LIMuted] at (8.8,3.69) {perturbed output};
  \node[font=\tiny\sffamily\bfseries, text=LIBlue] at (10.3,3.69) {score drop};
  \draw[LIBlue!35, line width=0.4pt] (6.85,3.46) -- (10.75,3.46);
  \node[lab, anchor=base west] at (6.85,3.13) {Target drop};
  \node[font=\footnotesize\sffamily, text=LIInk, anchor=base east] at (10.75,3.13) {$\geq$\,10};
  \node[font=\scriptsize\sffamily\bfseries, text=sevCrit, anchor=base west] at (6.85,2.78) {Distance from target};
  \node[font=\footnotesize\sffamily\bfseries, text=sevCrit, anchor=base east] at (10.75,2.78) {9};

  \draw[pstep] (6.6,0.7) rectangle (11.0,1.8);
  \node[ttl] at (8.8,1.47) {Refine metric \textcolor{LIBlue}{$m$}};
  \node[lab] at (8.8,1.03) {Coding agent, with human review};

  \draw[ploop, rounded corners=6pt] (11.0,3.0) -- (11.62,3.0) -- (11.62,1.25) -- (11.08,1.25);
  \draw[ploop, rounded corners=6pt] (6.6,1.25) -- (5.98,1.25) -- (5.98,3.0) -- (6.52,3.0);

  \draw[feed] (4.96,4.3) -- (6.52,4.3);

  \draw[ploop] (11.0,4.3) -- (12.92,4.3);

  \draw[draw=LIPanelLine, fill=LIPanelGray, rounded corners=4pt, line width=0.5pt] (13.0,0.35) rectangle (16.0,5.75);
  \node[ttl, anchor=west] at (13.2,5.38) {Policy-aligned};
  \node[ttl, anchor=west] at (13.2,4.98) {parsing metric};
  \node[font=\LARGE, text=LIBlue] at (14.5,4.1) {$m^{\star}$};
  \draw[LIPanelLine, line width=0.5pt] (13.2,3.5) -- (15.8,3.5);
  \node[font=\scriptsize\sffamily, text=sevCrit, anchor=base west] at (13.2,3.05) {Changed amount};
  \node[lab, anchor=base west] at (13.2,2.7) {Score drop 10};
  \node[font=\scriptsize, text=sevHarm, anchor=base east] at (15.8,2.7) {\ding{51}};
  \node[font=\scriptsize\sffamily, text=sevHarm, anchor=base west] at (13.2,2.1) {Field as a table};
  \node[lab, anchor=base west] at (13.2,1.75) {Score drop 0};
  \node[font=\scriptsize, text=sevHarm, anchor=base east] at (15.8,1.75) {\ding{51}};
\end{tikzpicture}
\caption{Experts express their severity policy through test scenarios, and we refine parsing metrics to follow that policy (\Cref{sec:metric-behavior,sec:framework,sec:metric-construction}).}
\label{fig:metric-design-process}
\end{figure}

\begin{samepage}
\section{Aligning Parsing Metrics with Severity Policies}
\label{sec:method}

Specifying how errors should affect the score is easier than designing a metric that meets those expectations.
A useful metric must reflect such judgments consistently across many kinds of content and errors.
Translating these expectations into scoring rules requires repeated implementation, testing and revision.
We therefore treat metric design as an optimization problem: we hold the expected score changes fixed and refine the metric implementation to meet them.
A \emph{severity policy} specifies how each error type should affect the score.
We record detected errors separately from their penalties so users can change the penalties and recompute scores without repeating error detection.
We express these requirements through test scenarios with target score changes (\Cref{fig:metric-design-process}).
\par
\end{samepage}

\subsection{Severity policies and test scenarios}
\label{sec:metric-behavior}
\label{sec:labs}

Which parsing errors matter depends on how the extracted information will be used.
We group parsing errors into \emph{error types}, such as changed numeric values, omitted fields, or incorrect header associations.
A \emph{severity policy} $\pi$ assigns each error type or benign variation $t$ a severity level $\pi(t)$, based on its consequences for the intended application.
We use four severity levels: critical, major, minor, and benign.
Benign variations preserve the information relevant to the application and receive no penalty.
The policy specifies the score change required for each severity level, such as a minimum drop for critical errors and no change for benign variations.

\paragraph{Assigning severity levels.}
A severity policy is expert judgment, not a measurement.
We assigned severity levels based on how each change could affect someone using the parsed content.
Users can assign different severity levels for their own applications.

\paragraph{Constructing test scenarios.}
We make the policy testable with a set of test scenarios $\mathcal S$.
A \emph{test scenario} $s \in \mathcal S$ consists of a page and its correct output, a controlled change to that output (a \emph{perturbation}), and the score change required by the policy.
One error type thus yields one scenario per page.
Each scenario $s$ has a target score-drop interval $I_\pi(s)$, the range of score drops the policy accepts.
The policy specifies a target interval for each severity level.
We use the interval for the severity level of the change being tested.

Three principles guide the construction.
A perturbation changes one thing, so the score drop measures the metric's response to that change alone.
We test whether a metric gives errors too small a penalty or penalizes benign variations that should receive none.
A metric's response to the same change can depend on the page's content and length.
We therefore apply each perturbation to many pages to check that the metric follows the policy consistently.

\paragraph{An example.}
Suppose the policy classifies changed amounts as critical and requires a score drop of at least ten points on a 0--100 scale.
We construct a test scenario that changes a reported amount from \$4,200 to \$4,700 on a particular form page.
Its target follows directly from the policy: the score must fall by at least ten points.
A second scenario presents the same field and value in a two-column table.
If the policy treats this as a benign variation, its target is no score change.
A target constrains the final score drop, not one parameter of the metric: a single change can be recorded as several errors and also lower the parsing score, so a ten-point target does not mean a ten-point weight.

We next evaluate how well a parsing metric meets these requirements.

\subsection{Evaluating parsing metrics under a severity policy}
\label{sec:framework}

We evaluate a parsing metric by checking whether its score drops meet the severity policy's targets across the test scenarios.

\paragraph{What the metric measures.}
A parsing metric assigns a score to a parsed output by comparing it with the page's annotations.
We write $m(y,r)$ for the score of output $y$ against annotations $r$, with higher scores indicating better parsing quality.

\paragraph{Measuring a metric's response.}
For each test scenario $s$, we score the correct output $y_s$ and perturbed output $\tilde y_s$ against the same annotations $r_s$.
The score drop produced by metric $m$ is
\begin{equation}
  \Delta m(s) = m(y_s,r_s) - m(\tilde y_s,r_s).
  \label{eq:metric-response}
\end{equation}
\paragraph{Distance from the policy target.}
We measure how far the observed drop falls outside the target score-drop interval $I_\pi(s)$ defined in \Cref{sec:metric-behavior}:
\begin{equation}
  d_\pi(m,s) = \operatorname{dist}\!\left(\Delta m(s),\,I_\pi(s)\right).
  \label{eq:policy-distance}
\end{equation}
The distance is zero when the score drop lies within the target interval.
Otherwise, we measure the gap to the nearest boundary.
These tests identify where the metric meets or misses the policy's requirements and guide its subsequent refinement.

\subsection{Refining metrics through scenario feedback}
\label{sec:metric-construction}
\label{sec:lab-derivation}

We use the test scenarios to guide metric development.
A coding agent proposes changes to the implementation, and we test each revision against the same policy targets to decide which changes to keep.

\paragraph{Metric implementation.}
We consider a family of candidate parsing metrics $\mathcal M$.
Candidates can differ in how they compare outputs, match corresponding content and aggregate discrepancies.
These choices determine how a metric responds to the errors and benign variations in the test scenarios.
The policy, the test scenarios and their targets are not part of the search and stay fixed.
\Cref{sec:score-page} describes how we combine parsing scores and severity caps in \pbtwo{}.

\paragraph{Optimization objective.}
We minimize the sum of squared distances defined in \Cref{sec:framework}:
\begin{equation}
  m^\star \in
  \operatorname*{arg\,min}_{m\in\mathcal M}
  \sum_{s\in\mathcal S}
  d_\pi(m,s)^2.
  \label{eq:metric-design-objective}
\end{equation}

\paragraph{Refinement process.}
We cannot search $\mathcal M$ exhaustively, so we solve the problem approximately by local search.
We start with a metric that already follows the policy well (\Cref{sec:validation}).
A coding agent then inspects the scenarios that miss their target, proposes a revision to the metric and evaluates it on the full set of test scenarios.
We review each change and keep it when it lowers this sum, breaks no scenario that already met its target, and does not overfit to a particular scenario.
We stop when we can no longer find a revision that improves the objective while meeting these conditions.
The result follows the policy more closely than the metric we started from.

\paragraph{An illustrative iteration.}
One Tables test inserts a decimal point into a numeric cell, such as changing 35 to 3.5.
The metric treated changes to numeric labels as rewording and assigned too small a penalty.
We revised the comparison rule to recognize the changed number.
Across the 45 tested tables, the share meeting the ten-point target rose from 56\% to 98\%.

\begin{figure}[!t]
\centering
\begin{tikzpicture}[x=1cm, y=1cm]
  \foreach \name/\ry in {Text/3.1, Formatting/2.48, Tables/1.86, Charts/1.24, Visual grounding/0.62, Forms/0} {
    \node[font=\scriptsize\sffamily, text=LIInk, anchor=east, inner sep=0pt] at (2.45,\ry) {\name};
  }

  \node[anchor=west, font=\small\sffamily\bfseries, text=LIInk, inner sep=0pt] at (0,3.95) {a\enspace Pages per dimension};
  \foreach \ry/\old/\new/\tot in {3.1/507/508/508, 2.48/507/861/861, 1.86/503/1753/1{,}753, 1.24/568/568/568, 0.62/500/535/535, 0/0/600/600} {
    \pgfmathsetmacro{\wo}{\old/380}
    \pgfmathsetmacro{\wn}{\new/380}
    \fill[pbNext] (2.7,\ry-0.15) rectangle (2.7+\wn,\ry+0.15);
    \ifdim\wo pt>0pt \fill[LIFlowGray!55!white] (2.7,\ry-0.15) rectangle (2.7+\wo,\ry+0.15); \fi
    \node[font=\scriptsize\sffamily, text=LIInk, anchor=west, inner sep=3pt] at (2.7+\wn,\ry) {\tot};
  }
  \fill[LIFlowGray!55!white] (2.7,-0.75) rectangle (2.92,-0.57);
  \node[font=\tiny\sffamily, text=LIMuted, anchor=west, inner sep=2.5pt] at (2.92,-0.66) {\pbone{}};
  \fill[pbNext] (4.55,-0.75) rectangle (4.77,-0.57);
  \node[font=\tiny\sffamily, text=LIMuted, anchor=west, inner sep=2.5pt] at (4.77,-0.66) {added in \pbtwo{}};

  \node[anchor=west, font=\small\sffamily\bfseries, text=LIInk, inner sep=0pt] at (8.9,3.95) {b\enspace Agreement with the severity policy};
  \foreach \ry in {3.1, 2.48, 1.86, 1.24, 0.62, 0} {
    \draw[LIPanelLine, line width=0.5pt] (9.6,\ry) -- (15.225,\ry);
  }
  \foreach \ry/\old/\new/\oldlab in {3.1/25.3/88.5/25.3, 2.48/-15.5/86.6/{$-$15.5}, 1.86/-9.7/87.8/{$-$9.7}, 1.24/26.6/86.1/26.6, 0.62/-6/82/{$-$6}} {
    \pgfmathsetmacro{\xo}{9.6+(\old+25)*0.045}
    \pgfmathsetmacro{\xn}{9.6+(\new+25)*0.045}
    \draw[LIFlowGray!55!white, line width=1.4pt] (\xo,\ry) -- (\xn,\ry);
    \fill[LIFlowGray] (\xo,\ry) circle (2.6pt);
    \fill[pbNext] (\xn,\ry) circle (3pt);
    \node[font=\tiny\sffamily, text=LIMuted, anchor=east, inner sep=4pt] at (\xo,\ry) {\oldlab};
    \node[font=\scriptsize\sffamily\bfseries, text=pbNext, anchor=west, inner sep=4pt] at (\xn,\ry) {\new};
  }
  \pgfmathsetmacro{\xn}{9.6+(97.2+25)*0.045}
  \fill[pbNext] (\xn,0) circle (3pt);
  \node[font=\scriptsize\sffamily\bfseries, text=pbNext, anchor=west, inner sep=4pt] at (\xn,0) {97.2};
  \node[font=\tiny\sffamily, text=LIMuted, anchor=south, inner sep=0pt] at (11.86,3.4) {\pbone{}};
  \node[font=\tiny\sffamily\bfseries, text=pbNext, anchor=south, inner sep=0pt] at (14.71,3.4) {\pbtwo{}};
  \foreach \v in {0,50,100} {
    \pgfmathsetmacro{\gx}{9.6+(\v+25)*0.045}
    \draw[LIPanelLine, line width=0.5pt] (\gx,-0.36) -- (\gx,-0.44);
    \node[font=\tiny\sffamily, text=LIMuted, anchor=north, inner sep=1pt] at (\gx,-0.48) {\v};
  }
  \node[font=\tiny\sffamily, text=LIMuted, anchor=north west, inner sep=1pt] at (15.5,-0.48) {PCS};
\end{tikzpicture}
\caption{\pbtwo{} covers more pages than \pbone{}, and its metrics follow the severity policy more closely (\Cref{tab:suite-results}).}
\label{fig:pb1-vs-pb2}
\end{figure}
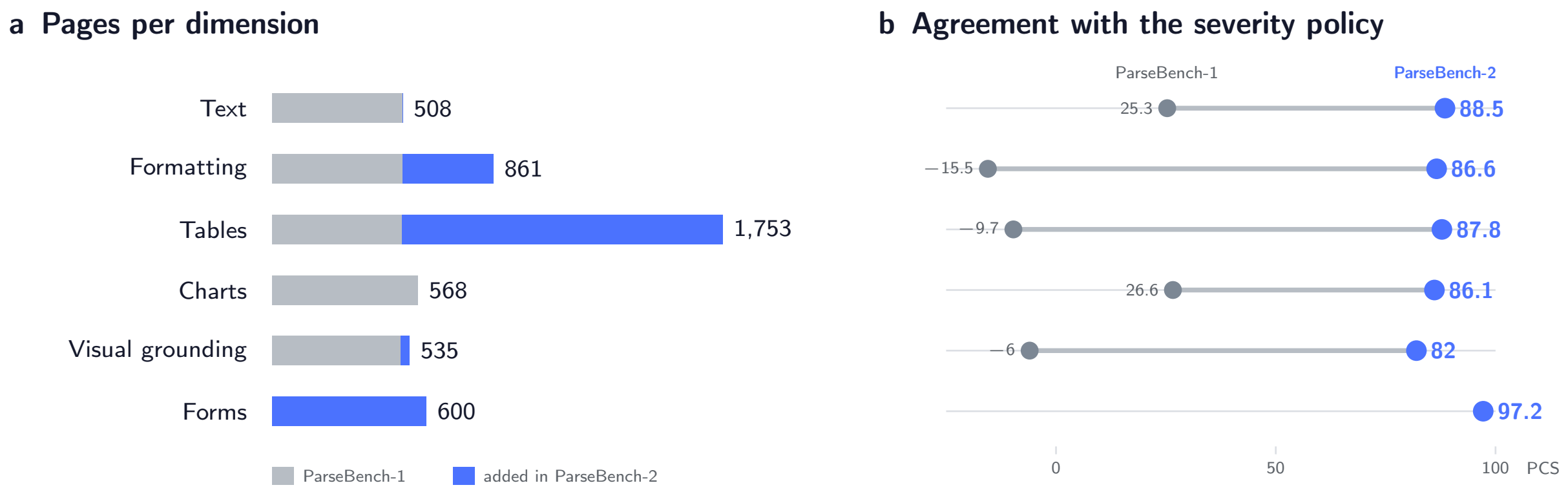

\section{\pbtwo{}: Broad Coverage and Severity-Aware Scoring}
\label{sec:dataset}

\pbtwo{} combines broad coverage of document types and difficult cases with metrics that reflect the severity of parsing errors (\Cref{fig:pb1-vs-pb2}).
We expand \pbone{} to \docsCasesHeadline{} pages across six dimensions, including cases that test whether parsers preserve table structure, strikethrough and empty form fields (\Cref{sec:dataset-coverage}).
We also provide more complete annotations, including full data tables for all 1{,}025 charts.

We use the method in \Cref{sec:method} to develop metrics from our severity policy (\Cref{sec:benchmark-policy}).
In preliminary tests, these metrics follow the policy more closely than the compared benchmark metrics in all six suites (\Cref{tab:suite-results}).
Our metrics detect errors such as changed values, incorrect header associations, and missing or invented content.
We report scores and detected errors for each dimension so users can see what a parser gets wrong and why it loses points.

\subsection{Dataset coverage and annotation quality}
\label{sec:dataset-coverage}

We evaluate six parsing capabilities as separate dimensions: Text, Formatting, Tables, Charts, Visual grounding and Forms.
We group annotated pages into collections, each focused on a parsing capability or a specific difficult case.
\pbtwo{} contains 13 collections with \docsCasesHeadline{} pages from \docsHeadline{} source documents (\Cref{app:datasets}).

We use a core collection for each dimension to evaluate how well parsers handle a broad range of documents.
We build on \pbone{}'s enterprise document coverage, including scanned, multi-column, multilingual and handwritten pages.

We add targeted collections for Formatting, Tables and Forms to test difficult cases that broad document coverage alone may miss.
We add targeted collections of financial statements, insurance rate filings, redlined text and blank public forms.

\paragraph{Expanding dataset coverage.}
\label{sec:features}
We retain 2{,}024 of \pbone{}'s 2{,}078 pages and add 2{,}293 pages.
The new Forms core collection contains 500 government-form pages from 30 countries.
We also add 86 pages for Visual grounding, one for Text and 1{,}706 in targeted collections.
We use the targeted collections to check whether parsers attach table values to the correct headers, preserve strikethrough that changes meaning, and retain empty form fields without inventing values.

\paragraph{Improving annotation quality.}
\label{sec:dataset-core}
We revise the annotations of the retained pages so that the metrics can compare complete content.
\pbone{} checked up to ten data points per chart.
We provide a full data table for each of the 1{,}025 charts so the metric can detect changed or missing values beyond those selected points.
Text annotations transcribe the whole page as printed, including running heads, page numbers and formatting.
Layout annotations have boxes tight to the ink and a box for every word.

We use frontier models to further refine the annotations, especially for charts and forms.
We review and approve the proposed corrections.
We check that each page's annotations receive a perfect score when used as both the expected content and the parser output.
We also correct any confirmed annotation errors.
We manually reviewed each parser's lowest-scoring pages in at least three rounds to identify annotation and scoring errors.

\subsection{Severity-aware scoring}
\label{sec:benchmark-policy}
\label{sec:metrics}
\label{sec:score-dims}

\paragraph{Severity policy.}
The policy reflects our own expert judgment of how an error affects a reader who uses the output.
We classify an error as critical when it can mislead a reader, such as a wrong form value.
Major errors leave a visible gap, such as a missing value with its caption intact.
Minor errors preserve the meaning, such as a typo.
Benign variations preserve the information while changing its representation.
We require a score drop of at least ten points for a critical error, at least five for a major error, and 1 to 25 for a minor error.
Benign variations must leave the score unchanged.

\paragraph{Scenarios.}
We build six severity test suites, one for each dimension.
Together, they contain 330 perturbations covering 240 error types and 90 benign variations (\Cref{tab:suites-main}).
They range from swapped column headers and a flipped negation (critical) to extra HTML markup (benign).
We list the error types, their severity levels, and the benign variations in \Cref{app:test-suites}.

\paragraph{Resulting metrics.}

For each dimension, we provide annotated document pages and a metric for scoring parser outputs.
A severity test suite checks whether that metric follows the policy.

We use the method in \Cref{sec:method} to optimize each metric for agreement with the severity policy, revising its scoring rules as needed.
For example, the metrics detect table values linked to the wrong headers, lost strikethrough that changes meaning, and incorrectly extracted chart values.
They allow small differences when chart values must be estimated from an axis.
The code defines the resulting metrics, and \Cref{app:metrics} provides agent-generated explanations of how they work.

\paragraph{Adding a severity cap.}
\label{sec:score}
\label{sec:score-page}
We optimize the parsing metric $m(y,r)$ introduced in \Cref{sec:framework} against the severity policy's targets.
To enforce the severity policy more strictly, we supplement the parsing metric with a severity cap $c$ for Text, Formatting, Tables, Charts and Forms.
The cap limits how high the score can be when serious errors are detected, even if most of the page is correct.
The cap is 1 minus the combined penalties for detected errors, with a minimum of zero.
\phantomsection\label{eq:framework-cap}
The final benchmark score is
\begin{equation}
  S = \min\!\left(m(y,r),c\right).
  \label{eq:headline}
\end{equation}

\subsection{Score aggregation}
\label{sec:score-compose}
\label{sec:score-feature}

\paragraph{Dimension scores.}
We want each dimension to reflect both broad parsing ability and performance on difficult cases.
Averaging all pages together would give larger collections more influence on the dimension score.

We therefore compute a separate average for the core collection and for each group of targeted cases (\Cref{sec:features}).
We then combine these averages using fixed weights, so adding pages to a group does not increase its influence:
\begin{equation}
  D \;=\; \sum_g w_g\,\bar S_g.
  \label{eq:slots}
\end{equation}
Here, $D$ is the dimension score, $\bar S_g$ is the average page score in group $g$, and $w_g$ is that group's weight.
The weights sum to one.

We choose the weights to balance broad parsing ability with performance on difficult cases (\Cref{tab:app-dataset-composition}).

\paragraph{Overall score.}
We average the six dimension scores, giving each parsing capability equal influence on the overall score.
We report the dimension scores alongside the overall score so users can see where a parser is strong or weak.

\section{Experiments}
\label{sec:experiments}

We evaluate both the scoring metrics and the parsing systems.
We use test scenarios to compare how well the metrics align with the severity policy (\Cref{sec:validation}).
We then evaluate parsing systems on \pbtwo{} to identify their strengths, diagnose serious errors, and compare accuracy and cost (\Cref{sec:parser-evaluation}).

\subsection{Metric evaluation}
\label{sec:validation}

We test how closely the metrics developed with our method follow the severity policy in \Cref{sec:benchmark-policy}.
We compare them with the \pbone{} metrics~\citep{zhang2026parsebench} and other baseline metrics on the six severity test suites in \Cref{tab:suites-main}.

\paragraph{Evaluation protocol.}
\label{sec:lab-protocol}
We evaluate each metric on the same test scenarios, which cover parsing errors and benign variations.
We measure the rate of policy violations at each severity level.
We summarize policy agreement with a \emph{policy-compliance summary} (PCS), which starts at 100 and decreases as test scenarios violate the policy:
\begin{equation}
  \text{PCS} = 100\left[1-\left(
    0.5\,v_{\text{critical}}
    + 0.3\,v_{\text{major}}
    + 0.2\,v_{\text{minor}}
    + v_{\text{benign}}
  \right)\right].
  \label{eq:pcs}
\end{equation}
Here, $v_\ell$ is the fraction of scenarios that violate the policy at severity level $\ell$.
For errors, a violation means that the score drop falls outside the required interval (\Cref{sec:metric-behavior}).
For benign variations, it means a drop of 0.5 points or more on the 0--100 page-score scale.

We give benign variations the same total weight as the three error levels combined.
This prevents a metric from scoring highly simply by penalizing every difference.
All metrics are evaluated on the same PCS scale: higher is better, 100 is the ideal value, and scores can be negative.

We evaluate the metrics on test scenarios generated by applying each perturbation to correct outputs from 100 pages in each of the five real-page suites.
We evaluate the metrics on 16{,}059 graded scenarios across these five suites.

To measure each component's contribution, we compare PCS for the parsing metric $m$, the severity cap $c$ and their combination $S$.

\paragraph{Results.}
\label{sec:val-summary}
The \pbtwo{} metrics have the highest PCS among the benchmark metrics in all six suites (\Cref{tab:suite-results}).
The full scenario results show that our metrics meet most policy targets at every severity level in each suite (\Cref{app:test-suites}).
For example, the Formatting metric meets the target on 96\% of critical cases, while the \pbone{} Semantic Formatting metric meets it on 5\%.

Without the severity cap, the parsing metrics already achieve higher PCS than all compared baselines on Text, Formatting and Charts.
The severity cap makes a large contribution to policy agreement.
For Tables, adding the cap raises PCS from 50.0 for the parsing metric alone to 87.8 for the combined score.

The metrics penalize at most 3.4\% of tested benign-variation scenarios in any suite.
For Tables, our metric penalizes none, compared with 80\% for \pbone{}'s GTRM metric.

\begin{table}[!htbp]
\RCsetup\small\setlength{\tabcolsep}{3pt}
\begin{tabularx}{\linewidth}{lrrrr>{\raggedright\arraybackslash}Xr>{\raggedright\arraybackslash}X}
\toprule
& \multicolumn{3}{c}{\sffamily\bfseries \pbtwo{}} & \multicolumn{2}{c}{\sffamily\bfseries \pbone{}~\citep{zhang2026parsebench}} & \multicolumn{2}{c}{\sffamily\bfseries Best other baseline} \\
\cmidrule(lr){2-4}\cmidrule(lr){5-6}\cmidrule(lr){7-8}
\addlinespace[3pt]
\RChead{Test suite} & \RChead{\shortstack{Parsing\\metric $m$}} & \RChead{\shortstack{Severity\\cap $c$}} & \RChead{\shortstack{Final\\score $S$}} & \RChead{Score} & \RChead{Metric} & \RChead{Score} & \RChead{Metric} \\
\midrule\RCstart
Text & 52.2 & 88.3 & \textbf{88.5} & 25.3 & Content Faithfulness & 29.5 & OmniDocBench $1-$NED~\citep{ouyang2024omnidocbench} \RCnext
Formatting & 40.7 & 86.5 & \textbf{86.6} & $-15.5$ & Semantic Formatting & $-4.7$ & FinTOC harmonic mean (adapted)~\citep{maarouf2021fintoc} \RCnext
Tables & 50.0 & 87.5 & \textbf{87.8} & $-9.7$ & GTRM & 58.2 & PulseBench-Tab T-LAG~\citep{pandey2026pulsebenchtab} \RCnext
Charts & 46.4 & 83.3 & \textbf{86.1} & 26.6 & \textsc{ChartData\allowbreak Point\allowbreak Match} & 30.6 & EpiCurveBench~\citep{berkane2026epicurvebench} \RCnext
Forms & 20.4 & 97.2 & \textbf{97.2} & -- & -- & 37.4 & Generic exact-match field F1 (ours) \RCnext
Visual grounding & -- & -- & \textbf{82} & $-6$ & EPR & -- & -- \RClast
\bottomrule
\end{tabularx}
\caption{Policy-compliance summary (PCS; higher is better) for \pbtwo{} and baseline metrics.}
\label{tab:suite-results}
\end{table}

\paragraph{Additional results: LLM-as-judge.}
We also evaluate LLM-as-judge as an alternative to our scoring metrics.
We use Claude Haiku 5.5 with the PAIR prompt~\citep{reducto2026pair}, which evaluates how faithfully outputs preserve a page's information.
We test it on the error types and benign variations from the five suites.
The judge sees a page and two outputs, then chooses the better output or declares a tie.
We use one page per error type and one judgment per comparison.

The judge usually prefers a correct output over one containing an error.
When both outputs contain errors, however, it selects the output with the more severe error in 29\% of comparisons.
It also often favors one output over another when they differ only by a benign variation.
These results motivate defining how errors and benign variations should affect the score, then refining metrics to meet those expectations.

We use the published PAIR prompt as a starting point.
Future work could explore model selection and prompt refinement to better align LLM judgments with severity policies.

\subsection{Parser evaluation}
\label{sec:parser-evaluation}
\label{sec:results}
\label{sec:robustness-tradeoffs}
\label{sec:leaderboard}

\paragraph{Evaluation protocol.}
\label{sec:setup-pipelines}
We evaluate \nmethods{} document-parser configurations, \nvlms{} frontier VLM configurations and \nopen{} open-weight model configurations.
We compare these systems on \docsCasesHeadline{} pages across six dimensions and rank them within each group.
\Cref{app:experimental-setup} describes the full evaluation protocol.

\Cref{tab:top3} shows the ten highest-scoring systems across all three groups.
\Cref{app:full-results} reports results for every system.

\begin{table}[!t]
\RCsetup\small\setlength{\tabcolsep}{2pt}
\begin{tabularx}{\linewidth}{@{}>{\raggedright\arraybackslash}Xlr@{\hspace{9pt}}rrrrrr@{\hspace{9pt}}r@{}}
\toprule
& & & \multicolumn{6}{c}{\sffamily\bfseries Scores by dimension} & \sffamily\bfseries Cost \\
\cmidrule(lr){4-9}
\addlinespace[3pt]
\RChead{System} & \RChead{Type} & \RChead{Overall} & \RChead{Text} & \RChead{Format.} & \RChead{Tables} & \RChead{Charts} & \RChead{\shortstack[r]{Visual\\grounding}} & \RChead{Forms} & \RChead{\$/page} \\
\midrule
\RCstart
LlamaParse (Agentic Plus) & Parser & \textbf{90.1} & 90.3 & \textbf{93.6} & \textbf{94.6} & \textbf{92.6} & 87.5 & \textbf{81.8} & 0.028\RCnext
LlamaParse (Agentic) & Parser & 86.0 & 87.0 & 92.7 & 89.9 & 77.4 & 88.3 & 80.9 & 0.013\RCnext
Claude Opus 5.5 & Frontier VLM & 85.7 & \textbf{92.7} & 89.4 & 94.1 & 74.1 & 86.2 & 77.8 & 0.065\RCnext
LlamaParse (Cost Effective) & Parser & 84.6 & 85.5 & 91.9 & 87.2 & 73.1 & \textbf{88.6} & 81.2 & 0.004\RCnext
Gemini 4 argon (low) & Frontier VLM & 84.2 & 89.7 & 89.2 & 92.8 & 68.2 & 88.1 & 77.2 & 0.027\RCnext
Reducto & Parser & 83.9 & 81.9 & 82.8 & 84.9 & 89.1 & 84.1 & 80.9 & 0.030\RCnext
Extend (Parse 2) & Parser & 83.2 & 80.2 & 83.3 & 84.2 & 85.1 & 84.8 & 81.5 & 0.025\RCnext
Pulse (Ultra 2) & Parser & 82.7 & 86.1 & 88.7 & 84.7 & 86.8 & 73.5 & 76.4 & 0.015\RCnext
Extend (Parse Light) & Parser & 81.6 & 78.3 & 78.7 & 81.4 & 85.4 & 84.7 & 81.2 & 0.006\RCnext
Datalab (Accurate) & Parser & 81.3 & 91.1 & 92.5 & 90.3 & 67.8 & 70.1 & 75.8 & 0.010\RClast

\bottomrule
\end{tabularx}
\caption{Ten highest-scoring systems on \pbtwo{} using default configurations. Bold marks the best score in each column across all ranked systems.}
\label{tab:top3}
\end{table}

\paragraph{Key findings.}
\begin{itemize}[leftmargin=*,itemsep=2pt]
\item \textbf{Document parsers lead overall.}
\bestoverallname{} scores \bestoverall{} overall, compared with \bestvlm{} for the best frontier VLM and \resOpenBest{} for the best open-weight model.
\item \textbf{Strengths differ across dimensions.}
Frontier VLMs have higher median scores on Text and Tables, while document parsers have higher medians on Charts and Visual grounding.
\item \textbf{Charts reveal large capability gaps.}
Scores range from \resAllMinCharts{} to \resAllMaxCharts{}.
Several default configurations describe charts without extracting their data.
\end{itemize}

\phantomsection
\paragraph{Cost and accuracy}
\label{sec:results-cost}

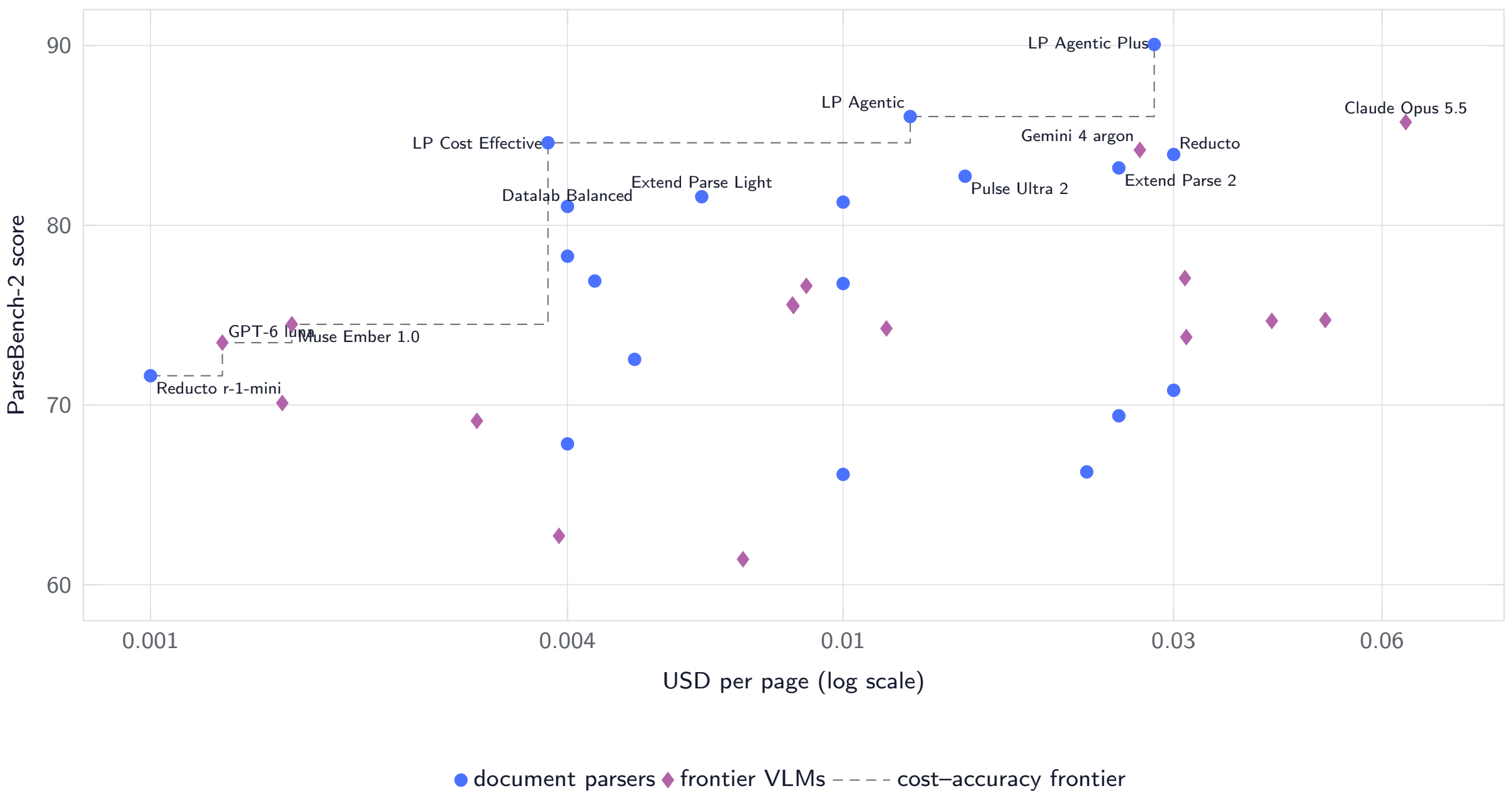
\begin{figure}[!t]
\centering
\begin{tikzpicture}
\begin{axis}[pbaxis, width=\textwidth, height=8cm, xmode=log, xmin=0.0008, xmax=0.09,
  xlabel={USD per page (log scale)}, ylabel={\pbtwo{} score}, ymin=58, ymax=92, ytick={60,70,80,90}, grid=major,
  legend style={at={(0.5,-0.22)}, anchor=north, legend columns=3}, xtick={0.001,0.004,0.01,0.03,0.06}, xticklabels={0.001,0.004,0.01,0.03,0.06},
  costlabel/.style={nodes near coords, point meta=explicit symbolic, visualization depends on={value \thisrow{anchor}\as\costanchor},
    nodes near coords style={font=\tiny, text=LIInk, inner sep=1.5pt, anchor=\costanchor}}]
\addplot[only marks, mark=*, mark size=1.8pt, pbNext, keep if={group}{parser}, costlabel] table[x=cost, y=v2, meta=label] {results_cost.dat};
\addlegendentry{document parsers}
\addplot[only marks, mark=diamond*, mark size=2.2pt, pbLegacy, keep if={group}{vlm}, costlabel] table[x=cost, y=v2, meta=label] {results_cost.dat};
\addlegendentry{frontier VLMs}
\addplot[const plot, densely dashed, thin, black!55] table[x=cost, y=v2] {results_cost_frontier.dat};
\addlegendentry{cost\textendash accuracy frontier}
\end{axis}
\end{tikzpicture}
\caption{\pbtwo{} score versus average cost per page for \nmethods{} document parsers and \nvlms{} frontier VLMs. The dashed line shows the highest score achieved within each cost budget. LP denotes LlamaParse.}
\label{fig:quality-cost}

\end{figure}

Costs vary \resCostRatio{}-fold among document parsers, but higher-priced systems do not consistently score better (\Cref{fig:quality-cost}).
The highest-scoring parser costs less than half as much per page as the best frontier VLM.
Frontier VLMs also offer competitive choices at the low-cost end.
Among parsers on the cost--accuracy frontier, the first step above the cheapest option gives the largest score gain.
Each later step costs two to four times as much for a few additional points.

\begin{figure}[!tbp]
\centering
\begin{tikzpicture}[x=1cm,y=1cm,
  exhead/.style={anchor=north west,inner sep=0pt,font=\small\sffamily\bfseries,text=LIInk},
  exsub/.style={anchor=north west,inner sep=0pt,font=\scriptsize\sffamily,text=LIMuted},
  extext/.style={anchor=north west,inner sep=0pt,align=left,text width=5.25cm,font=\scriptsize\sffamily,text=LIInk}]
  \node[exhead] at (0,6.85) {a\enspace Source page};
  \node[exhead] at (4.4,6.85) {b\enspace LlamaParse};
  \node[exsub] at (4.4,6.43) {Agentic Plus};
  \node[exhead] at (10.6,6.85) {c\enspace Mistral OCR 4.1};
  \node[exsub] at (10.6,6.43) {annotation};

  \node[anchor=north,inner sep=0pt] at (1.95,6.03)
    {\includegraphics[width=3.9cm,height=5.6cm,keepaspectratio]{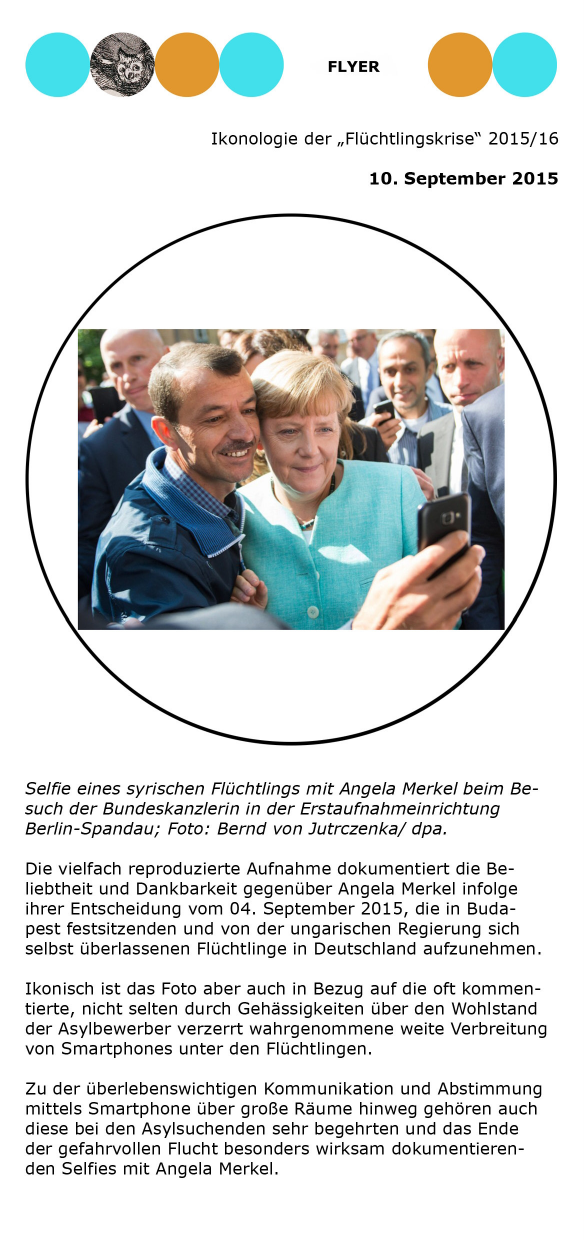}};
  \draw[LIPanelLine,rounded corners=2pt] (4.4,0.55) rectangle (10.2,6.05);
  \draw[LIPanelLine,rounded corners=2pt] (10.6,0.55) rectangle (16.4,6.05);

  \node[extext,font=\scriptsize\sffamily\bfseries] at (4.68,5.78) {FLYER};
  \node[extext] at (4.68,5.15) {Ikonologie der \quotedblbase Fl\"uchtlingskrise\textquotedblleft{} 2015/16};
  \node[extext,font=\scriptsize\sffamily\bfseries] at (4.68,4.35) {10. September 2015};
  \node[extext] at (4.68,3.53) {\textit{Selfie eines syrischen Fl\"uchtlings mit Angela Merkel beim Besuch der Bundeskanzlerin \ldots}};
  \node[extext] at (4.68,2.10) {Die vielfach reproduzierte Aufnahme dokumentiert die Beliebtheit und Dankbarkeit gegen\"uber Angela Merkel \ldots};

  \node[extext] at (10.88,5.78) {FLYER};
  \filldraw[fill=sevCrit!7!white,draw=sevCrit!65,line width=0.6pt,rounded corners=2pt]
    (10.85,3.56) rectangle (16.15,5.37);
  \node[anchor=north west,inner sep=0pt,font=\tiny\sffamily\bfseries,text=sevCrit!85!black]
    at (11.05,5.18) {INVENTED CONTENT};
  \node[anchor=north west,inner sep=0pt,font=\scriptsize\sffamily,text=LIInk] at (11.05,4.72) {%
    \setlength{\tabcolsep}{7pt}\renewcommand{\arraystretch}{1.15}%
    \begin{tabular}{@{}lr@{}}
      \textbf{Category} & \textbf{Percentage} \\
      Category A & 50\% \\
      Category B & 50\%
    \end{tabular}};
  \node[extext] at (10.88,3.20) {Ikonologie der \quotedblbase Fl\"uchtlingskrise\textquotedblleft{} 2015/16};
  \node[extext] at (10.88,2.39) {10. September 2015};
  \node[extext] at (10.88,1.75) {\textit{Selfie eines syrischen Fl\"uchtlings mit Angela Merkel beim Besuch der Bundeskanzlerin \ldots}};

  \draw[LIPanelLine] (0,0.25) -- (16.4,0.25);
  \node[anchor=west,inner sep=0pt,font=\scriptsize\sffamily,text=LIMuted] at (0,-0.15) {Content Faithfulness};
  \node[font=\small\sffamily,text=LIMuted] at (7.3,-0.15) {98.6};
  \node[font=\small\sffamily,text=LIMuted] at (13.5,-0.15) {98.6};
  \node[anchor=west,inner sep=0pt,font=\scriptsize\sffamily\bfseries,text=LIInk] at (0,-0.78) {\pbtwo{} Text};
  \node[font=\small\sffamily\bfseries,text=LIBlue!85!black] at (7.3,-0.78) {100};
  \node[font=\small\sffamily\bfseries,text=sevCrit!85!black] at (13.5,-0.78) {58.7};
\end{tikzpicture}
\caption{Content Faithfulness gives both outputs the same score despite the invented percentage table in Mistral's output. \pbtwo{} identifies the error. Output excerpts are rendered with image placeholders omitted and the invented content highlighted.}
\label{fig:severity-text-example}
\end{figure}

\phantomsection
\paragraph{Revealing severe errors in parser outputs.}
\label{sec:results-errors}
Severity-aligned evaluation distinguishes outputs that existing metrics score alike.
On one text page, Mistral OCR 4.1 (annotation) invents a percentage table absent from the document, while LlamaParse (Agentic Plus) preserves the annotated content.
Content Faithfulness assigns both outputs 98.6.
\pbtwo{} identifies the invented content and scores the outputs 58.7 and 100, respectively (\Cref{fig:severity-text-example}).
\Cref{app:cap-per-system} reports error counts and severity-cap results for every system.

\section{Related Work}
\label{sec:related}

\subsection{Benchmarks for document parsing}

\paragraph{Subtask benchmarks.}
Early benchmarks each target one parsing capability on a narrow set of documents.
PubTabNet~\citep{zhong2020imagebasedtablerecognitiondata} and FinTabNet~\citep{zheng2020gte} evaluate table structure recognition on academic and financial documents, DocLayNet~\citep{pfitzmann2022doclaynet} evaluates layout detection but not content extraction, and ChartQA~\citep{masry2022chartqa} tests chart reasoning through question answering rather than extraction of the plotted data.

\paragraph{End-to-end benchmarks.}
OmniDocBench~\citep{ouyang2024omnidocbench}, OCRBench~v2~\citep{fu2025ocrbenchv2} and READoc~\citep{li2024readoc} evaluate several capabilities together, mostly on academic and general documents, and score structured elements mainly with similarity metrics.
olmOCR-Bench~\citep{poznanski2025olmocr} uses binary property tests rather than whole-output similarity, drawing many of its pages from arXiv papers.
\pbone{}~\citep{zhang2026parsebench} evaluates five capabilities on enterprise documents and checks charts at selected values.

\paragraph{Gaps addressed by \pbtwo{}.}
Most of these benchmarks either focus on narrow document types or check only selected content, so many errors on real enterprise pages go untested.
\pbtwo{} closes this gap by extending \pbone{}'s enterprise coverage with forms, revised annotations, complete chart data tables and targeted hard cases (\Cref{sec:dataset}).

\subsection{Evaluating parse quality}

\paragraph{Similarity metrics.}
Edit distance and BLEU~\citep{papineni2002bleu} compare text, TEDS~\citep{zhong2020imagebasedtablerecognitiondata} compares table trees, and GriTS~\citep{smock2023gritsgridtablesimilarity} compares table grids.
They give partial credit, but a changed digit adds little over a long page, while equivalent markup can cost points.

\paragraph{Rule-based tests.}
Rule-based benchmarks such as olmOCR-Bench and \pbone{}~\citep{poznanski2025olmocr,zhang2026parsebench} make each requirement an explicit check, but a pass rate counts every check alike, and a strict rule can fail an output that gives the same content in another valid format.

\paragraph{Model-based judges.}
LLM judges trade hand-written rules for model judgment.
They accept equivalent formats and reworded content without a rule for each case, and \citet{horn2026beyond} report stronger agreement with human judgments than TEDS and GriTS for tables.
This flexibility comes at a cost: strong judges approach human agreement, yet still show position and verbosity bias~\citep{zheng2023judging}, so consistent and interpretable scores need a capable model and a detailed rubric that defines each error.
PAIR~\citep{reducto2026pair} ranks two parses with an LLM judge against a shared baseline, so its win rates depend on the judge and the baseline.

\paragraph{Gaps addressed by \pbtwo{}.}
None of these approaches states how much each error should cost.
\pbtwo{} instead scores parsing errors under an explicit severity policy (\Cref{sec:score}).

\subsection{Evaluating and designing metrics}

\paragraph{Prior work.}
In machine translation, MQM-based evaluation has experts mark each error with a severity and weights the counts~\citep{freitag2021experts}, while learned metrics such as BLEURT~\citep{sellam2020bleurt} and COMET~\citep{rei2020comet} predict human ratings and leave their priorities implicit.
Controlled tests check behavior directly: CheckList~\citep{ribeiro2020checklist} organizes tests around capabilities and expected behavior, and \citet{sai2021perturbation} apply perturbations to evaluation metrics.

\paragraph{Gaps addressed by \pbtwo{}.}
What these precedents leave open is how document-parsing metrics should respond to errors of different severity.
We therefore define severity targets for parsing errors and test metrics against them (\Cref{sec:metric-behavior,sec:validation}).

\section{Conclusion}
\label{sec:conclusion}

We introduced a method for developing parsing metrics from explicit severity policies.
We express how errors should affect scores through test scenarios, then use coding agents with human review to refine metrics against those expectations.
\pbtwo{} applies this method across six parsing dimensions, combining broad document coverage, detailed annotations and targeted difficult cases with metrics designed to reflect error severity.

On the test scenarios used for metric development, our metrics follow the stated policy more closely than the compared alternatives while rarely penalizing benign variations (\Cref{sec:val-summary}).
Our evaluation reveals differences in parser capabilities and consequential errors that high content-recovery scores can obscure.
By making evaluation priorities explicit and reporting the errors behind each score, \pbtwo{} helps users understand parser failures and adjust penalties to reflect what matters in their applications.
Future work will test how well these scores reflect errors in downstream applications.

\clearpage
\appendix
\crefalias{section}{appendix}
\crefalias{subsection}{appendix}
\crefalias{subsubsection}{appendix}

\etocdepthtag.toc{appendixmatter}
\begingroup
\etocsettagdepth{mainmatter}{none}
\etocsettagdepth{appendixmatter}{subsubsection}
\etocsettocstyle{\section*{Appendix Contents}\small}{}
\tableofcontents
\endgroup
\clearpage

\section{Benchmark Definition}
\label{app:benchmark-definition}
\subsection{Dataset Composition}
\label{app:datasets}

The \pbtwo{} dataset groups annotated pages into core and targeted collections.
Core collections cover a broad range of documents, while targeted collections test specific difficult cases.
\Cref{tab:app-dataset-composition} lists the collections, their page counts and their weights in each dimension score (\Cref{eq:slots}).
Each page has its own annotations and forms one test case.

\begin{table}[!htbp]
\RCsetup
\begin{tabularx}{\linewidth}{>{\raggedright\arraybackslash}Xlcr}
\toprule
\RChead{Collection} & \RChead{Role} & \RChead{Weight} & \RChead{Pages} \\
\midrule
\RCgroup{Text}{508 pages}
Text & core & 1 & 508 \RClast
\RCgroup{Formatting}{861 pages, 508 shared with Text}
Text pages$^{\dagger}$ & core & $\frac{3}{4}$ & 508 \RCnext
Headings & targeted & $\frac{1}{8}$ & 100 \RCnext
Strikethrough$^{\ddagger}$ & targeted & $\frac{1}{16}$ & 128 \RCnext
Underline$^{\ddagger}$ & targeted & $\frac{1}{16}$ & 125 \RClast
\RCgroup{Tables}{1{,}753 pages}
Tables & core & $\frac{1}{2}$ & 500 \RCnext
Complex table headers & targeted & $\frac{1}{8}$ & 149 \RCnext
SERFF rate tables & targeted & $\frac{1}{8}$ & 412 \RCnext
Financial tables & targeted & $\frac{1}{8}$ & 319 \RCnext
Dense tables & targeted & $\frac{1}{8}$ & 373 \RClast
\RCgroup{Charts}{568 pages}
Charts & core & 1 & 568 \RClast
\RCgroup{Visual grounding}{535 pages}
Layout & core & 1 & 535 \RClast
\RCgroup{Forms}{600 pages}
Forms & core & $\frac{7}{8}$ & 500 \RCnext
Blank public forms & targeted & $\frac{1}{8}$ & 100 \RClast
\midrule\RCstart
\textbf{Full dataset} & & & \textbf{\docsCasesHeadline} \RClast
\bottomrule
\end{tabularx}
\caption{Collections, page counts and weights in each dimension score. $^{\dagger}$Formatting reuses the Text core pages. We count these pages once in the dataset total. $^{\ddagger}$Strikethrough and underline share a targeted group with total weight $\frac{1}{8}$.}
\label{tab:app-dataset-composition}
\end{table}

\clearpage
\subsection{The Severity Policy}
\label{app:severity-policy}
\label{app:suites-overview}

Our severity policy specifies how parsing errors and benign variations should affect a score (\Cref{sec:benchmark-policy}).
The six test suites contain 330 perturbations covering 240 error types and 90 benign variations.
We assign severity levels according to how each change could affect someone using the parsed content.
The policy sets a target score drop for each level:
\begin{itemize}[leftmargin=*,itemsep=1pt]
  \item \textbf{Critical} (87 error types): a change that can mislead a reader, such as a wrong amount or a value under the wrong header. The required drop is at least 10 points.
  \item \textbf{Major} (71): lost or misplaced content that leaves a visible gap, such as a missing value or a moved block. The required drop is at least 5 points.
  \item \textbf{Minor} (58): an error that preserves meaning, such as a typo or a missing bold style. The required drop is 1 to 25 points.
  \item \textbf{Benign} (90 variations): a change in representation that preserves the relevant information, such as using Markdown instead of HTML for the same table. The score should not change.
\end{itemize}
In the evaluation, a drop below 0.5 points on a benign variation counts as no penalty.
We also list 24 \emph{debatable} error types, for which the appropriate severity depends on the application's requirements.
We exclude these from the graded evaluation.
These assignments reflect our judgment, and users can choose different severity levels for their applications.

\paragraph{Coverage.}
\Cref{tab:suites-main} summarizes the error types and benign variations in each dimension and gives examples.

\begin{table}[!htbp]
\RCsetup\small\setlength{\tabcolsep}{4pt}
\begin{tabularx}{\linewidth}{lccccc>{\raggedright\arraybackslash}X}
\toprule
& \multicolumn{5}{c}{\sffamily\bfseries Error types by level, benign variations} & \\
\cmidrule(lr){2-6}
\RChead{Test suite} & \RChead{C} & \RChead{M} & \RChead{m} & \RChead{H} & \RChead{D} & \RChead{Example perturbation (C; M; m; H)} \\
\midrule\RCstart
Text & 13 & 16 & 5 & 8 & 9 & negation flipped; words dropped; spelling typos; paragraphs run together \RCnext
Formatting & 5 & 8 & 24 & 15 & 3 & strikethrough lost; insertion underline lost on a redline; bold and italic stripped; \texttt{\_\_bold\_\_} for \texttt{**bold**} \RCnext
Tables & 27 & 18 & 9 & 5 & 3 & column headers swapped; two rows merged; empty row inserted; extra HTML markup \RCnext
Charts & 22 & 9 & 3 & 14 & 5 & order of magnitude off; series missing; series names abbreviated; units written in cells \RCnext
Forms & 13 & 13 & 8 & 34 & 3 & digit misread in a value; value dropped, caption kept; typo in a text value; fields written as a two-column table \RCnext
Visual grounding & 7 & 7 & 9 & 14 & 1 & biggest paragraph missed; 50 junk boxes; tiny icon missed; every class wrong \RClast
\bottomrule
\end{tabularx}
\caption{Coverage of the six severity test suites \prelim{}. C, M, m: critical, major and minor error types. H: benign variations. D: debatable error types, excluded from grading. The last column gives examples in the order C, M, m, H.}
\label{tab:suites-main}
\end{table}

\paragraph{Evaluation coverage.}
For the main metric comparison (\Cref{tab:suite-results}), we apply each perturbation to correct outputs from 100 real pages in its suite.
A page and a perturbation together form a test scenario.
We skip a scenario if the perturbation leaves the output unchanged, for example when a page has no strikethrough to remove.
This leaves 16{,}059 graded scenarios across the five real-page suites.
Ten error types (3 critical, 3 major and 4 minor) and 12 benign variations change none of the sampled pages, so this evaluation does not test them.

Visual grounding uses one synthetic page with fixed variants of the predicted boxes and text.
We evaluate its score drops against the same severity targets, although the metric has no cap.

\paragraph{Reading the policy tables.}
\Cref{tab:policy-text,tab:policy-formatting,tab:policy-tables,tab:policy-charts,tab:policy-forms,tab:policy-layout} list the perturbations by dimension and severity level.
Within each level, we order them from most to least severe according to our ratings.
The category groups related changes, and the description explains how each perturbation changes a correct output.
\Cref{app:test-suites} reports whether each metric's response meets the policy target.

{\RClongsetup\scriptsize\setlength{\tabcolsep}{4pt}\renewcommand{\arraystretch}{1.05}
\begin{longtable}{@{}>{\raggedright\arraybackslash}p{4.3cm}>{\raggedright\arraybackslash}p{2.7cm}>{\raggedright\arraybackslash}p{8.4cm}@{}}
\caption{Severity policy for Text: 43 error types (13 critical, 16 major, 5 minor, 9 debatable) and 8 benign variations.}\label{tab:policy-text}\\
\toprule
\RChead{Perturbation} & \RChead{Category} & \RChead{Description} \\
\midrule
\endfirsthead
\multicolumn{3}{l}{\scriptsize\sffamily\color{LIMuted}Severity policy for Text, continued}\\
\toprule
\RChead{Perturbation} & \RChead{Category} & \RChead{Description} \\
\midrule
\endhead
\bottomrule
\endlastfoot
\RCband{3}{CRITICAL ERRORS} \\*
Negation flipped & Word errors & "not", "no" and "never" are dropped, or "not" is added after is, are, was, will or can. One or two words change and the meaning inverts. \\
Output cut off & Missing or extra content & The output stops early, as with a token limit. \\
Date with day and month swapped & Wrong facts & A date's day and month are exchanged, such as March 4 $\to$ April 3 or 10 November $\to$ 11 October. The resulting date looks plausible but is incorrect. \\
Paragraphs dropped & Missing or extra content & Whole paragraphs disappear, as when a region is missed by detection. \\
Two values swapped between clauses & Wrong facts & Two figures in one paragraph trade places (about 62\% \ldots{} the remaining 38\% becomes about 38\% \ldots{} the remaining 62\%). Every number is still on the page, each in the wrong clause. \\
Wrong digits & Character errors & One digit is changed in some numbers, dates and phone numbers. Text is otherwise perfect, so this is the error that matters most and moves word metrics least. \\
Hallucinated summary sentence & Missing or extra content & A fluent closing sentence that is not on the page is added after the last paragraph, with a believable figure: "In summary, the report shows a 12\% improvement over the previous year." \\
Invented sentences & Missing or extra content & Plausible sentences that are not on the page are added inside paragraphs. \\
OCR confusion in an identifier & Wrong facts & Digits in an identifier are replaced by similar-looking letters (0 $\to$ O, 1 $\to$ l), such as CA3310021 $\to$ CA33lOO2l or 01632 $\to$ Ol632. The identifier becomes incorrect despite the small visual change. \\
Read straight across columns & Reading order & The classic layout failure: each output line joins the left column's line with the right column's line at the same height, so sentences from different columns are spliced together. \\
Unit changed & Wrong facts & Units next to numbers are misread (mg/L $\to$ g/L, \pounds{}3.20 $\to$ \$3.20, 20 km $\to$ 20 m). Numbers are right, the quantities are wrong. \\
Email address corrupted & Wrong facts & One character changes in each email address (quality@riversidewater.org $\to$ quality@riversidewaters.org), so the address no longer works. \\
Names and places misspelled & Word errors & Proper nouns are changed by one or two letters (Riverside to Riverview, Norco to Norca). Small character change, large factual error. \\
\addlinespace[2pt]
\RCband{3}{MAJOR ERRORS} \\*
Real-word substitutions & Word errors & Words are replaced by other valid words, such as form/from, then/than, of/or and on/in. Some content words are replaced by other words on the page. These substitutions can evade a spell checker. \\
\% and currency signs dropped & Wrong facts & Percent and currency signs disappear (62\% $\to$ 62, \pounds{}86.50 $\to$ 86.50), so readers cannot tell shares from counts or amounts from plain numbers. \\
OCR character confusions & Character errors & Typical OCR confusions such as rn$\to$m, l$\to$1, O$\to$0, e$\to$c and lost accents. \\
Words dropped & Character errors & Individual words go missing, as with low-contrast or faint text. \\
Paragraph repeated & Missing or extra content & A paragraph is emitted two or more times, a common VLM looping failure. \\
Phrase omitted & Word errors & A run of 3 to 8 consecutive words disappears from some paragraphs, as when a line inside a paragraph is skipped. \\
Random words inserted & Word errors & Words that are not on the page appear inside sentences, as small hallucinations. \\
Sentence cut at a column break & Missing or extra content & The last paragraph of the first column stops in the middle of its final sentence, as when a parser loses the lines that continue at the top of the next column. The rest of the page is intact. \\
Footnote spliced mid-sentence & Reading order & A short closing note (the last paragraph of the page) is moved into the middle of the last sentence of the paragraph before it, on its own line, as when a parser reads a footnote at a page or column break. All words are present. \\
Paragraphs shuffled & Reading order & Some blocks are moved to random positions, like a layout model that orders regions wrongly. \\
Adjacent words swapped & Word errors & Pairs of neighboring words exchange positions, such as "water delivered" and "delivered water". All words remain present. \\
Headings dropped & Missing or extra content & Section headings, such as Testing and treatment or Results at a glance, are omitted. The paragraphs below them remain intact. \\
List items' contents swapped & Reading order & In a bulleted list the first two items trade places while the markers stay put: each item is complete, but the list now reads in the wrong order. Only pages with a list change. \\
One sentence repeated & Missing or extra content & A single sentence is emitted twice in a row inside its paragraph, a small looping failure. \\
Table of contents duplicated & Missing or extra content & A block of short entries (here the results table, read like a table of contents) is emitted twice in a row, a looping failure on lists of short lines. Only pages with such a block change. \\
Sentences swapped in a paragraph & Reading order & In paragraphs with several sentences, the first two sentences trade places. All words are present. \\
\addlinespace[2pt]
\RCband{3}{MINOR ERRORS} \\*
Spelling typos & Character errors & Letters inside words are inserted, deleted, doubled, swapped with a neighbour or replaced by a nearby keyboard key ("treatmnet", "chloramnie", "deliverred"). \\
Look-alike characters & Character errors & Some Latin letters are replaced by identical-looking Cyrillic ones (a, e and o). The page looks right but search and copy-paste break. \\
Date format changed & Formatting only & Dates are written in another order (March 4 $\to$ 4 March, 14 October $\to$ October 14). Same dates, different notation than the page. \\
Thousands separators removed & Formatting only & Large numbers lose their separators (312,480 $\to$ 312480, 18,906 $\to$ 18906). Same values, different notation than the page. \\
Parentheses dropped & Character errors & Round brackets disappear, including in phone numbers ((951) 351-6170 $\to$ 951 351-6170) and asides. \\
\addlinespace[2pt]
\RCband{3}{DEBATABLE ERROR TYPES (listed, not graded)} \\*
Accents stripped & Character errors & Diacritics are removed: informaci\'on becomes informacion, $\mu$g becomes ug. Common in ASCII-only OCR engines. \\
Columns in reverse order & Reading order & Every block is transcribed correctly, but the right-hand column appears before the left-hand column. \\
Header and footer dropped & Page furniture & The running header and footer are omitted. ParseBench v2.1 counts them as content. olmOCR-bench requires their omission, while OmniDocBench ignores them. \\
Headings in capitals & Formatting only & The title and headings are transcribed in capitals, as when a parser copies a small-caps style. \\
Line-end hyphenation kept & Formatting only & Line breaks are kept and words split at the end of a line keep their hyphen ("distri-\textbackslash{}nbution"). The words themselves are broken in the output. \\
List items run together & Formatting only & List items lose their bullets and line breaks and become one paragraph. The words remain, but the list structure is lost. \\
Punctuation dropped & Character errors & Commas, periods, semicolons and parentheses go missing. Words and numbers are intact, though decimal points inside numbers are kept. \\
Table flattened to lines & Formatting only & Each table cell becomes its own line. Every value survives, but the row and column structure is lost. \\
Words split or joined & Character errors & Word boundaries are wrong: a space disappears between two words or appears inside one ("thedistribution", "distri bution"). Letters are all present. \\
\addlinespace[2pt]
\RCband{3}{BENIGN VARIATIONS} \\*
Different Markdown style & Formatting only & Headings become bold lines and list bullets switch from - to *. Same text, different markup. \\
Extra blank lines & Formatting only & Extra empty lines appear between paragraphs and after headings. Text is identical. \\
Hard line breaks & Formatting only & Paragraphs keep the line breaks of the printed column. In Markdown this is still one paragraph and the text is identical. \\
Hyphenated line breaks kept & Formatting only & Up to six long words keep the hyphen of a printed line break, with the break ("distri-\textbackslash{}nbution") or as a space ("informa- tion"). Lines are not otherwise rewrapped and the words are the page's words. \\
Paragraphs run together & Formatting only & Paragraph breaks inside each column are lost, so consecutive paragraphs come out as one block. Words and order are unchanged. \\
Random emphasis added & Formatting only & Some words are wrapped in bold or italic markers that the page does not have. Text is identical. \\
Table as HTML & Formatting only & The table is emitted as an HTML table instead of a pipe table. Cell text is identical. \\
Typographic characters & Formatting only & Straight quotes become curly, hyphens become en dashes, fi becomes a single ligature character and spaces before units become non-breaking. \\
\addlinespace[2pt]
\end{longtable}}

{\RClongsetup\scriptsize\setlength{\tabcolsep}{4pt}\renewcommand{\arraystretch}{1.05}
\begin{longtable}{@{}>{\raggedright\arraybackslash}p{4.3cm}>{\raggedright\arraybackslash}p{2.7cm}>{\raggedright\arraybackslash}p{8.4cm}@{}}
\caption{Severity policy for Formatting: 40 error types (5 critical, 8 major, 24 minor, 3 debatable) and 15 benign variations.}\label{tab:policy-formatting}\\
\toprule
\RChead{Perturbation} & \RChead{Category} & \RChead{Description} \\
\midrule
\endfirsthead
\multicolumn{3}{l}{\scriptsize\sffamily\color{LIMuted}Severity policy for Formatting, continued}\\
\toprule
\RChead{Perturbation} & \RChead{Category} & \RChead{Description} \\
\midrule
\endhead
\bottomrule
\endlastfoot
\RCband{3}{CRITICAL ERRORS} \\*
Strikethrough lost & Strikethrough and scripts & Deleted text loses its strikethrough and reads as live text: a voided clause or an old fee reads as current. \\
Exponent lost (10$^9$ $\to$ 109) & Strikethrough and scripts & A superscript exponent becomes ordinary digits glued to its base: 10$<$sup$>$5$<$/sup$>$ reads as 105, 10$<$sup$>$-3$<$/sup$>$ as 10-3. \\
Live text struck through & Strikethrough and scripts & A phrase of live text is struck through: the reader takes it as deleted. \\
Digit turned into exponent & Strikethrough and scripts & The last digit of a plain number is set as a superscript: 2026 reads as 202$^6$. \\
Exponent written as subscript & Strikethrough and scripts & Exponents are emitted as subscripts: 10$<$sub$>$5$<$/sub$>$. \\
\addlinespace[2pt]
\RCband{3}{MAJOR ERRORS} \\*
All headings turned into body text & Headings & Every heading is plain text: the page has no structure left. \\
Insertion underline lost (redline) & Strikethrough and scripts & On a redline, underline marks inserted text. Without it the reader cannot tell new terms from old ones. \\
One heading turned into body text & Headings & A section heading is emitted as an ordinary paragraph. \\
List items run together & Lists & The items of one list are joined into a single line of text. \\
Parent and child levels swapped & Headings & A section and its first subsection swap levels: the subsection now contains its parent. \\
All headings at one level & Headings & Every heading is \#\#: title, sections and subsections are siblings. \\
Body paragraph turned into heading & Headings & A full sentence of body text is emitted as a heading. \\
List nesting flattened & Lists & Sub-items lose their indentation and become siblings of their parent item. \\
\addlinespace[2pt]
\RCband{3}{MINOR ERRORS} \\*
All headings shown as bold lines & Headings & Every heading is a bold line, a common habit of OCR-first parsers. \\
All bold and italic stripped & Emphasis & The parser emits no emphasis at all: every bold and italic span is plain text. \\
Heading shown as a bold line & Headings & A heading is emitted as a standalone bold line: it still looks like a heading but is no longer one. \\
List markers dropped & Lists & List bullets and numbers are omitted. Each item remains on its own line. \\
Code block emitted as text & Lists & A fenced code block loses its fences and reads as a paragraph. \\
Footnote and unit superscripts lost & Strikethrough and scripts & Superscripts after a word or symbol become plain digits: billion1, site.12, m2. \\
Numbered list turned into bullets & Lists & A numbered list comes out with bullets (or bullets with numbers). \\
Bold lost & Emphasis & Bold phrases and bold label lines lose their bold. \\
One bold phrase lost & Emphasis & A single inline bold phrase loses its bold. \\
One heading one level too deep & Headings & A heading is one level deeper than it should be, so it hangs under the wrong parent. \\
Short line promoted to heading & Headings & A short line of body text becomes a heading. \\
Subscripts lost (CO$_2$ $\to$ CO2) & Strikethrough and scripts & Chemical subscripts become plain digits. Readable, but the formula is no longer marked up. \\
Underline lost & Emphasis & Underlined text (a cross-reference, a defined term) is plain. Only on pages without strikethrough. \\
Italic lost & Emphasis & Italic phrases (species, defined terms, variables) become plain. \\
Item nested under its sibling & Lists & A top-level item is indented under the item before it. \\
List item split in two & Lists & A wrapped list item is read as two items. \\
Paragraph turned into a bullet & Lists & A body paragraph is emitted as a list item. \\
Whole paragraph in bold & Emphasis & A body paragraph is emitted entirely in bold. \\
Bold added to plain words & Emphasis & Two plain words in running text are set in bold. \\
Bold rendered as italic & Emphasis & An inline bold phrase comes out in italic. \\
Highlight lost & Emphasis & A highlighted warning loses its highlight. \\
Bold span extended by a word & Emphasis & The bold run swallows the next word: **Lakeside plant** instead of **Lakeside** plant. \\
Inline code lost & Emphasis & Inline code (a command, a function name) is plain text. \\
Italic added to plain words & Emphasis & Two plain words in running text are set in italic. \\
\addlinespace[2pt]
\RCband{3}{DEBATABLE ERROR TYPES (listed, not graded)} \\*
Bold label line promoted to heading & Headings & A standalone bold line ("Procedure") becomes a \#\#\# heading. The reference treats bold lines as the lowest title level. \\
Second half of the page missing & Equivalent syntax & The output stops half way. Formatting of what is there is perfect. \\
Struck text omitted & Strikethrough and scripts & The struck text is left out entirely. The live text is right, but the record of what was deleted is gone. \\
\addlinespace[2pt]
\RCband{3}{BENIGN VARIATIONS} \\*
All heading levels one deeper & Equivalent syntax & \# becomes \#\#, \#\# becomes \#\#\#: the hierarchy is the same. \\
Bold split word by word & Equivalent syntax & **heating** **demand** instead of **heating demand**. \\
Closing \#s and bold inside headings & Equivalent syntax & \#\# **Abstract** \#\# instead of \#\# Abstract. \\
Every item numbered 1. & Equivalent syntax & Markdown renumbers 1. 1. 1. as 1, 2, 3. \\
Everything written as HTML & Equivalent syntax & $<$h2$>$, $<$b$>$, $<$i$>$, $<$del$>$, $<$ul$>$$<$li$>$ and $<$p$>$ instead of Markdown. \\
HTML entities & Equivalent syntax & \&amp;, \&\#39; and \&nbsp; instead of the characters. \\
LaTeX exponents and indices & Equivalent syntax & \$10\^{}\{5\}\$ and \$\textbackslash{}mathrm\{CO\}\_\{2\}\$ instead of HTML tags. \\
Ordinal superscript lost (1st) & Strikethrough and scripts & 1$<$sup$>$st$<$/sup$>$ written as 1st. Nothing a reader needs is lost. \\
Output wrapped in ```markdown & Equivalent syntax & The whole page arrives inside a ```markdown fence, as some VLMs return it. \\
Punctuation outside the bold & Equivalent syntax & **Electric**: instead of **Electric:**. \\
Setext headings & Equivalent syntax & Title underlined with ===, sections with ---. \\
Spacing and indentation & Equivalent syntax & Extra blank lines, trailing spaces and 4-space nesting. \\
Unicode superscripts and subscripts & Equivalent syntax & 10$^5$, CO$_2$ and m$^2$ written with Unicode characters. \\
\_\_bold\_\_ and \_italic\_ & Equivalent syntax & Underscore emphasis instead of asterisks. \\
\textbullet{} bullets and 1) numbers & Equivalent syntax & Bullet glyphs and parenthesis numbering instead of - and 1. \\
\addlinespace[2pt]
\end{longtable}}

{\RClongsetup\scriptsize\setlength{\tabcolsep}{4pt}\renewcommand{\arraystretch}{1.05}
\begin{longtable}{@{}>{\raggedright\arraybackslash}p{4.3cm}>{\raggedright\arraybackslash}p{2.7cm}>{\raggedright\arraybackslash}p{8.4cm}@{}}
\caption{Severity policy for Tables: 57 error types (27 critical, 18 major, 9 minor, 3 debatable) and 5 benign variations.}\label{tab:policy-tables}\\
\toprule
\RChead{Perturbation} & \RChead{Category} & \RChead{Description} \\
\midrule
\endfirsthead
\multicolumn{3}{l}{\scriptsize\sffamily\color{LIMuted}Severity policy for Tables, continued}\\
\toprule
\RChead{Perturbation} & \RChead{Category} & \RChead{Description} \\
\midrule
\endhead
\bottomrule
\endlastfoot
\RCband{3}{CRITICAL ERRORS} \\*
Table missing entirely & Missing or extra & The parser returns no table at all, only a placeholder line, as when it fails on a complex layout. \\
Column headers swapped & Values in the wrong place & Two column headers trade places, so both columns are labelled with the other's period or measure. \\
Decimal point inserted & Numbers & A speck or a line break is read as a decimal point inside a whole number: 97 $\to$ 9.7. \\
Decimal point misplaced & Numbers & The decimal point is lost or moves one place: 13.8 $\to$ 138, 0.41 $\to$ 4.1. \\
Headers shifted one column & Values in the wrong place & The lowest header row slides one column right, so every column is labelled with its left neighbour's header. \\
Minus signs added & Numbers & Some positive values come out negative, as when a dash or a parenthesis next to the number is misread. \\
Sign or comparator lost & Numbers & Accounting negatives lose their parentheses, minus signs vanish, and $<$ or $>$ disappear from ranges: (21,330) $\to$ 21,330, $<$200 $\to$ 200. \\
Wrong digits & Cell content & One digit is changed in some numeric cells. \\
Wrong year in a column header & Values in the wrong place & One column header reads another year (2023 $\to$ 2022), so a whole column of figures is attributed to the wrong period. \\
Empty cell skipped, values shift left & Structure & In some rows an empty cell is dropped, so every value after it slides one column left. \\
Footnote mark glued to a number & Numbers & A superscript footnote mark is read as one more digit of the number: 1,284$^1$ $\to$ 1,2841. \\
Last row (total) dropped & Missing or extra & The last body row, often the total, is left out, as when a parser stops at a rule above the total line. \\
Plausible row invented & Missing or extra & A row that is not on the page is added before the last row, with a believable label ("Other") and believable numbers. \\
Row labels swapped & Values in the wrong place & Two row labels trade places, so both rows carry the other row's numbers. \\
Rows missing & Missing or extra & Some data rows are left out. \\
Two values swapped within a row & Values in the wrong place & In one row the values of the first two value columns trade places (2023 and 2024 swapped), so both periods carry the other's figure. \\
Values shifted one row & Values in the wrong place & Row labels stay in place but every row's values come from the row below (the last row takes the first row's values), as when a parser loses one row boundary. Every number is now under the wrong label. \\
Values swapped between rows & Values in the wrong place & A value trades places with the value in another row of the same column, as when a parser misreads row boundaries. \\
Values swapped within a row & Values in the wrong place & In some rows two values trade columns, so each sits under the other's header. Every value is still present. \\
Column missing & Missing or extra & One data column is left out entirely. \\
Header rows missing & Row and column structure & Column headers are omitted. The data rows remain intact, but their columns no longer have labels. \\
OCR character confusions & Cell content & Character confusions such as 0$\to$O, 1$\to$l, 5$\to$S, rn$\to$m inside cells. \\
Adjacent cells merged & Structure & Two neighbouring values end up in one cell ("48,215 41,906") and the row loses a column. \\
Empty cells filled in & Text in cells & Cells that are empty on the page get invented content (a flag, a zero or a dash with a number), a typical hallucination on sparse tables. \\
European number format & Numbers & Thousands and decimal separators are swapped (48,215 $\to$ 48.215, 0.92 $\to$ 0,92). Each number now reads as a different value to anyone using the other convention. \\
Invented extra table & Missing or extra & A second small table that is not on the page is added after the real one. \\
Table transposed & Structure & Rows become columns. Every value keeps its row and column labels, but the layout differs from the page. \\
\addlinespace[2pt]
\RCband{3}{MAJOR ERRORS} \\*
Two rows merged & Row and column structure & Two neighbouring rows collapse into one, each cell holding both values ("9,412 7,208"). \\
Values missing & Text in cells & Some filled cells come out empty, as with faint or overprinted text. \\
Merged-cell spans lost & Structure & Merged cells lose their colspan and rowspan attributes. A header that covered two columns now covers one, shifting subsequent cells left and making rows uneven. Cell text remains unchanged. \\
Two tables merged into one & Structure & Two separate tables on the page come out as one table: the second table's header becomes an ordinary row and its rows are padded to the first table's width. Only applies to pages with two tables (the Two tables on one page sample). \\
Units dropped & Numbers & Units next to numbers disappear and a Units column is left blank: 450 W $\to$ 450, mg/dL $\to$ (empty). \\
Values rounded & Numbers & Decimal values lose a digit of precision (0.92 $\to$ 0.9, 13.8 $\to$ 14, 55.8\% $\to$ 56\%). \\
Group labels missing & Row and column structure & Labels that span several rows (Lipid panel, Metabolic) are left blank, so those rows lose their grouping. \\
One cell split in two & Row and column structure & In one row a long label is split across two cells, pushing every value in that row one column to the right. \\
Table split into several tables & Structure & One table is split into several tables, as at a page or section boundary. Only the first part retains the header. \\
Long cells truncated & Text in cells & Longer cell texts are cut after their first word or two, as when text overflows a narrow column. \\
Row label wrapped into a new row & Row and column structure & A long row label wraps onto a second line that becomes its own row with empty values ("Research and" / "development"). \\
Body rows reordered & Structure & Some data rows are moved. Each row is intact, but the table no longer reads in the page's order. \\
Title row dropped & Row and column structure & The table's title row (Consolidated statement of operations, USD thousands) is lost, along with the unit it states. Applies to tables with a title row. \\
Two columns swapped & Structure & Two data columns trade places, headers included. Every value stays under its own header. \\
Two header rows joined into one & Structure & Multi-row headers are collapsed into one row by joining the texts of each column ("Three months ended Sep 30, 2024"). \\
Column duplicated & Missing or extra & One data column appears twice side by side. \\
Row labels abbreviated & Text in cells & Row labels are shortened to abbreviations the page does not use (Research and development $\to$ R\&D, North America $\to$ NA, Total cholesterol $\to$ Total chol.). \\
Row repeated & Missing or extra & One data row appears twice in a row. \\
\addlinespace[2pt]
\RCband{3}{MINOR ERRORS} \\*
Column headers reworded & Values in the wrong place & Column headers are paraphrased (Three months ended $\to$ 3 months ended, Reference range $\to$ Ref. range, Region $\to$ Area). The columns still mean the same thing to a careful reader, but the labels no longer match the page. \\
Header repeated mid-table & Row and column structure & The header rows appear a second time in the middle of the body, as when a table continues across a page break and the repeated header is kept as data. \\
Empty column inserted & Row and column structure & A blank column appears at the right edge, as when a ruling line or margin is read as a column. \\
Empty row inserted & Row and column structure & A blank spacer row appears between the header and the data, as when vertical whitespace on the page is read as a row. \\
Number format changed & Cell content & Thousands separators are removed and accounting negatives become minus signs: 48,215 $\to$ 48215, (21,330) $\to$ -21330. Same values. \\
Trailing zeros dropped & Numbers & Decimals lose their trailing zeros (0.60 $\to$ 0.6, 12.0 $\to$ 12). The value is equal but its stated precision is gone. \\
Letter case changed & Cell content & Row labels are written in capitals. \\
Currency symbol moved & Numbers & Currency symbols are written after the number (\$0.41 $\to$ 0.41 \$). Same amount, different notation. Applies to tables with currency values. \\
Punctuation dropped inside cells & Text in cells & Commas, periods and hyphens inside text cells are lost ("Net income per share, diluted" $\to$ "Net income per share diluted", "Asia-Pacific" $\to$ "Asia Pacific"). Numbers are untouched. \\
\addlinespace[2pt]
\RCband{3}{DEBATABLE ERROR TYPES (listed, not graded)} \\*
Header cells not marked as headers & Format & Header cells use td instead of th. The text and layout remain identical, but the markup no longer identifies the cells as headers. \\
Markdown pipe table & Format & The table as a Markdown pipe table. Cell text is identical, but spans cannot be expressed: a spanning cell keeps its text in the first position and leaves the rest empty, and only the first row counts as a header. \\
Merged cells repeated in every position & Format & Each merged cell is replaced by ordinary cells containing repeated copies of its text. For example, "Three months ended" is repeated across two columns or "Lipid panel" across three rows. The visible content is preserved, but the merge is lost. \\
\addlinespace[2pt]
\RCband{3}{BENIGN VARIATIONS} \\*
Bold row labels & Cell content & Row labels are wrapped in Markdown bold markers. \\
Empty cells as dashes & Cell content & Blank cells are written as an em dash. \\
Extra HTML markup (thead, styles, bold) & Format & Same table with thead and tbody sections, inline style attributes and bold header text. \\
Title above the table & Format & A spanning title row is emitted as a heading just above the table instead of inside it. \\
Typographic variants & Cell content & En dashes become hyphens, $\times$ becomes x, $\mu$ becomes u, and a non-breaking space appears before units. \\
\addlinespace[2pt]
\end{longtable}}

{\RClongsetup\scriptsize\setlength{\tabcolsep}{4pt}\renewcommand{\arraystretch}{1.05}
\begin{longtable}{@{}>{\raggedright\arraybackslash}p{4.3cm}>{\raggedright\arraybackslash}p{2.7cm}>{\raggedright\arraybackslash}p{8.4cm}@{}}
\caption{Severity policy for Charts: 39 error types (22 critical, 9 major, 3 minor, 5 debatable) and 14 benign variations.}\label{tab:policy-charts}\\
\toprule
\RChead{Perturbation} & \RChead{Category} & \RChead{Description} \\
\midrule
\endfirsthead
\multicolumn{3}{l}{\scriptsize\sffamily\color{LIMuted}Severity policy for Charts, continued}\\
\toprule
\RChead{Perturbation} & \RChead{Category} & \RChead{Description} \\
\midrule
\endhead
\bottomrule
\endlastfoot
\RCband{3}{CRITICAL ERRORS} \\*
Negative signs lost & Wrong numbers & Bars below the axis are transcribed as positive values. Only applies to charts with negative values. \\
Order of magnitude off & Wrong numbers & A decimal point is misread, so a value is ten times too large or too small (412.5 $\to$ 41.25). \\
Scale lost & Scale and units & Values are written in the base unit without saying so, 412500000 instead of 412.5 million, or as fractions without a \% sign. Nothing in the output tells the reader the scale. \\
Cumulative values read & Wrong numbers & Stacked series are read from the top of each stack instead of by their individual heights. Later series therefore include the values of earlier ones. \\
Digit typo & Values & One digit is wrong in some values, as from a misread label. \\
Pie slice dropped, rest renormalized & Missing or invented data & The last slice of a pie is left out and the remaining shares are rescaled to add up to 100\%, so every share left is wrong too. Only applies to pie charts. \\
Secondary axis read on the primary scale & Wrong numbers & The last series belongs to a right-hand axis but is read against the left one, so its values come out scaled by the ratio of the two axes ($\times$4 here). Only applies to charts with several series. \\
Series names swapped & Structure & The first two series get each other's names, as from misreading the legend colors. Every value sits under the wrong label. \\
Shares instead of values & Wrong numbers & Values are replaced by their share of the category total, such as North America 46.6\% instead of 412.5. The labels remain correct, but the quantities change. \\
Two series summed into one & Missing or invented data & The first two series are added together into one column named after both, as when stacked segments are read as a single bar. Only applies to charts with several series. \\
Values copied from the neighbouring chart & Values attached to the wrong label & Two charts share the same layout, categories and series but have different values. The first chart's extracted table uses the second chart's values. The second chart's table follows separately. \\
Values shifted one category & Structure & Every series is shifted by one position along the axis, so each value sits next to the wrong category label. The first category is left empty. \\
Wrong unit stated & Labels & The numbers are right but the stated unit is wrong ("Values in USD thousands" for a chart in millions, "Values in percentage points of GDP" for shares). \\
Category labels swapped & Values attached to the wrong label & Two axis labels are read in the wrong order, so every series is wrong for both categories. \\
Value copied from a neighbour & Wrong numbers & One point repeats the value of the point next to it instead of its own, as when a data label is attached to the wrong bar. \\
Values 8--15\% off & Values & Some values are clearly wrong, 8 to 15\% away from the truth. \\
Values swapped between categories & Values attached to the wrong label & Two neighbouring categories trade values in the first series (Q2 gets Q3's bar and the reverse). \\
Values swapped between series & Values attached to the wrong label & For some categories the values of two series trade places, as when bar colors are confused for one group. \\
All values one gridline too high & Wrong numbers & Every value is read one gridline above its true height, a systematic misreading of the axis. \\
Invented series & Missing or extra data & An extra series that is not in the chart is added with plausible values. \\
Legend lost & Structure & Legend names are replaced by Series 1, Series 2 and so on. The values remain correct but lose their attribution. The policy treats this as critical for multiple series and major for a single series. \\
Invented category & Missing or invented data & A category that is not on the axis is added with plausible values (an extra quarter, year or vendor). \\
\addlinespace[2pt]
\RCband{3}{MAJOR ERRORS} \\*
Series missing & Missing or invented data & The last data series in the legend is left out entirely (a trendline is not a data series: see Trendline left out). Only applies to charts with several series. \\
Category missing & Missing or invented data & One whole category is left out of the extraction. \\
Points missing & Missing or extra data & Some cells of the data series are left empty (a trendline is not a data series). \\
Category labels truncated & Labels & Long axis labels are cut short (Q1 2023 $\to$ Q1, Transsion $\to$ Trans, 2016 $\to$ 16), so some categories become ambiguous. \\
Series repeated & Missing or invented data & The first series appears twice as two identical columns. \\
Title dropped & Labels & The chart title and unit note are left out. The numbers are intact, but nothing says what they measure. \\
Wrong title & Labels & The chart title is replaced by a different one (Quarterly revenue by region $\to$ Quarterly profit by region), so the table claims to show another measure. \\
Row repeated & Missing or extra data & One category row appears twice, a common looping failure. \\
Total row added & Missing or invented data & A Total row that the chart does not show is appended, with values computed by summing each series. \\
\addlinespace[2pt]
\RCband{3}{MINOR ERRORS} \\*
Ranges instead of values & Wrong numbers & Each value is written as a small range around the true number (412.5 $\to$ 410--415), as a cautious reader of an unlabeled chart would. \\
Category axis mislabeled & Labels & The category column is headed with the wrong axis name (Quarter $\to$ Month, Year $\to$ Age group, Vendor $\to$ Region). \\
Series names abbreviated & Labels & Series names are shortened (North America $\to$ N. America, Asia-Pacific $\to$ APAC, Women $\to$ F). Recognizable, but not what the legend says. \\
\addlinespace[2pt]
\RCband{3}{DEBATABLE ERROR TYPES (listed, not graded)} \\*
Category labels reworded & Labels & Axis labels are written differently: Q1 2023 becomes 1Q23, 2016 becomes '16, Others becomes Other. \\
Labels misspelled & Labels & Series and category names carry a small typo (Europe $\to$ Eurpoe, Samsung $\to$ Samsnug). The data is right and recognisable. \\
Rounded to whole numbers & Values & Values are rounded to integers. Close on large numbers, several percent off on small ones. \\
Small reading noise & Values & Values receive small errors like those introduced when estimating from an axis. Such errors can be acceptable without printed data labels, but they are transcription errors when labels are present. \\
Values snapped to gridlines & Wrong numbers & Every value is rounded to the nearest gridline, as a lazy reader of an unlabeled chart would (412.5 $\to$ 400). Wrong on a labeled chart, borderline on an unlabeled one. \\
\addlinespace[2pt]
\RCband{3}{BENIGN VARIATIONS} \\*
Bullet list instead of a table & Output format & One bullet per category listing each series value. Complete and correct, but not a table. \\
Categories reversed & Structure & Rows are listed from last to first. Labels stay attached. \\
Decimal commas & Output format & Values use a decimal comma (412,5) as in many European locales. Same numbers. \\
HTML table & Output format & The same table emitted as HTML instead of Markdown. \\
Hedged value & Output format & One value per series carries a hedge, as a parser writes a reading it is unsure of: approximately 438.2, \~{}305.1, 224.8 (est.). Same numbers. \\
JSON instead of a table & Output format & The data is emitted as a JSON code block. Complete and correct, but not a table. \\
Negatives in parentheses & Output format & Negative values are written in accounting style, (3.2) instead of -3.2. Only applies to charts with negative values. \\
One table per series & Output format & Each series gets its own two-column table under a heading with the series name. \\
Rescaled, unit in header & Scale and units & Values are converted to another unit, which is stated in the column headers. For example, 412.5 million becomes 0.4125 under "North America (USD bn)". The quantities are unchanged. \\
Series in another order & Structure & Columns are listed in a different order than the legend. Labels stay attached. \\
Table transposed & Structure & Series become rows and categories become columns. Every value keeps both labels. \\
Trendline left out & Missing or invented data & The chart draws an Excel trendline over its bars (legend entry Linear (LDCs)) and the parser leaves it out. The trendline is computed from the data, so nothing a reader needs is lost. Only applies to charts with a trendline. \\
Units written in cells & Scale and units & Each cell carries its unit, like \$412.5M or 5.1\%. Same numbers, more verbose. \\
Value and share in one cell & Output format & Each cell includes its value and its share of the category total, such as 412.5 (46.6\%). All chart values remain present. This applies only to charts with multiple series whose values are not percentages. \\
\addlinespace[2pt]
\end{longtable}}

{\RClongsetup\scriptsize\setlength{\tabcolsep}{4pt}\renewcommand{\arraystretch}{1.05}
\begin{longtable}{@{}>{\raggedright\arraybackslash}p{4.3cm}>{\raggedright\arraybackslash}p{2.7cm}>{\raggedright\arraybackslash}p{8.4cm}@{}}
\caption{Severity policy for Forms: 37 error types (13 critical, 13 major, 8 minor, 3 debatable) and 34 benign variations.}\label{tab:policy-forms}\\
\toprule
\RChead{Perturbation} & \RChead{Category} & \RChead{Description} \\
\midrule
\endfirsthead
\multicolumn{3}{l}{\scriptsize\sffamily\color{LIMuted}Severity policy for Forms, continued}\\
\toprule
\RChead{Perturbation} & \RChead{Category} & \RChead{Description} \\
\midrule
\endhead
\bottomrule
\endlastfoot
\RCband{3}{CRITICAL ERRORS} \\*
Decimal point lost in an amount & Values & 52,287.47 is written as 5228747. The digits are unchanged, but the amount is one hundred times larger. \\
Digit misread in a value & Values & One digit changes in some numeric values (amount, date, id): the form now states a different number. \\
Two digits of an ID transposed & Values & An identifier with two neighbouring digits swapped (123-45-6789 $\to$ 123-54-6789): every digit is there, the number is wrong. \\
Values of two fields swapped & Values & Two filled fields exchange their values: each value is attached to the wrong caption. \\
Repeated block's values swapped & Values & Two blocks with the same captions (Applicant / Co-applicant, State line 1 / 2) exchange their values: each value is written under the wrong person or line. \\
Values shifted to the next field & Values & A run of values slides down by one field (a column read out of step): every value in the run sits under the wrong caption. \\
Selected option changed & Checkboxes and signatures & In a single-choice group the tick moves to another option: Married filing jointly instead of Single. \\
Text value misread as another word & Values & A text value is replaced by a different word: Dayton becomes Daytona Beach County. \\
Checkbox state flipped & Checkboxes and signatures & Standalone or multiple-choice checkboxes have their states reversed, so an unchecked box appears checked or vice versa. \\
Every option of a choice ticked & Checkboxes and signatures & Every option in a single-choice group is marked as selected. The intended answer can no longer be identified, as in the over-selection error reported by CheckboxQA. \\
Table column slipped a row & Values & In a form table one column's values sit one row below their own: each amount belongs to the wrong line. \\
Value invented in a blank field & Values & Blank fields contain plausible values that do not appear on the page. \\
Signature claimed on an unsigned line & Checkboxes and signatures & An unsigned line is reported as signed. \\
\addlinespace[2pt]
\RCband{3}{MAJOR ERRORS} \\*
All values dropped (blank template) & Values & The parser returns the blank form: every caption, no filled value. \\
Chosen option unticked (no choice shown) & Checkboxes and signatures & Every checkbox in a single-choice group is marked as empty. The selected answer is lost without selecting an alternative. \\
Filled field missing & Values & Filled fields are left out entirely, caption and value. \\
Ticked option missing & Checkboxes and signatures & A ticked option is left out of the output. \\
Value dropped (caption kept) & Values & Filled fields retain their captions but lose their values, making them appear blank. \\
Caption repeated as the value & Values & A filled field's value is replaced by its caption, such as Employer's name: Employer's name. \\
Signature lost & Checkboxes and signatures & A signed line is reported as blank. \\
Ticks dropped (option text kept) & Checkboxes and signatures & Checked options lose their box glyph: the option is listed but nothing says it was ticked. \\
Value truncated & Values & A multi-word value is cut after its first word: an address loses its city. \\
Values without their captions & Values & Values are written on their own lines, captions dropped: the value is there but no longer says what it is. \\
Amount rounded to whole units & Values & 52,287.47 written 52,287: the cents are gone, the amount is nearly right. \\
Choice group written without boxes & Checkboxes and signatures & A select group's options are written as plain text, no glyphs: which one was chosen is lost. \\
Value marked [illegible] & Values & A filled value replaced by an [illegible] marker: the parser abstains instead of guessing (KIE-HVQA rewards this over a hallucinated value). \\
\addlinespace[2pt]
\RCband{3}{MINOR ERRORS} \\*
Typo in a text value & Values & One letter in a text value changes, such as Sorensen $\to$ Soransen. Digits remain unchanged. \\
Form table flattened to lines & Captions and structure & A form table is written as one line per row (Row 1: Description \ldots{} Amount \ldots{}): values stay in their row but the grid is gone. \\
Blank fields omitted & Values & Empty fields are left out of the output: nothing false, but the form loses fields. \\
Extra text after a value & Values & A value runs into neighbouring printed text: 60082 (see attached). \\
Unticked boxes omitted & Checkboxes and signatures & Standalone and multi-choice boxes that are not ticked are left out. \\
Caption garbled & Captions and structure & Two words in a caption are misread. The value remains correct. \\
Empty boxes dropped (option text kept) & Checkboxes and signatures & Unchecked options retain their text but lose the empty checkbox symbol. They still appear unselected. \\
Form table rows in another order & Captions and structure & Form-table rows appear in reverse order, with each row intact. DocILE's line-item matching ignores row order, while ParseBench's table metric checks it. \\
\addlinespace[2pt]
\RCband{3}{DEBATABLE ERROR TYPES (listed, not graded)} \\*
Captions, then values, in two runs & Debatable & A two-column block read row by row: two captions, then their two values. The pairing is only implied by order. \\
Printed hint read into a blank field & Debatable & The instruction printed in an empty box (Select from the drop-down list) is transcribed as its value. \\
Second half of the page missing & Debatable & The output stops halfway down the page. \\
\addlinespace[2pt]
\RCband{3}{BENIGN VARIATIONS} \\*
Amounts with currency and cents & Format & 52,287.47 written \$52,287.47 and 300,000 written 300,000.00 (VRDU's numeric price match). \\
Ballot-box glyphs ($\boxtimes$ $\square$) & Format & Unicode boxes instead of [x] / [ ]. \\
Blank fields show their format hint & Format & An empty date box written MM/DD/YYYY, other empty boxes as a dotted line. \\
Blank fields written N/A & Format & Empty fields as N/A. \\
Blank fields written as \_\_\_\_ & Format & Empty fields as an underscore line. \\
Bold key: value paragraphs & Format & **Caption**: value lines, options inline after their caption. \\
Box states as Yes / No values & Format & Each option written as a field with a Yes or No value, the way extraction outputs write booleans. \\
Box states as words (Selected / Not selected) & Format & Each option followed by its state in words: Single: Selected, Married: Not selected. \\
Box-per-character field spaced out & Format & A comb field read one box at a time: 919293710 written 9 1 9 2 9 3 7 1 0. \\
Boxes after their option & Format & Yes [x] No [ ] instead of [x] Yes [ ] No. \\
Captions in capitals & Format & Every caption upper-cased. \\
Check marks and Azure markers & Format & $\checkmark$ for ticked, :unselected: for empty boxes. \\
Choice written as the caption's value & Format & Filing status: Single. \\
Dates normalised to ISO & Format & 08/23/2026 and 14/01/2022 written 2026-08-23 and 2022-01-14 (Kleister's canonical form). \\
Dates reformatted & Format & 1997-09-18 written 18/09/1997. \\
Dates with month names & Format & Dates use month names: 08/23/2026 becomes August 23, 2026, and 1997-09-18 becomes 18 September 1997. VRDU accepts these through type-aware date matching. \\
HTML table with checkbox inputs & Format & $<$table$>$ rows, boxes as $<$input type=checkbox checked$>$. \\
IDs and phone numbers re-punctuated & Format & 920-24-6595 written 920 24 6595, (666) 431-5871 written 666.431.5871. \\
JSON nested by section & Format & One JSON object per section, blanks as null (the dictionary output ANLS* scores). \\
JSON-like key/value lines & Format & "Caption": "value", boxes as true / false. \\
Long value wrapped onto the next line & Format & A multi-word value (an address) continues on the line below its caption line. \\
No colon after captions & Format & **Caption** value. \\
Plain Caption: value lines & Format & No markup at all, [X] / [ ] boxes inline. \\
Printed ids dropped & Format & Box and line numbers (1a, 12) left out of the captions. \\
Radio buttons as (X) / ( ) & Format & Boxes written as parenthesised radio marks: (X) Single ( ) Married. \\
Round radio glyphs ($\bullet$ $\circ$) & Format & Filled and empty circles instead of boxes. \\
Section headings left out & Format & Every field kept, the section headings (Part I, Employee) dropped. \\
Sections in another order & Format & Whole sections emitted in a different order (a two-column page read column by column). \\
Single choice: only the chosen option & Format & A single-choice group lists only its ticked option. \\
Thousands separators dropped & Format & 28,898.28 written 28898.28. \\
Two fields on one line & Format & A two-column block written row by row with both fields on one line: Name: Ana Ruiz   Date: 2026-04-27. \\
Two-column table & Format & Each section as a Field \textbar{} Value table. \\
Values in capitals & Format & Every filled value upper-cased, as some OCR engines return block capitals. \\
Values underlined & Format & Filled values wrapped in $<$u$>$\ldots{}$<$/u$>$, as the enriched forms output writes them. \\
\addlinespace[2pt]
\end{longtable}}

{\RClongsetup\scriptsize\setlength{\tabcolsep}{4pt}\renewcommand{\arraystretch}{1.05}
\begin{longtable}{@{}>{\raggedright\arraybackslash}p{4.3cm}>{\raggedright\arraybackslash}p{2.7cm}>{\raggedright\arraybackslash}p{8.4cm}@{}}
\caption{Severity policy for Visual grounding: 24 error types (7 critical, 7 major, 9 minor, 1 debatable) and 14 benign variations.}\label{tab:policy-layout}\\
\toprule
\RChead{Perturbation} & \RChead{Category} & \RChead{Description} \\
\midrule
\endfirsthead
\multicolumn{3}{l}{\scriptsize\sffamily\color{LIMuted}Severity policy for Visual grounding, continued}\\
\toprule
\RChead{Perturbation} & \RChead{Category} & \RChead{Description} \\
\midrule
\endhead
\bottomrule
\endlastfoot
\RCband{3}{CRITICAL ERRORS} \\*
One page-wide box &  & One Text box covers the entire page and contains all its text. \\
Biggest paragraph missed &  & Paragraph 5 is missing. All other elements are correct. \\
Five paragraphs in one box &  & One Text box contains all five paragraphs and also covers the picture and table. \\
Right box, wrong text &  & Paragraph 3 has the correct box but contains paragraph 4's text. \\
Coarse blocks &  & Two Text blocks cover large page regions: one from the title through paragraph 3, and one from the icon through paragraph 5. \\
Japanese paragraph garbled &  & Paragraph 4 has the correct box, but about half its characters are wrong or missing. \\
Loose boxes (IoU about 0.45) &  & Every box expands to 1.49 times its original width and height. \\
\addlinespace[2pt]
\RCband{3}{MAJOR ERRORS} \\*
Every box six times &  & Each annotated box is predicted once for every basic class. The correctly labeled copy appears first. \\
50 junk boxes &  & The correct output also includes 50 randomly placed boxes with random classes and no text. \\
Table box cuts the header row &  & The table box starts below its header row. All other content is correct. \\
Table box cuts the last column &  & The table box excludes its last column. All other content is correct. \\
Header merged into a paragraph &  & A section header and paragraph 1 share one Text box. \\
Two columns, one box over both &  & One box encloses both column segments of paragraph 5 and contains its full text. The box also covers the gutter and most of paragraph 6. \\
Loose boxes (IoU about 0.64) &  & Every box expands to 1.25 times its original width and height. \\
\addlinespace[2pt]
\RCband{3}{MINOR ERRORS} \\*
Empty boxes between table rows &  & The correct output also includes six empty Text boxes between table rows. \\
Five boxes on blank margins &  & The correct output also includes five empty boxes in the page margins. \\
Second box over a matched line &  & An extra Section box covers the first line of paragraph 2. Those words are already assigned to paragraph 2's correct box. \\
Box nicks the next paragraph &  & Paragraph 2's box extends over the first line of paragraph 3. All text and other boxes are correct. \\
Header and footer as page-wide bands &  & The page header and page number use full-width boxes containing the correct text. \\
Text boxes inside a picture &  & The correct output also includes four Text boxes inside the main picture. \\
Hyphen dropped, word left split &  & The header is output as "Quarterly infor mation report": the hyphen is lost and the word stays in two pieces. \\
Japanese paragraph, one character wrong &  & Paragraph 4 is correct except for one character (the character for three read as the character for two). \\
Tiny icon missed &  & The 18$\times$18 pt icon is missing. All other elements are correct. \\
\addlinespace[2pt]
\RCband{3}{DEBATABLE ERROR TYPES (listed, not graded)} \\*
Split at the wrong place &  & Paragraphs 1 and 2 are split into two boxes at the wrong boundary. The first contains lines 1--3 of paragraph 1, and the second contains its last line and all of paragraph 2. All text remains in order. Whether this matters depends on the benchmark. \\
\addlinespace[2pt]
\RCband{3}{BENIGN VARIATIONS} \\*
Boxes 1.5 pt off &  & Every box shifts 1.5 pt to the right, giving an IoU of about 0.95. \\
Boxes on optional sub-images &  & The correct output also includes Picture boxes around optional sub-images within the main picture. \\
Every class wrong &  & All boxes and text are correct, but every class label is wrong. Classes are reported separately. \\
Hyphenated word joined &  & The header is output as "Quarterly information report", the word rejoined across the line break. \\
Hyphenated word kept as printed &  & The two-line header is output as "Quarterly infor- mation report". The benchmark joins "infor- mation" on both sides. \\
Japanese paragraph, read right &  & Paragraph 4 is Japanese. Its box and text are correct, with each CJK character treated as one token. \\
Paragraph split into lines &  & The seven lines of paragraph 5 each have a separate box containing that line's text. \\
Perfect &  & Every element has a tight box, the correct class and the correct text. \\
Annotations by line, output by paragraph &  & Paragraph 5 has seven annotated line boxes. The output has one tight paragraph box. \\
Table labelled Picture &  & All content is correct, but the table box is labeled Picture. \\
Table plus tight cell boxes &  & The correct output also includes a tight Text box around each table word. \\
Two columns, one box per part &  & Paragraph 5 continues from the left column into the right. Each part has its own box and text. \\
Two columns, one element in two segments &  & Paragraph 5 spans two columns. The output represents it as one element with two box segments and the full paragraph text. \\
Two paragraphs in one box &  & Paragraphs 1 and 2 share one Text box containing both paragraphs. \\
\addlinespace[2pt]
\end{longtable}}

\clearpage
\subsection{Metric Definitions by Dimension}
\label{app:metrics}

This appendix explains how we score each parsing dimension (\Cref{sec:metrics}).
We describe how each metric matches outputs to annotations, measures content recovery and detects errors.
Text, Formatting, Tables, Charts and Forms combine a parsing score $m$ with a severity cap $c$ (\Cref{app:metric-cap}).
Visual grounding scores the location of parsed content directly, without a cap.

\subsubsection{Text}
\label{sec:text}

The Text metric checks whether the output preserves words, numbers and identifiers, keeps sentences intact, and follows the page's reading order.
It distinguishes moved text from missing text and a wrong number from a typo.
These distinctions can be lost in existing metrics.
On a 10-K page (a US annual report), \pbone{} Content Faithfulness deducts half a point for either a misspelled word or a wrong revenue share because it weights every rule equally.
On another example, $1-\text{NED}$ scores a moved sidebar at 63.8 and a deleted one at 81.9, penalizing the move more than the omission.

\paragraph{Matching text.}
We convert the output and annotations to plain text, removing supported Markdown, HTML and emphasis syntax.
We track numbers, identifiers, units, currency signs and negators as \emph{key tokens} because changing them can change a fact.
Greedy string tiling~\citep{wise1993gst} matches maximal runs of shared tokens, including blocks that have moved.
We then align the remaining tokens between these runs using a longest common subsequence (\Cref{fig:text-align}).
Moved text retains credit for its content but can lose points for reading order and a \emph{moved block} error.

\def\TxNref{107}
\def\TxMovedNed{63.8}
\def\TxMovedNext{95.0}
\def\TxMovedOrder{0.82}
\def\TxMovedProp{98.2}
\def\TxDroppedNed{81.9}
\def\TxDroppedNext{75.7}
\def\TxTenkNref{870}
\def\TxTenkCFself{0.961}

\begin{figure}[!htbp]
\centering
\begin{minipage}[b]{0.8\textwidth}
\centering
\begin{tikzpicture}[x=0.074cm, y=1cm, font=\scriptsize\sffamily]
  \def\blk#1#2#3#4#5{%
    \fill[#5!22] (#1,#3) rectangle ++(#2,0.42);
    \draw[#5!85!black, line width=0.6pt] (#1,#3) rectangle ++(#2,0.42);
    \foreach \t in {1,...,#2} {\draw[#5!60, line width=0.2pt] (#1+\t,#3) -- ++(0,0.09);}
    \node[#5!70!black, font=\scriptsize\sffamily\bfseries] at (#1+#2/2,#3+0.24) {#4};}
  \def\tile#1#2#3#4{%
    \fill[#4!12] (#1,1.6) -- (#1+#3,1.6) -- (#2+#3,0.42) -- (#2,0.42) -- cycle;}
  \tile{0}{0}{10}{pbNext}
  \tile{11}{31}{24}{pbNext}
  \tile{36}{56}{20}{pbNext}
  \tile{57}{11}{19}{sevMaj}
  \tile{77}{77}{19}{pbNext}
  \tile{97}{97}{15}{pbNext}
  \blk{0}{10}{1.6}{H}{pbNext}
  \blk{11}{24}{1.6}{P1}{pbNext}
  \blk{36}{20}{1.6}{P2}{pbNext}
  \blk{57}{19}{1.6}{P3}{sevMaj}
  \blk{77}{19}{1.6}{P4}{pbNext}
  \blk{97}{15}{1.6}{P5}{pbNext}
  \blk{0}{10}{0}{H}{pbNext}
  \blk{11}{19}{0}{P3}{sevMaj}
  \blk{31}{24}{0}{P1}{pbNext}
  \blk{56}{20}{0}{P2}{pbNext}
  \blk{77}{19}{0}{P4}{pbNext}
  \blk{97}{15}{0}{P5}{pbNext}
  \node[anchor=east, font=\scriptsize\bfseries] at (-1,1.81) {Annotations};
  \node[anchor=east, font=\scriptsize\bfseries] at (-1,0.21) {Out.};
  \draw[decorate, decoration={brace, mirror, amplitude=3pt}, black!60] (0,-0.08) -- (10,-0.08);
  \draw[decorate, decoration={brace, mirror, amplitude=3pt}, black!60] (31,-0.08) -- (112,-0.08);
  \node[anchor=north, black!75, align=center] at (71.5,-0.22) {heaviest in-order chain: H P1 P2 P4 P5 (88 tokens)};
  \draw[decorate, decoration={brace, mirror, amplitude=3pt}, sevMaj!90!black] (11,-0.08) -- (30,-0.08);
  \node[anchor=north, sevMaj!80!black, align=center] at (17,-0.22) {moved tile\\(19 tokens)};
  \node[anchor=south, align=center, font=\scriptsize\sffamily] at (56,2.08)
    {\textcolor{pbLegacy!80!black}{\textbf{1$-$NED = \TxMovedNed}}: delete P3 here, insert it there\quad
     \textcolor{pbNext}{\textbf{\textnext{} = \TxMovedNext}}};
  \node[anchor=north, align=center, font=\scriptsize\sffamily] at (56,-0.95)
    {words $=$ key $=$ 1.00;\quad order component $=1-19/\TxNref=\TxMovedOrder$\\
     one major \emph{moved block}: $c=95$,\ \ $S=\min(\TxMovedProp,\,95)=\TxMovedNext$};
\end{tikzpicture}
\end{minipage}

\caption{\textbf{Matching text after a block moves.} On this 107-token annual-report page, sidebar P3 appears first in the output. Blue bands retain the annotated order, and the orange band marks the moved block. All words and key tokens still match. The order component falls to 0.82, lowering the parsing score to 98.2. The move also produces one major error, so the cap sets the final score at 95. By comparison, $1-$NED deducts 36 points for the same move. H is the running header, and P1--P5 are paragraphs.}
\label{fig:text-align}
\end{figure}
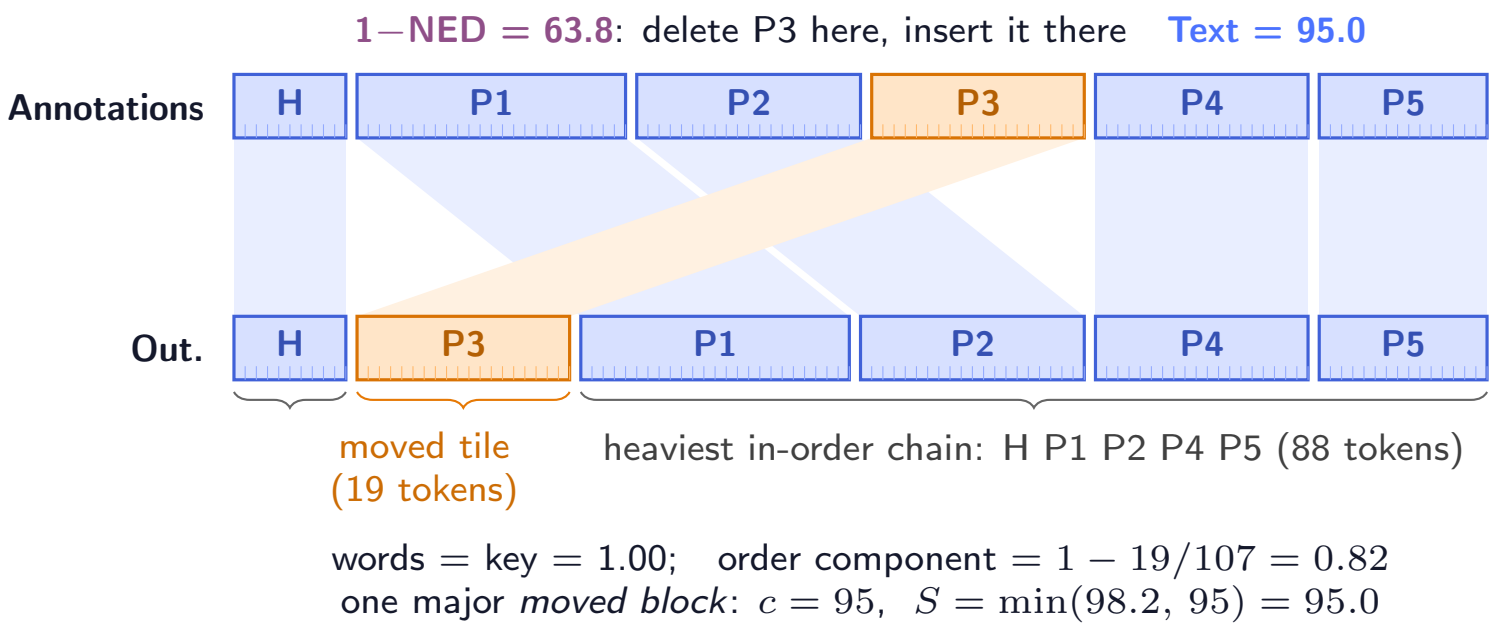

\paragraph{Parsing score.}
Four components measure text recovery.
\emph{Words} is the F1 of matched tokens, and \emph{key} is the F1 of key tokens.
\emph{Integrity} measures the share of sentence words in sentences uninterrupted by text from elsewhere on the page.
\emph{Order} measures how many matched tokens remain in the annotated reading order.
We find the set of blocks with the largest total token count that preserves this order, then divide by all matched tokens.
We subtract the weighted losses of these components from 1:
\begin{equation}
m_{\text{text}} = \max\!\Big(0,\ 1 - \big[\,(1-\text{words}) + 1.5\,\rho\,(1-\text{key}) + 0.25\,(1-\text{integrity}) + 0.1\,(1-\text{order})\big]\Big),
\label{eq:text-prop}
\end{equation}
Here, $\rho = \min(1, n_c/10)$ reduces the influence of key tokens when a page contains few of them.
The count $n_c$ is the larger of the key-token counts in the annotations and the output.
An undefined component contributes a value of 1.
When $\rho=1$, a key-token F1 of 0.9 lowers the parsing score by 15 points.
Order has the smallest weight because moved text remains available to the reader.

\paragraph{Error severity.}
We classify errors from unmatched token runs.
Wrong numbers, changed negators and omissions of at least 12 tokens are critical.
We detect changed negators through lexical rules.
Shorter omissions, wrong words and moved blocks are major.
Spelling and accent changes are minor.
The cap follows \Cref{eq:cap} with $K=400$ tokens.
Moved blocks and interrupted sentences together lower it by at most 30 points, preserving credit when the words are intact but the reading order is wrong.

On a 56-token insurance notice, benign formatting variations score 100.
Swapping two paragraphs produces one major error and a score of 95.0.
Changing ``30 days'' to ``60 days'' produces one critical error and a cap of 90.
The parsing score is lower, at 83.2, so it determines the final score.

\FloatBarrier
\subsubsection{Formatting}
\label{sec:formatting}

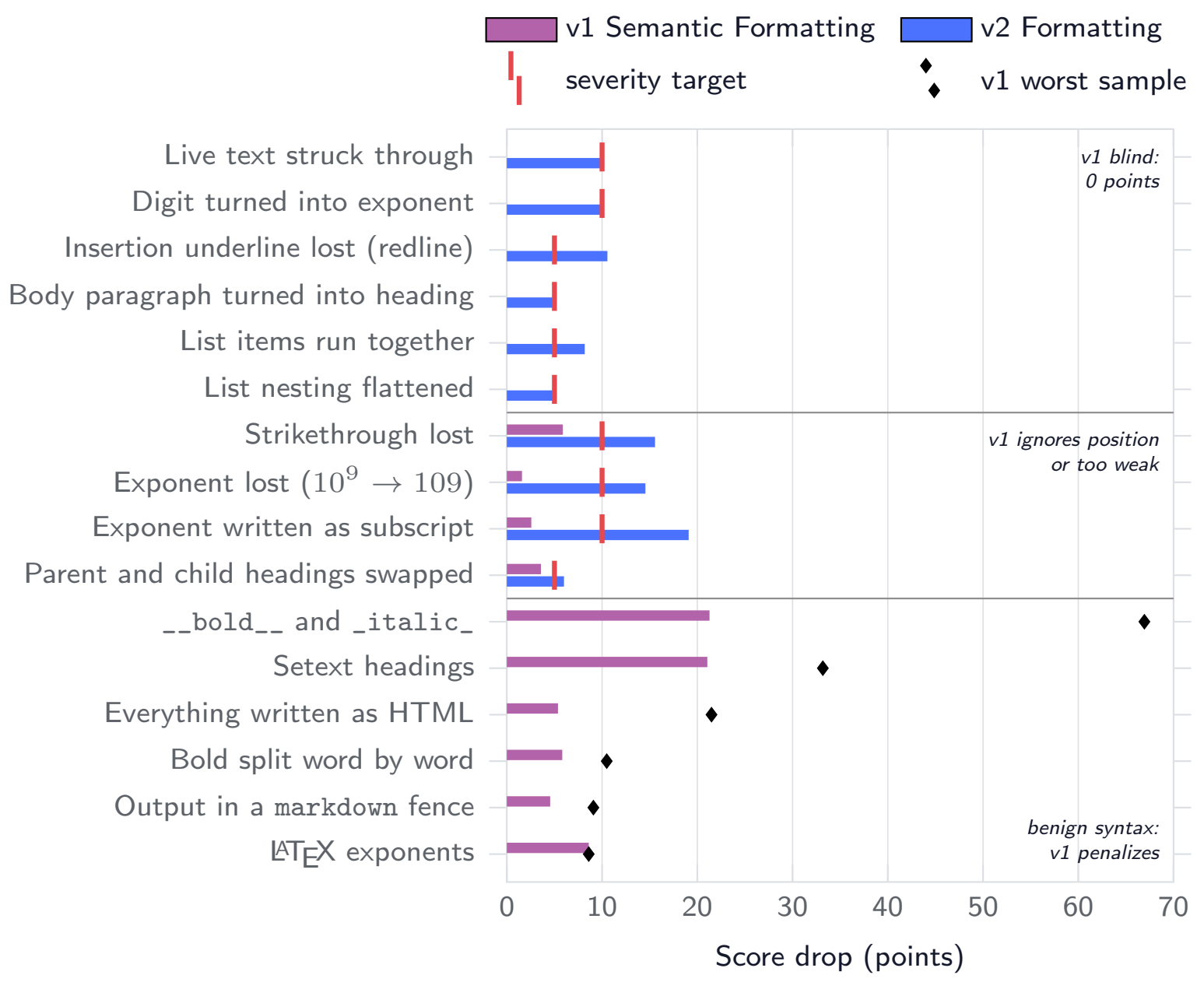
\begin{figure}[!htbp]
\centering
\begin{pbnarrow}[0.66]
\begin{tikzpicture}
\begin{axis}[pbaxis, xbar, scale only axis, width=0.52\columnwidth, height=6.4cm, bar width=2.5pt,
  xmin=0, xmax=70, xtick={0,10,20,30,40,50,60,70}, xmajorgrids,
  ymin=-0.6, ymax=15.6, ytick={0,...,15}, yticklabels from table={format_v1v2.dat}{label},
  y tick label style={font=\scriptsize}, clip=false,
  xlabel={Score drop (points)},
  legend style={at={(0.5,1.02)}, anchor=south, legend columns=2, /tikz/every even column/.append style={column sep=4pt}}]
\addplot[area legend, fill=pbLegacy, draw=none, bar shift=1.5pt] table[x=v1, y=y] {format_v1v2.dat};
\addplot[area legend, fill=pbNext, draw=none, bar shift=-1.5pt] table[x=v2, y=y] {format_v1v2.dat};
\addplot[only marks, mark=|, mark size=3.5pt, very thick, sevCrit, keep if={grp}{blind}] table[x=floor, y=y] {format_v1v2.dat};
\addplot[forget plot, only marks, mark=|, mark size=3.5pt, very thick, sevCrit, keep if={grp}{weak}] table[x=floor, y=y] {format_v1v2.dat};
\addplot[only marks, mark=diamond*, mark size=1.6pt, black, keep if={grp}{harmless}] table[x=v1max, y=y] {format_v1v2.dat};
\legend{v1 Semantic Formatting, v2 \formatnext{}, severity target, v1 worst sample}
\draw[black!45] (axis cs:0,9.5) -- (axis cs:70,9.5) (axis cs:0,5.5) -- (axis cs:70,5.5);
\node[font=\tiny\itshape, anchor=north east, align=right] at (axis cs:70,15.5) {v1 blind:\\0 points};
\node[font=\tiny\itshape, anchor=north east, align=right] at (axis cs:70,9.4) {v1 ignores position\\or too weak};
\node[font=\tiny\itshape, anchor=south east, align=right] at (axis cs:70,-0.5) {benign syntax:\\v1 penalizes};
\end{axis}
\end{tikzpicture}
\end{pbnarrow}
\caption{\textbf{Score drops for formatting errors and benign variations.} Bars show mean drops on the Formatting suite's sample pages (\Cref{sec:lab-protocol}), using the perturbation tool's 50\% strength setting. Top: errors \pbone{} misses. Middle: errors it penalizes too little because it accepts the formatted text anywhere on the page. Bottom: benign syntax variations, which should receive no penalty. Red ticks mark the required minimum drops: 10 points for critical errors and 5 for major errors. A missing bar means zero drop. Diamonds show the largest \pbone{} drop on a single sample, up to \fmtVoneHarmlessMax{} points. \pbtwo{} meets every target shown.}
\label{fig:format-v1v2}
{\footnotesize\par}
\end{figure}

The Formatting metric checks formatting that affects meaning or document structure, including strikethrough, exponents, headings and lists.
\pbone{} Semantic Formatting uses one rule per annotated span and accepts only one markup syntax for each rule.
It also leaves invented formatting unpenalized.
On a page of bold labels, writing \texttt{\_\_bold\_\_} instead of \texttt{**bold**} can lose up to \fmtVoneHarmlessMax{} points (\Cref{fig:format-v1v2}).
We score the 480 core text pages with at least one formatting item.

\paragraph{Matching formatting.}
We map supported syntaxes to a common representation.
Each line has a role, such as a paragraph, a heading with a level, or a list item with a nesting depth.
Each character records its styles, including bold, italic, underline, strikethrough, superscript, subscript, highlight and inline code.
Thus \texttt{**}, \texttt{\_\_} and \texttt{<b>} all represent bold text.
ATX, setext and \texttt{<hN>} syntax all represent headings.

We align words using the Text metric's procedure, including across moved blocks.
A \emph{scored item} is a style span, scored superscript or subscript token, heading, list item or code block.
We count items in the annotations and items added to aligned output text.
For headings, we compare the relative levels of consecutive headings.
A uniform shift of all heading levels therefore receives no penalty.

\paragraph{Parsing score.}
A lost, added or changed item is one error with a loss of 1.0 (critical), 0.5 (major) or 0.2 (minor).
We sum these losses and divide by the number of scored items plus 3:
\begin{equation}
  m_{\text{fmt}} = 1 - \frac{\sum_a \text{loss}_a}{n_{\text{items}} + 3}.
  \label{eq:format-prop}
\end{equation}
Here, $\text{loss}_a$ is the loss for item $a$, and correct items contribute zero.
Adding 3 to the denominator limits the effect of one missing item on a page with little formatting.
The severity cap still enforces a substantial penalty for a serious error on such a page.
The ratio of critical to minor item losses is 5:1, compared with 10:1 for cap penalties.
This gives repeated minor errors more influence on content recovery.

\paragraph{Error severity.}
Lost or added strikethrough is critical because it changes whether text appears to be deleted.
A changed exponent, such as $10^9$ read as 109, is also critical.
Lost underline is major on redline pages, where it marks inserted text.
Other emphasis errors are minor.
The cap follows \Cref{eq:cap} with $K=10$ items, without scaling minor errors.

On a short contract amendment, equivalent markup and a uniform heading shift score 100.
A missing bold label loses 1.7 points.
A struck fee rendered as plain text caps the score at 90.

\paragraph{Targeted strikethrough and underline collections.}
The general item score gives too much credit to outputs that omit all strikethrough or underline.
Such outputs would score 52 on the strikethrough collection and 93 on the underline collection.
We therefore annotate every decorated span in these two targeted collections (\Cref{sec:features}).
We call these annotations \emph{decoration inventories}.
For these pages, the parsing score is the F1 of decorated words in the aligned text.
It penalizes both missing decoration and decoration added elsewhere on the page.
The cap is unchanged.
In simulations that remove all decoration from the 253 pages, scores fall to 14 for strikethrough and 20 for underline.

\FloatBarrier
\subsubsection{Tables}
\label{sec:tables}

The Tables metric checks cell values, their row and column labels, and the order of tables and rows.
\pbone{} scored tables with GTRM, the mean of GriTS~\citep{smock2023gritsgridtablesimilarity} and \textsc{TableRecordMatch}.
In our examples, GTRM scores a transposed table at 13.2, below a table with half its rows missing.
Yet ten wrong numbers in a 459-cell table still score 98.7.

\begin{figure}[!htbp]
\centering
\begin{tikzpicture}[font=\footnotesize\sffamily,
  c/.style={draw=black!35, minimum height=0.5cm, inner sep=2pt, anchor=center, text depth=0.25ex},
  h/.style={c, fill=pbRef, font=\footnotesize\sffamily\bfseries}]
  \def\H{0.5}
  \node[h, minimum width=2.4cm, minimum height=1.0cm] (region) at (1.2,-0.25) {Region};
  \node[h, minimum width=2.4cm, fill=pbLegacy!25] (rev) at (3.6,0) {Revenue (USD m)};
  \node[h, minimum width=2.4cm] at (6.0,0) {Share (\%)};
  \node[h, minimum width=1.2cm] at (3.0,-\H) {2023};
  \node[h, minimum width=1.2cm, fill=pbLegacy!25] (y24) at (4.2,-\H) {2024};
  \node[h, minimum width=1.2cm] at (5.4,-\H) {2023};
  \node[h, minimum width=1.2cm] at (6.6,-\H) {2024};
  \node[c, minimum width=2.4cm, fill=pbNext!20] (lab) at (1.2,-2*\H) {Europe};
  \node[c, minimum width=1.2cm] at (3.0,-2*\H) {846};
  \node[c, minimum width=1.2cm, fill=pbHi, font=\footnotesize\sffamily\bfseries] (val) at (4.2,-2*\H) {901};
  \node[c, minimum width=1.2cm] at (5.4,-2*\H) {30.4};
  \node[c, minimum width=1.2cm] at (6.6,-2*\H) {29.3};
  \node[draw=pbNext!60, fill=pbNext!20, rounded corners=2pt, inner sep=4pt] (t1) at (1.2,-4.4*\H) {Europe};
  \node[draw=pbLegacy!60, fill=pbLegacy!25, rounded corners=2pt, inner sep=4pt] (t2) at (4.0,-4.4*\H) {Revenue (USD m) / 2024};
  \node[draw=pbHi!80!black, fill=pbHi, rounded corners=2pt, inner sep=4pt, font=\footnotesize\sffamily\bfseries] (t3) at (7.0,-4.4*\H) {901};
  \node[anchor=east] at (t1.west) {$\big($};
  \node at ($(t1.east)!0.5!(t2.west)$) {,};
  \node at ($(t2.east)!0.5!(t3.west)$) {,};
  \node[anchor=west] at (t3.east) {$\big)$};
  \node[anchor=north, font=\scriptsize\sffamily, black!60] at (t1.south) {row-label path};
  \node[anchor=north, font=\scriptsize\sffamily, black!60] at (t2.south) {column-header path};
  \node[anchor=north, font=\scriptsize\sffamily, black!60] at (t3.south) {value};
  \draw[pbarr, pbNext] (lab.south) -- (t1.north);
  \draw[pbarr, black!60] (val.south) -- ++(0,-0.35) -| (t3.north);
\end{tikzpicture}
\caption{\textbf{Scoring a value together with its headers.} The value 901 belongs to Europe (blue) and Revenue (USD m) / 2024 (pink). The metric represents these labels and the value as a triple. Equivalent Markdown or HTML, \texttt{th} or \texttt{td} tags, and spanning or repeated labels yield the same triples. Placing the value under a different year changes the triple and the fact it represents.}
\label{fig:table-triples}
\end{figure}
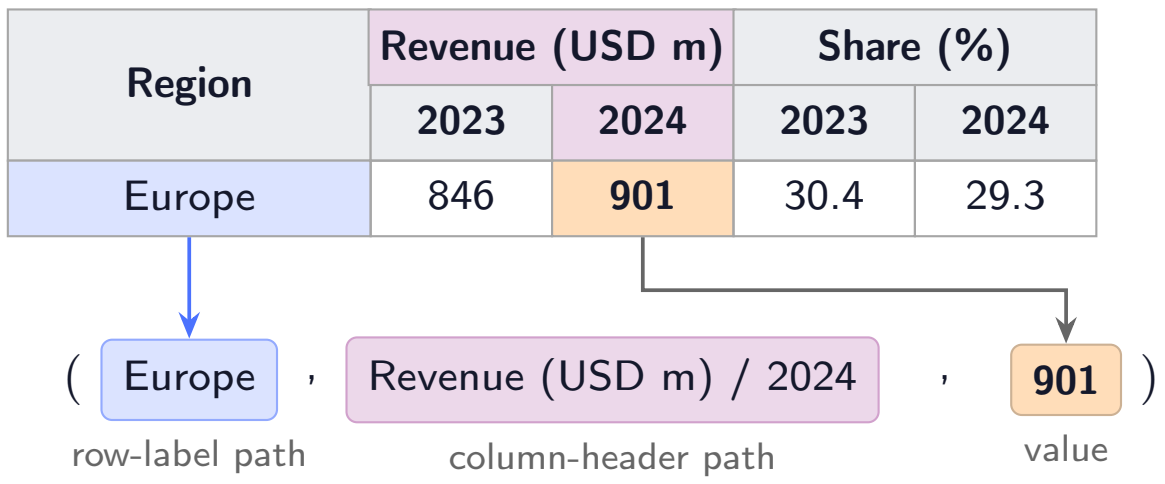

\paragraph{Representing table content.}
A value is meaningful only in the context of its headers.
In \Cref{fig:table-triples}, 901 is Europe's revenue for 2024.
Placing it under 2023 changes the fact.
We represent each data cell as a triple containing its full row-label path, full column-header path and value:
$(r_1/\cdots/r_a,\allowbreak\ c_1/\cdots/c_b,\allowbreak\ v)$.
The table is the multiset of these triples.

We convert HTML and Markdown pipe tables to grids and expand spanning cells.
The choice of \texttt{<th>} or \texttt{<td>} does not affect the score.
A dash in place of a blank cell is a benign variation, whether it appears in the body, a label, a corner or a section row.
A dash that replaces a nonempty value remains an error.

TEDS~\citep{zhong2020imagebasedtablerecognitiondata} compares HTML trees, while GriTS aligns cells on a 2D grid.
Both average differences across cells, so one wrong value can have little effect.
They can also assign a small penalty to a value under the wrong header when the grid retains its shape.
\textsc{TableRecordMatch} matches rows as records but ignores row order and cannot represent hierarchical row labels.
Our metric checks each value together with its headers.
The cap ensures that one wrong digit lowers an otherwise correct table's score by at least 10 points.
\Cref{tab:suite-results} compares it with these metrics on the Tables test suite.

\paragraph{Matching tables and cells.}
We pair tables using shared row and column labels.
For each annotated table, we infer the title rows, header rows and label columns, then apply that structure to the paired output.
We also try the output's own inferred structure, a transposed reading, and a reading that treats dash-only cells as blanks.
We keep the reading that best matches the annotations.
This preserves content credit for a transposed table, while the cap penalizes the change in orientation.
These alternatives cover common representations but not every possible format.

Within each pair of tables, we match triples one to one with a Hungarian assignment~\citep{kuhn1955hungarian}, allowing matches only between compatible labels.
We ignore output tables that share no keys with the annotations because some tables on a page may be unannotated.
If an annotated table has fewer than three values, we count empty markers such as dashes and ``n/a'' as values.
This allows us to score tables made mostly of such markers.

\paragraph{Parsing score.}
Each matched pair of triples receives a weight equal to its label similarity multiplied by its value similarity.
The value similarity is 1 for equal values, 0.95 when only letter case differs, 0.9 for the same number written differently, 0.5 when a unit is lost and 0 when any digit differs.
We use these weights to compute precision and recall, then subtract a row-order penalty from their F1:
\begin{equation}
\begin{split}
  \text{content} &= \mathrm{F1}\Big(\tfrac{\sum w}{|G|},\ \tfrac{\sum w}{|O|}\Big),\\
  m_{\text{table}} &= \max\big(0,\ \text{content} - 0.3\,(1 - \text{row\_order})\big),
\end{split}
  \label{eq:table-prop}
\end{equation}
Here, $G$ and $O$ are the annotated and output triples in the paired tables, and $\mathrm{F1}(x,y) = 2xy/(x+y)$.
The term row\_order is the share of annotated rows in the longest run that retains the annotated order.
Missing and invented cells both reduce the score.
An output with no table scores 0.

\paragraph{Error severity.}
A changed digit or a value under the wrong header is critical.
A changed number in a header, such as 2025 read as 2024 or 202.5, is also critical.
So are swapped or shifted headers, which change the meaning of the values below them.
We penalize a reworded label once per table.
It is minor if it differs by one typo and major otherwise.

A transposition that preserves every header--value association is one major error under our policy for consumers that read tables row by row.
Any lost, changed or misplaced cells can still produce critical errors.
The cap follows \Cref{eq:cap} with $K=80$ annotated cells.
Three or more cell errors at the same severity level within one column count as one column error.
We also group errors from an inserted column, a change of decimal convention across a whole table (minor), or a table moved in page order.
On a small revenue table, one changed digit gives a content F1 of 91.7 and a cap of 90.
The cap determines the final score.

\FloatBarrier
\subsubsection{Charts}
\label{sec:charts}

The Charts metric checks every annotated data point, including its category, series and value.
We score figures that plot data values.
Diagrams are not yet scored (\Cref{sec:conclusion}).
\pbone{}'s \textsc{ChartDataPointMatch} checked about ten selected values per chart.
Its 4{,}864 rules covered 13\% of the plotted values.
Some rules also expected the value of a neighboring bar.
Checking selected values leaves other errors untested, such as an invented series, a wrong unit or a missing title.

\paragraph{Matching chart data.}
We annotate each chart with a complete data table, title, chart type, axis labels, unit, categories and series (\Cref{sec:dataset-coverage}).
Complete annotations let us score all data points and penalize invented ones.
The scorer accepts tables in either orientation, JSON blocks and bullet lists.
It normalizes labels and converts values using units in cell suffixes, headers and scale notes.
An explicitly stated unit that contradicts the annotations makes the value incorrect.
We match rows and columns to categories and series with a Hungarian assignment.
If an output combines several charts in one table, a row or column bearing a chart's title is assigned only to that chart.

\paragraph{Allowing for reading error.}
Printed data labels can be transcribed directly, while values read from an axis must be estimated.
We therefore use different tolerances for labeled and unlabeled charts.
Let $t$ be the annotated value and $p$ the output value.
For labeled charts, the allowance is $a = 1\%\cdot\max(|t|,|p|)$, unless the annotations specify another relative tolerance.
Using the larger magnitude makes the tolerance symmetric in $t$ and $p$, as in Python's \texttt{math.isclose}.
It gives an overshoot at most 1\% more allowance than an undershoot: 1.01 versus 1.00 for an annotated value of 100.

For unlabeled charts, $a = \max\big(1\%\cdot|t|,\ 2\%\cdot\text{range}\big)$, where range is the range of annotated values.
We chose 2\% as an allowance comparable to reported human reading errors on bar and line charts~\citep{cleveland1984graphical,heer2010crowdsourcing}.
We have not measured human reading error on this benchmark.
Values within the allowance receive full credit.
Beyond it, closeness falls linearly to zero at $3a$:
\begin{equation}
  \nu(t, p) = \begin{cases} 1 & |p-t| \le a\\ \max\!\big(0,\ 1 - \frac{|p-t| - a}{2a}\big) & \text{otherwise,}\end{cases}
  \label{eq:chart-vscore}
\end{equation}
We also set $\nu = 0$ when the signs differ beyond the allowance.

\paragraph{Parsing score.}
Each matched point receives credit for both its value and its labels.
We multiply value closeness $\nu$ by label similarity $\beta$, then compute F1 over the annotated and output points:
\begin{equation}
  R = \frac{\sum_i \beta_i \nu_i}{n_{\text{ref}}},\quad
  Pr = \frac{\sum_i \beta_i \nu_i}{n_{\text{out}}},\quad
  m_{\text{chart}} = \mathrm{F1}(R, Pr),
  \label{eq:chart-prop}
\end{equation}
Here, $\beta_i$ is 1 when the labels match after normalization.
It uses string similarity for a typo and gives less credit for a rewording.
The counts $n_{\text{ref}}$ and $n_{\text{out}}$ are the numbers of annotated and output points in the assigned chart grids.

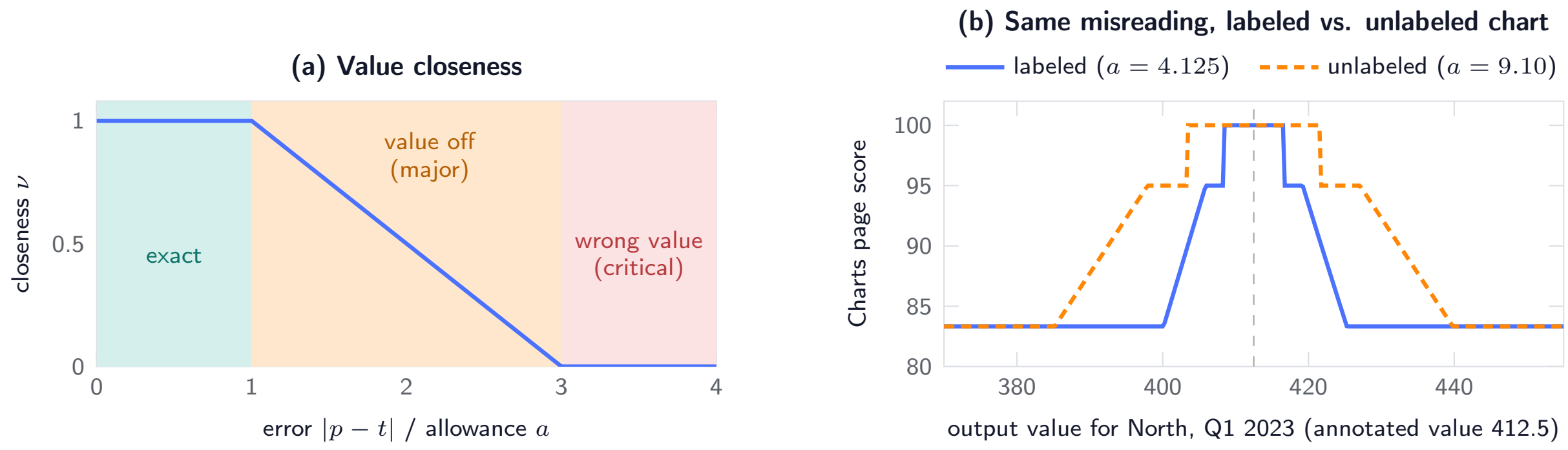
\begin{figure}[!htbp]
\centering
\def\chtolLab{4.125}\def\chtolUnl{9.10}

\begin{minipage}[b]{0.48\textwidth}\centering
\begin{tikzpicture}
\begin{axis}[pbaxis, width=\linewidth, height=4.3cm, xmin=0, xmax=4, ymin=0, ymax=1.08,
  xlabel={error $|p-t|$ / allowance $a$}, ylabel={closeness $\nu$}, ytick={0,0.5,1}, xtick={0,1,2,3,4},
  title={\footnotesize (a) Value closeness}, title style={yshift=-4pt}]
  \fill[sevHarm!18] (axis cs:0,0) rectangle (axis cs:1,1.08);
  \fill[sevMaj!20] (axis cs:1,0) rectangle (axis cs:3,1.08);
  \fill[sevCrit!16] (axis cs:3,0) rectangle (axis cs:4,1.08);
  \addplot[pbNext, very thick] table[x=x, y=nu] {chart_tol_nu.dat};
  \node[font=\scriptsize, sevHarm!70!black] at (axis cs:0.5,0.45) {exact};
  \node[font=\scriptsize, sevMaj!70!black, align=center] at (axis cs:2.15,0.85) {value off\\(major)};
  \node[font=\scriptsize, sevCrit!80!black, align=center] at (axis cs:3.5,0.45) {wrong value\\(critical)};
\end{axis}
\end{tikzpicture}
\end{minipage}\hfill
\begin{minipage}[b]{0.48\textwidth}\centering
\begin{tikzpicture}
\begin{axis}[pbaxis, width=\linewidth, height=4.3cm, xmin=370, xmax=455, ymin=80, ymax=102,
  xlabel={output value for North, Q1 2023 (annotated value 412.5)}, ylabel={\chartnext{} page score}, ytick={80,85,90,95,100},
  legend style={at={(0.5,1.03)}, anchor=south, legend columns=2, /tikz/every even column/.append style={column sep=6pt}},
  title={\footnotesize (b) Same misreading, labeled \vs{} unlabeled chart}, title style={yshift=9pt}]
  \draw[black!35, dashed] (axis cs:412.5,80) -- (axis cs:412.5,102);
  \addplot[pbNext, very thick] table[x=p, y expr=100*\thisrow{score}] {chart_tol_labeled.dat};
  \addlegendentry{labeled ($a=\chtolLab$)}
  \addplot[sevMaj, very thick, densely dashed] table[x=p, y expr=100*\thisrow{score}] {chart_tol_unlabeled.dat};
  \addlegendentry{unlabeled ($a=\chtolUnl$)}
\end{axis}
\end{tikzpicture}
\end{minipage}
\caption{\textbf{Allowing for uncertainty when reading chart values.} (a)~A value within allowance $a$ receives full credit. Closeness falls linearly to zero at $3a$. Errors beyond $a$ are major, becoming critical beyond $3a$. (b)~We vary one of six values in a two-series chart. The dashed line marks the annotated value, 412.5. With printed labels, a value just outside the allowance caps the score at 95. As the error grows, the parsing score $m$ falls below the cap and eventually reaches $5/6$, or about 83.3. An output of 420 scores 93.5 here and 95.0 on an eight-value chart, where one value has less influence. Without printed labels, 420 falls within the wider allowance and receives full credit.}
\label{fig:chart-tolerance}

\end{figure}

\paragraph{Error severity.}
A numeric error larger than $a$ and at most $3a$ is a major \ev{value off}.
A larger error is a critical \ev{wrong value} (\Cref{fig:chart-tolerance}).
For cell errors, such as wrong or missing values, the cap follows \Cref{eq:cap} with $K=10$ cells.
Structural errors, such as a missing series, a lost legend or an invented category, receive fixed penalties regardless of page length.
Omission penalties within a chart cannot exceed the penalty for omitting the entire chart.

On a labeled two-series bar chart, reading 412.5 as 512.5 produces a critical \ev{wrong value} and a page score of 87.5.
Reading it as 420, an error of 1.8\%, produces a major \ev{value off} and a score of 95.0.
In this example, rescaled units and abbreviated labels receive no penalty, although they fail the \pbone{} rules.

The leaderboard uses \emph{full-table mode}, which scores every annotated chart cell.
For pages annotated only with selected points, the scorer retains a separate mode that uses the \pbone{} rules to measure recall.

\FloatBarrier
\subsubsection{Visual grounding}
\label{sec:layout}

\newcommand{\layWIcon}{1.3}
\newcommand{\layWPfive}{15.5}
\newcommand{\layWCount}{8.3}
\newcommand{\layAreaIcon}{0.15}
\newcommand{\layAreaPageNo}{0.17}
\newcommand{\layGrowVoneLast}{2.2}
\newcommand{\layGrowMinS}{1.5}
\newcommand{\layGrowMin}{4}
\newcommand{\layGrowMinIoU}{0.44}
\newcommand{\layGrowRebound}{22}

The Visual grounding metric checks whether parsed content can be located on the source page.
Each output block, such as a paragraph, table, picture or page header, must have a box around the content it represents.
For text blocks, the text must also match the content at that location.

\paragraph{Allowing different block sizes.}
A parser may combine two paragraphs into one block or split a paragraph into lines.
We accept either choice when the boxes correctly locate the content.
The output defines the block size, and the annotated element and word boxes let us check its location and text.

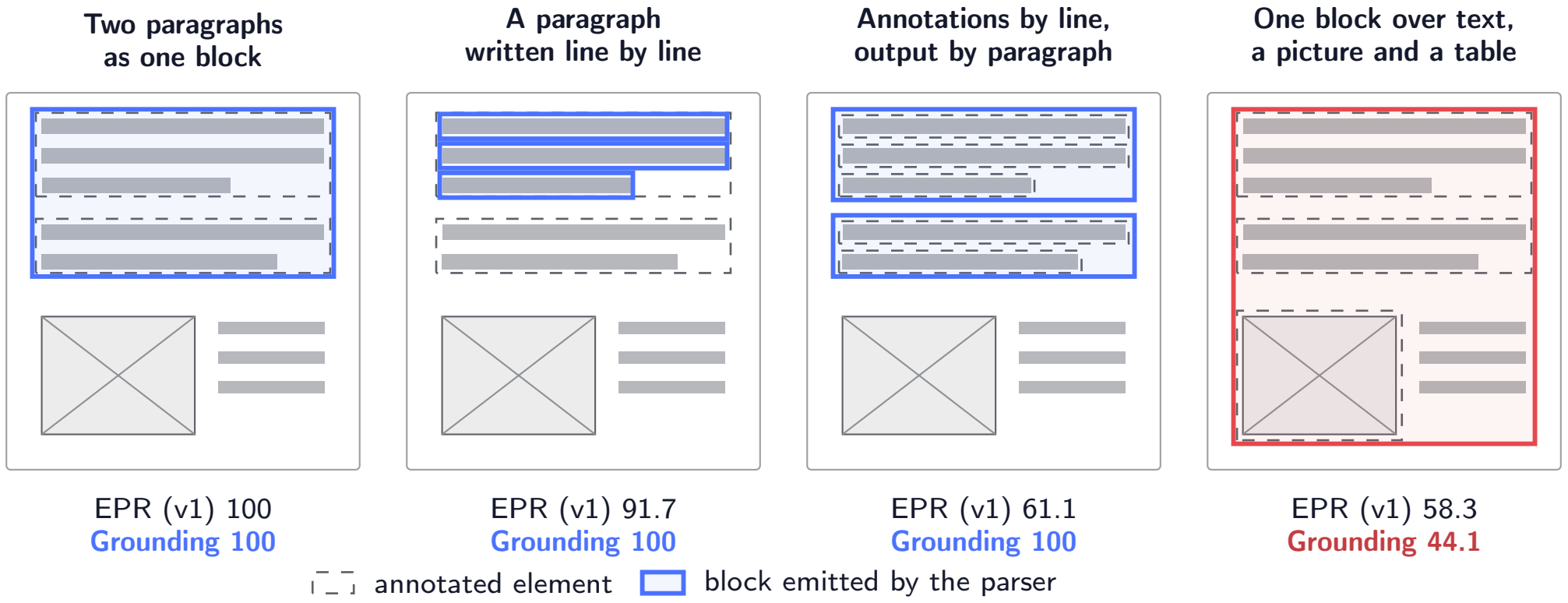
\begin{figure}[!htbp]
\centering
\begin{tikzpicture}[font=\scriptsize\sffamily,
  ln/.style={fill=LIMuted!45, draw=none},
  refb/.style={draw=LIMuted, dashed, line width=0.5pt},
  pred/.style={draw=pbNext, line width=1.1pt, fill=pbNext, fill opacity=0.06},
  ttl/.style={anchor=south, font=\scriptsize\sffamily\bfseries, align=center},
  sc/.style={anchor=north, align=center, font=\scriptsize\sffamily}]
\newcommand{\lgpage}[1]{%
  \draw[LIMuted!60, fill=white, rounded corners=1pt] (#1,0) rectangle ++(3.0,3.2);
  \foreach \y/\w in {2.85/2.4, 2.6/2.4, 2.35/1.6}{\fill[ln] (#1+0.3,\y) rectangle ++(\w,0.13);}
  \foreach \y/\w in {1.95/2.4, 1.7/2.0}{\fill[ln] (#1+0.3,\y) rectangle ++(\w,0.13);}
  \draw[LIMuted, fill=LIMuted!12] (#1+0.3,0.3) rectangle ++(1.3,1.0);
  \draw[LIMuted!70] (#1+0.3,0.3) -- ++(1.3,1.0) (#1+0.3,1.3) -- ++(1.3,-1.0);
  \foreach \y in {1.15,0.9,0.65}{\fill[ln] (#1+1.8,\y) rectangle ++(0.9,0.1);}}
\lgpage{0}
\draw[refb] (0.25,2.32) rectangle (2.75,3.03); \draw[refb] (0.25,1.67) rectangle (2.75,2.13);
\draw[pred] (0.22,1.64) rectangle (2.78,3.06);
\node[ttl] at (1.5,3.3) {Two paragraphs\\as one block};
\node[sc] at (1.5,-0.1) {EPR (v1) 100\\\textcolor{pbNext}{\textbf{Grounding 100}}};
\lgpage{3.4}
\draw[refb] (3.65,2.32) rectangle (6.15,3.03); \draw[refb] (3.65,1.67) rectangle (6.15,2.13);
\foreach \y/\w in {2.85/2.4, 2.6/2.4, 2.35/1.6}{\draw[pred] (3.68,\y-0.04) rectangle ++(\w+0.04,0.21);}
\node[ttl] at (4.9,3.3) {A paragraph\\written line by line};
\node[sc] at (4.9,-0.1) {EPR (v1) 91.7\\\textcolor{pbNext}{\textbf{Grounding 100}}};
\lgpage{6.8}
\foreach \y/\w in {2.85/2.4, 2.6/2.4, 2.35/1.6, 1.95/2.4, 1.7/2.0}{\draw[refb] (7.07,\y-0.03) rectangle ++(\w+0.06,0.19);}
\draw[pred] (7.02,2.29) rectangle (9.58,3.06); \draw[pred] (7.02,1.64) rectangle (9.58,2.16);
\node[ttl] at (8.3,3.3) {Annotations by line,\\output by paragraph};
\node[sc] at (8.3,-0.1) {EPR (v1) 61.1\\\textcolor{pbNext}{\textbf{Grounding 100}}};
\lgpage{10.2}
\draw[refb] (10.45,2.32) rectangle (12.95,3.03); \draw[refb] (10.45,1.67) rectangle (12.95,2.13); \draw[refb] (10.45,0.25) rectangle (11.85,1.35);
\draw[pred, draw=sevCrit, fill=sevCrit] (10.42,0.22) rectangle (12.98,3.06);
\node[ttl] at (11.7,3.3) {One block over text,\\a picture and a table};
\node[sc] at (11.7,-0.1) {EPR (v1) 58.3\\\textcolor{sevCrit!85!black}{\textbf{Grounding 44.1}}};
\draw[refb] (2.6,-1.05) rectangle ++(0.35,0.18); \node[anchor=west] at (3.0,-0.96) {annotated element};
\draw[pred] (5.4,-1.05) rectangle ++(0.35,0.18); \node[anchor=west] at (5.8,-0.96) {block emitted by the parser};
\end{tikzpicture}
\caption{\textbf{Accepting different block sizes in Visual grounding.} Merging two paragraphs or splitting one into lines preserves their text and location. Our metric scores both representations at 100. \pbone{} EPR checks each annotated element separately, so it penalizes split output and paragraph blocks compared with line-level annotations. Both metrics penalize a single box that spans five elements across a picture and a table. Scores cover the full synthetic 12-element page, of which the drawing shows a portion.}
\label{fig:layout-granularity}
\end{figure}

Layout benchmarks such as PubLayNet and DocLayNet use COCO mean average precision to compare predicted boxes with annotated elements of each class~\citep{zhong2019publaynet,pfitzmann2022doclaynet,lin2014coco}.
\pbone{} Element Pass Rate (EPR) uses looser overlap requirements, allowing two paragraphs in one block to pass.
However, it still checks each annotated element's text separately.
In \Cref{fig:layout-granularity}, EPR scores a paragraph split into lines at 91.7 and a paragraph block against line-level annotations at 61.1.
Our metric scores these representations and the merged paragraphs at 100.

\begin{figure}[!htbp]
\centering
\begin{tikzpicture}
\begin{groupplot}[group style={group size=4 by 1, horizontal sep=0.95cm},
  pbaxis, width=4.55cm, height=3.9cm, ymin=-3, ymax=108, grid=major,
  legend style={at={(0.5,-0.36)}, anchor=north, legend columns=1, fill=none}]
\nextgroupplot[xlabel={IoU with the annotated box}, ylabel={credit (\%)}, xmin=0, xmax=1, title={\footnotesize (a) per-element credit}]
\addplot[pbLegacy, very thick, const plot] table[x=iou, y expr=100*\thisrow{v1grow}] {layout_ramp.dat};
\addlegendentry{v1 EPR, box too large}
\addplot[pbLegacy, very thick, dashed, const plot] table[x=iou, y expr=100*\thisrow{v1shrink}] {layout_ramp.dat};
\addlegendentry{v1 EPR, box too small}
\addplot[pbNext, very thick] table[x=iou, y expr=100*\thisrow{v3}] {layout_ramp.dat};
\addlegendentry{Visual grounding metric: ramp}
\nextgroupplot[xlabel={box side scale $s$ (all boxes)}, title={\footnotesize (b) loose boxes}, xmin=1, xmax=2.6, clip=false]
\addplot[pbLegacy, very thick] table[x=s, y=v1] {layout_grow.dat};
\addlegendentry{v1 EPR}
\addplot[pbNext, very thick] table[x=s, y=v3] {layout_grow.dat};
\addlegendentry{Visual grounding metric}
\draw[black!50, densely dotted] (axis cs:2.236,-3) -- (axis cs:2.236,108) node[pos=0.55, left, font=\tiny, black!70, inner sep=1pt] {IoU 0.2};
\nextgroupplot[xlabel={random empty boxes added}, title={\footnotesize (c) junk boxes}, xmin=0, xmax=100]
\addplot[pbLegacy, very thick] table[x=n, y=v1] {layout_junk.dat};
\addlegendentry{v1 EPR}
\addplot[pbNext, very thick, mark=*, mark size=1pt] table[x=n, y=v3] {layout_junk.dat};
\addlegendentry{Visual grounding metric}
\nextgroupplot[xlabel={copies of every box}, title={\footnotesize (d) duplicate boxes}, xmin=0.7, xmax=6.3, xtick={1,...,6}]
\addplot[pbLegacyB, very thick, mark=square*, mark size=1.3pt] table[x=k, y=v1int] {layout_dup.dat};
\addlegendentry{v1 EPR, box and class only}
\addplot[pbLegacy, very thick, mark=triangle*, mark size=1.6pt] table[x=k, y=v1] {layout_dup.dat};
\addlegendentry{v1 EPR, full rule}
\addplot[pbNext, very thick, mark=*, mark size=1.3pt] table[x=k, y=v3] {layout_dup.dat};
\addlegendentry{Visual grounding metric}
\end{groupplot}
\end{tikzpicture}
\caption{\textbf{Effects of loose, empty and duplicate boxes.} (a)~EPR passes oversized boxes down to IoU 0.2 and undersized boxes down to 0.5. Visual grounding gives increasing credit from IoU 0.5 to 0.9. (b)--(d)~Scores on a synthetic 12-element page as boxes expand, empty boxes are added, or boxes are repeated with different classes. EPR accepts loose boxes until its overlap threshold fails and does not penalize added empty boxes. In (d), its box-and-class variant also ignores duplicates, while the full rule checks text. Visual grounding penalizes all three changes. In (b), scaling width and height by $s$ gives IoU $1/s^2$. The dotted line marks IoU 0.2. The slight later rise occurs when expanded boxes are matched to groups of neighboring elements.}
\label{fig:layout-epr}
\end{figure}
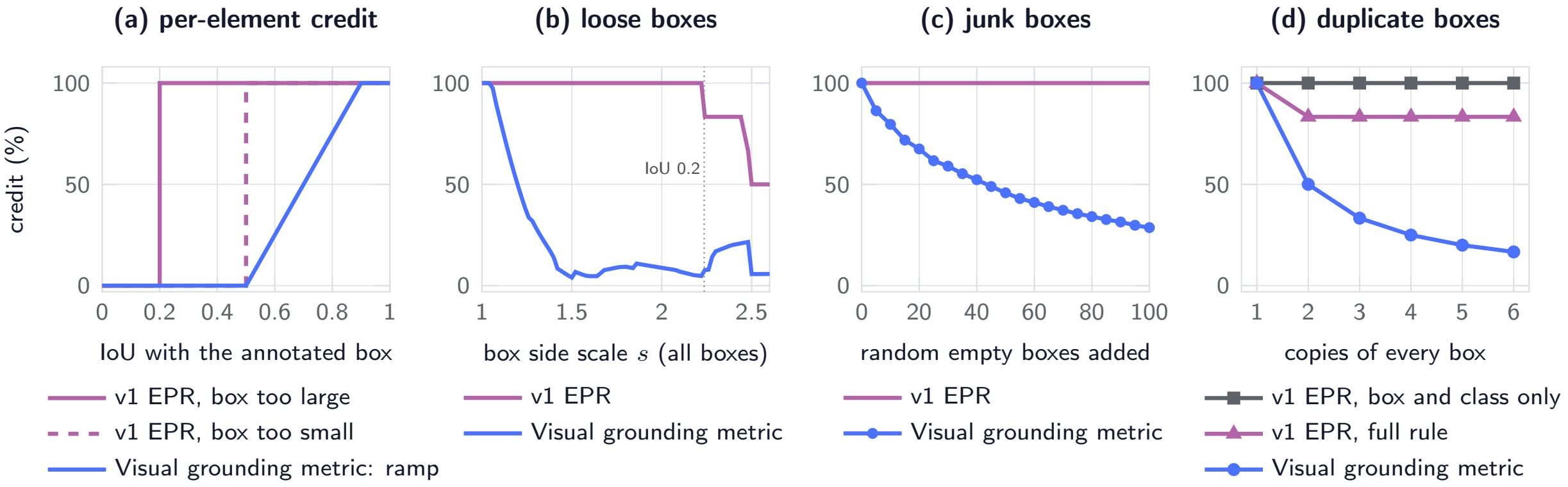

EPR passes an element when a prediction overlaps it sufficiently, has the expected basic class and contains its text.
It does not penalize extra boxes on blank areas.
On a synthetic 12-element page, adding 50 such boxes leaves EPR at 100\%.
A box can also pass until it is about $\layGrowVoneLast\times$ the element's width and height.
Finally, EPR gives the same penalty to a missing icon and a missing paragraph (\Cref{fig:layout-epr}).
Our metric gives partial credit for loose boxes, penalizes extra boxes and weights elements by size.

\paragraph{Matching content to boxes.}
To match an annotated element, a predicted box must cover at least half of it, and at least 20\% of the predicted box must overlap it.
These are the overlap requirements used by \pbone{}.
The predicted text must also match at a token F1 of at least 0.8.
We do not check the predicted class.
For an eligible match, credit $s_g$ rises linearly from 0 at an intersection over union (IoU) of 0.5 to 1 at 0.9.
With tight annotations, an IoU of 0.9 allows boxes about 5\% larger in each dimension.

We score several consecutive text elements as a group when one output block covers them.
We also group several output boxes that cover one annotated element.
Within each group, we compare the combined text and the enclosing boxes, separately for each column.
Page headers and footers can be grouped even when their parts are not consecutive.
We match tables and pictures one to one.
A table box must cover 90\% of its words.
Predictions that duplicate an element, cover blank areas or contain words already assigned elsewhere count as \emph{extra boxes}.

\paragraph{Page score.}
We weight each element by $w=\sqrt{\text{area}}$.
This gives a missing paragraph more weight than a missing icon, while limiting the influence of a large picture.
Extra boxes add to the denominator with the same weighting:
\begin{equation}
  \text{Grounding} = \frac{\sum_{g\in\text{Ref}} w_g\, s_g}{\sum_{g\in\text{Ref}} w_g + \sum_{p\in\text{extra}} w_p},
  \label{eq:epr-v3}
\end{equation}
Here, Ref contains the annotated elements, and extra contains the extra predictions.
Larger extra boxes reduce the score more (\Cref{fig:layout-epr}c).
The dimension score is the mean of page scores.
It uses no severity cap or per-error severity penalties.
The test suite still uses severity-based targets to evaluate how the score responds to misplaced or missing content (\Cref{app:suites-overview}).

We chose the IoU bounds and area weights for uses such as citation and highlighting.
We have not validated these choices in downstream applications or against independent annotations.
Our sensitivity analysis also leaves them fixed.

\FloatBarrier
\subsubsection{Forms}
\label{sec:forms}

\begin{figure}[!htbp]
\centering
\begin{tikzpicture}[font=\scriptsize\sffamily,
  fcap/.style={anchor=west, text=LIMuted, font=\tiny\sffamily},
  fval/.style={anchor=west, font=\scriptsize\sffamily},
  fbox/.style={draw=LIMuted!60, line width=0.4pt},
  obox/.style={draw=LIMuted!50, fill=pbRef, rounded corners=1.5pt, inner sep=3pt, anchor=north west, align=left, font=\tiny\ttfamily},
  fhead/.style={anchor=west, font=\scriptsize\sffamily\bfseries}]
\node[fhead] at (0,5.05) {a\enspace The form (9 annotated items)};
\draw[draw=LIMuted!70, fill=white, rounded corners=2pt] (0,0) rectangle (5.2,4.75);
\node[anchor=west, font=\scriptsize\sffamily\bfseries] at (0.15,4.5) {Applicant};
\foreach \y/\c/\v in {3.95/Full name/Ingrid Sorensen, 3.35/Date of birth/{09/18/1987}, 2.75/Policy number/PL-4471-2093, 2.15/Annual income/{\$48,250.00}}{
  \node[fcap] at (0.15,\y+0.2) {\c};
  \draw[fbox] (0.15,\y-0.2) rectangle (3.4,\y+0.08);
  \node[fval] at (0.2,\y-0.06) {\v};}
\node[fcap] at (3.55,2.35) {Middle name};
\draw[fbox] (3.55,1.95) rectangle (5.05,2.23);
\node[fcap] at (0.15,1.55) {Prior claims};
\draw[fbox] (1.35,1.47) rectangle ++(0.16,0.16); \node[fval] at (1.53,1.55) {Yes};
\draw[fbox] (2.1,1.47) rectangle ++(0.16,0.16); \node[fval] at (2.28,1.55) {No};
\draw[line width=0.7pt] (2.12,1.49) -- (2.24,1.61) (2.12,1.61) -- (2.24,1.49);
\draw[fbox] (0.15,1.0) rectangle ++(0.16,0.16); \draw[line width=0.7pt] (0.17,1.02) -- (0.29,1.14) (0.17,1.14) -- (0.29,1.02);
\node[fval] at (0.35,1.08) {I agree to the terms};
\node[fcap] at (0.15,0.62) {Applicant signature};
\draw[LIMuted] (0.15,0.22) -- (3.4,0.22);
\draw[line width=0.6pt] (0.3,0.25) .. controls (0.6,0.55) and (0.8,0.05) .. (1.1,0.38) .. controls (1.3,0.55) and (1.5,0.2) .. (1.8,0.33);
\node[fhead] at (5.6,5.05) {b\enspace Two outputs, both 100};
\node[obox, text width=3.15cm] at (5.6,4.75) {- **Full name**: Ingrid Sorensen\\- **Date of birth**: 1987-09-18\\- **Policy number**: PL-4471-2093\\- **Annual income**: 48,250.00\\- **Middle name**:\\- **Prior claims**: Yes [ ] No [x]\\- **I agree to the terms**: [x]\\- **Applicant signature**: signed};
\node[obox, text width=3.15cm] at (5.6,2.3) {| Full name | Ingrid Sorensen |\\| Date of birth | 18/09/1987 |\\| Policy number | PL-4471-2093 |\\| Annual income | \$48,250 |\\| Middle name | |\\| Prior claims | $\square$ Yes $\boxtimes$ No |\\| I agree to the terms | $\boxtimes$ |\\| Applicant signature | signed |};
\node[anchor=north west, text=sevHarm!80!black, font=\tiny\sffamily\bfseries] at (5.6,-0.05) {key: value lines and a table, other date and amount formats: $S=100$};
\node[fhead] at (9.75,5.05) {c\enspace One change at a time};
\foreach \y/\chg/\pp/\ff/\ss/\col/\cls/\evt in {
  4.35/{PL-4471-2093 $\to$ PL-4477-2093}/91.7/90.0/90.0/sevCrit/C/wrong value,
  3.45/{Agreement box unticked}/91.7/90.0/90.0/sevCrit/C/box state flipped,
  2.55/{Annual income left empty}/95.8/95.0/95.0/sevMaj/M/value lost,
  1.65/{Sorensen $\to$ Sorenson}/98.3/99.0/98.3/sevMin/m/value typo}{
  \node[anchor=west, font=\scriptsize\sffamily] at (9.75,\y) {\chg};
  \node[anchor=west, font=\tiny\sffamily, text=LIMuted] at (9.75,\y-0.33) {$m$ \pp\quad $c$ \ff};
  \node[anchor=west, font=\tiny\sffamily] at (11.35,\y-0.33) {\sevbadge{\col}{\cls}\ \ev{\evt}};
  \node[anchor=east, font=\scriptsize\sffamily\bfseries] at (14.85,\y) {$S$ \ss};}
\node[anchor=west, font=\tiny\sffamily, text=LIMuted, text width=5cm, align=left] at (9.75,0.55) {$S=\min(m,c)$; a wrong value caps the page at 90, a lost value costs 5 and a typo 1.7.};
\end{tikzpicture}
\caption{\textbf{Scoring form values and their labels.} (a)~A synthetic form with nine scored items, including text fields, an empty field, checkboxes and a signature. (b)~Key--value lines and a table preserve the same information despite differences in date, amount and checkbox notation. Both score 100. (c)~Changing one entry at a time shows how wrong values, omissions and typos affect the score. $m$ is the parsing score, $c$ is the severity cap and $S$ is the final score.}
\label{fig:forms-example}
\end{figure}
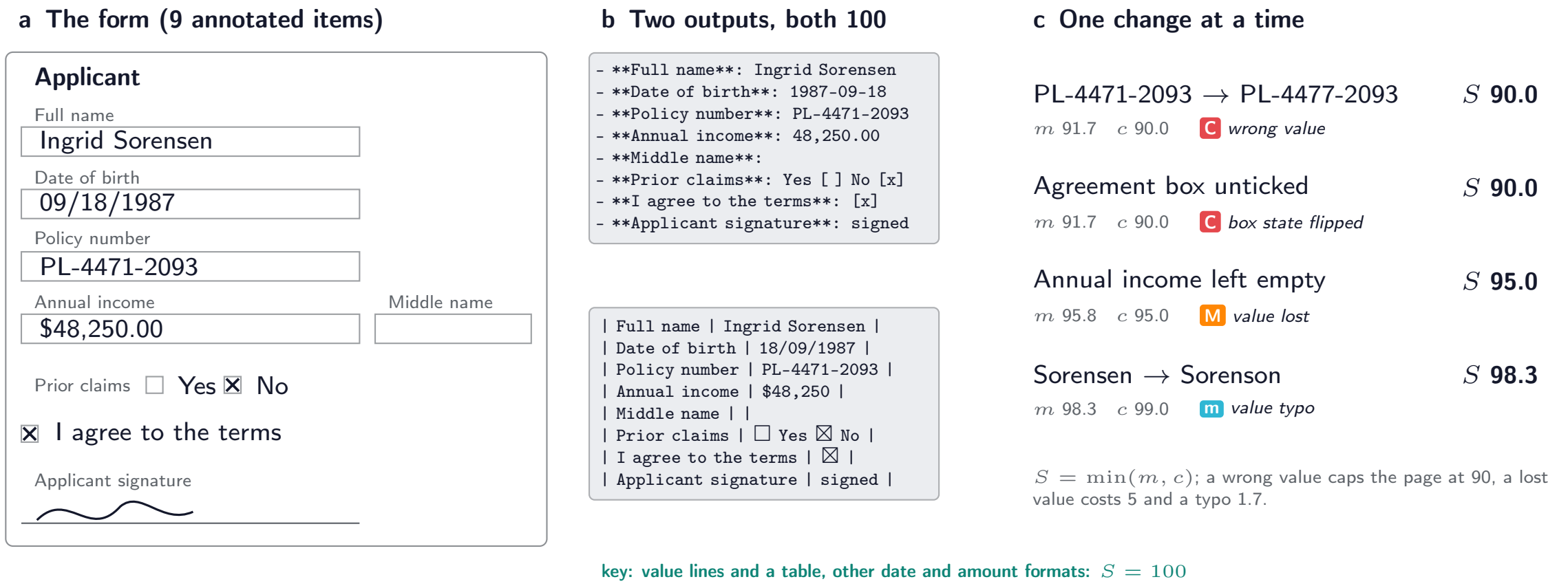

A \emph{form} contains labeled fields, checkboxes, choices and signature lines for a person to fill in.
The Forms metric checks the recovered values and checkbox states, together with the labels that give them meaning (\Cref{fig:forms-example}).
It accepts several ways of representing the same content.
We score the page Markdown.
A provider's separate key--value output is scored only when it is included in that Markdown (\Cref{sec:options}).

\paragraph{Value correctness.}
We compare the entered text of each field with its annotated value.
For a checkbox, we check whether the output correctly reports it as checked or unchecked.

\paragraph{Semantic binding.}
We also check whether each value or checkbox state is attached to the correct label.
For example, \emph{Ingrid Sorensen} belongs to \emph{Full Name} in \Cref{fig:forms-example}.
Reading the name correctly but placing it under \emph{Date of birth} still changes the information.
A parser can therefore fail by misreading a value or by attaching it to the wrong field.
Printed identifiers, section headings and table context help distinguish fields with similar labels.

\paragraph{Equivalent representations.}
In \Cref{fig:forms-example}b, key--value lines and a two-column table preserve the same values, checkbox states and label associations.
They therefore receive the same score.
We also accept common checkbox symbols and equivalent date and amount formats.

\paragraph{Error severity.}
Forms uses the same item loss weights and severity-cap framework as Formatting (\Cref{sec:formatting,app:metric-cap}).
Our policy assigns severity according to the effect on the field's information:
\begin{itemize}[leftmargin=*,itemsep=2pt]
  \item \textbf{Critical:} a wrong value, a value attached to the wrong label, a value invented in a blank field, or a checkbox whose state is flipped.
  \item \textbf{Major:} a missing value, an omitted tick or a lost signature.
  \item \textbf{Minor:} a typo or an omitted blank field.
  \item \textbf{Benign variation:} an equivalent representation of the same content, such as key--value lines instead of a two-column table.
\end{itemize}

\Cref{fig:forms-example}c shows how changing one entry at a time affects the score while the rest of the form remains correct.
A wrong policy number or a flipped checkbox lowers the score from 100 to 90.0.
An omitted annual income scores 95.0, and a typo in the surname scores 98.3.

\Cref{tab:suite-results} compares the metric with a generic exact-match field F1 baseline on the Forms test suite.
\FloatBarrier

\subsection{Severity Cap}
\label{app:metric-cap}

Text, Formatting, Tables, Charts and Forms use the same procedure to compute the severity cap (\Cref{sec:score}).
Each metric first normalizes the output and aligns it with the annotations.
It then identifies errors in the mismatches, such as a missing passage or a table value under the wrong header.
The scorer records each error's type and severity level.
It groups some errors that share a cause, such as a whole table written in a different decimal convention.

One wrong number can mislead a reader regardless of page length.
We therefore apply the full penalty to the first error at each severity level.
For later errors, we reduce the penalty on long pages to account for the amount of content being parsed.

The scaling factor is $\min(1, K/N)$, where $N$ measures page content and $K$ is a constant for the metric.
Text counts tokens, Tables and Charts count cells, Formatting counts scored items, and Forms counts fields.
We use annotation counts for $N$, except in Formatting, which also counts items added by the output.
Minor errors in Text, Formatting and Forms are not scaled by page length.
For errors that use this scaling, the \emph{effective error count} at severity level $\ell$ is
\begin{equation}
  n^{\mathrm{eff}}_\ell \;=\; \min(n_\ell, 1) \;+\; \max(0, n_\ell - 1)\cdot\min\!\Big(1, \frac{K}{N}\Big),
  \label{eq:neff}
\end{equation}
where $n_\ell$ counts detected errors at level $\ell \in \{\crit, \maj, \mnr\}$ (critical, major, minor).
For unscaled minor errors, we use the error count directly.
The severity cap is 1 minus the combined penalties, with a minimum of zero:
\begin{equation}
  c = \max\!\Big(0,\ 1 - w_{\crit}\, n^{\mathrm{eff}}_{\crit} - w_{\maj}\, n^{\mathrm{eff}}_{\maj}
  - \min\big(0.15,\; w_{\mnr}\, n^{\mathrm{eff}}_{\mnr}\big)\Big),
  \label{eq:cap}
\end{equation}
with $w_{\crit}=0.10$, $w_{\maj}=0.05$ and $w_{\mnr}=0.01$.
Minor errors together can lower the cap by at most 15 points.
This limits the effect of many small errors.

For example, Text uses $K=400$ tokens.
On a 1{,}200-token page, five wrong numbers count as $1 + 4\cdot\frac{400}{1200} = 2.33$ effective critical errors, giving a cap of 76.7.
On a 300-token page, all five receive the full penalty, giving a cap of 50.
We adjust these weights and constants to meet the policy's targets (\Cref{sec:metric-construction}).
The targets specify the required score response, while the weights and constants implement it.

\clearpage
\section{Metric Evaluation}
\label{app:metric-evaluation}
\subsection{Metric Sources and Adaptations}
\label{app:metric-sources}

\Cref{tab:metric-sources} maps the metric columns in \Cref{tab:grid-text,tab:grid-formatting,tab:grid-tables,tab:grid-charts,tab:grid-forms,tab:grid-layout} to their sources.
Citations identify the published definitions or software projects.
We run local suite implementations, with adaptations noted below.
Our baselines, design variants and ablations are identified separately.

{\RClongsetup\small\setlength{\tabcolsep}{5pt}\renewcommand{\arraystretch}{1.02}
\begin{longtable}{@{}>{\raggedright\arraybackslash}p{0.39\textwidth}>{\raggedright\arraybackslash}p{\dimexpr0.61\textwidth-10pt\relax}@{}}
\caption{Sources and implementation notes for every metric in the severity test suites. Names match the grid columns, with related variants grouped together.}\label{tab:metric-sources}\\
\toprule
\RChead{Metric columns} & \RChead{Source and suite implementation} \\
\midrule
\endfirsthead
\multicolumn{2}{l}{\scriptsize\sffamily\color{LIMuted}Metric sources and adaptations, continued}\\
\toprule
\RChead{Metric columns} & \RChead{Source and suite implementation} \\
\midrule
\endhead
\bottomrule
\endlastfoot

\RCband{2}{TEXT} \\*
\pbtwo{} Text, Move-tolerant fidelity (component average), Next: reading-order component, Ordered critical tokens F1, Bag-of-words recall
& Our metric (\Cref{sec:text}), component variants and internal baselines. \\
Content Faithfulness, Order rule pass rate
& \pbone{}~\citep{zhang2026parsebench}. The suite generates order rules from consecutive sentences. \\
Character accuracy ($1-$CER), Word accuracy ($1-$WER)
& Levenshtein distance~\citep{levenshtein1966binary} with JiWER-style default normalization~\citep{jiwer2026}, expressed as accuracy and clipped at zero. \\
ISRI word accuracy
& ISRI/ocreval~\citep{ocreval2026}. The local implementation aligns lowercased Latin-1 words with a longest common subsequence. \\
$1-$edit distance, Unique-word F1
& Nougat evaluation~\citep{blecher2023nougat}. Character edit similarity and set F1 after removing tables. \\
BLEU-4
& BLEU~\citep{papineni2002bleu}, using whitespace tokens and no smoothing. \\
METEOR
& METEOR~\citep{banerjee2005meteor}. Local approximation with suffix stripping and no WordNet synonym matching. \\
Full-page OCR score, Document parsing STEDS
& OCRBench v2~\citep{fu2025ocrbenchv2}: the four-score OCR composite and document-tree structure score. \\
Word F1 (multiset)
& CC-OCR~\citep{yang2024ccocr}. F1 over normalized word counts. \\
NID (Indel similarity)
& DP-Bench~\citep{upstage2026dpbench}. Markup-stripped Markdown stands in for extracted element text. \\
CCT accuracy
& Unstructured evaluation code~\citep{unstructured2026eval}. The suite approximates concatenated element text with stripped Markdown. \\
$1-$tokens added, Tokens found
& SCORE~\citep{li2025score}. Local tokenization replaces the upstream tokenization. \\
Word order (Kendall)
& READoc~\citep{li2024readoc}, using a Kendall-based comparison of shared word order~\citep{kendall1938tau}. \\
Word edit distance
& Docling evaluation code~\citep{docling2026eval}. A regular-expression tokenizer replaces NLTK tokenization. \\
Text $1-$NED, Text $1-$NED (MGAM matching), Reading order $1-$ED
& OmniDocBench~\citep{ouyang2024omnidocbench,omnidocbench2026code}. Local block and segmentation matching use greedy assignment. \\
Text + reading order
& FinixDoc~\citep{wang2026finixdoc}. Our text-only adaptation averages text and reading-order scores, omitting its table component. \\
chrF (character n-gram F)
& chrF~\citep{popovic2015chrf}, also used by KITAB-Bench~\citep{heakl2025kitab}. Character n-grams of orders 1--6 with $\beta=2$. \\
Unit-test pass rate
& olmOCR-bench tests~\citep{allenai2026olmocrbench}. Tests are generated from annotations, with a local fuzzy-matching approximation. \\
Heuristic score
& Marker benchmark code~\citep{marker2026benchmark}. Local block matching and order weighting omit the Pandoc round trip. \\
Flexible character accuracy
& Clausner et al.~\citep{clausner2020flexible}. A simplified line search uses fixed coefficients and a limited candidate shortlist. \\

\RCband{2}{FORMATTING} \\*
\pbtwo{} Formatting, \nolinkurl{formatting_v2} proportional only, \nolinkurl{formatting_v2} cap only
& Our metric and ablations (\Cref{sec:formatting}). \\
Semantic Formatting, Normalized text styling, Normalized title accuracy, All formatting rules pass rate
& \pbone{}~\citep{zhang2026parsebench} and its component rule scores. Rules are generated from the page annotations. \\
FormatTest pass rate
& olmOCR-bench's FormatTest~\citep{allenai2026olmocrbench}, with exact substring matching for annotated styled spans. \\
Tag-sequence $1-$NED
& MDEval~\citep{chen2025mdeval}. A local Markdown converter produces the HTML tag sequence. \\
Heading TEDS
& READoc~\citep{li2024readoc}. Tree comparison of ATX headings and their relative nesting. \\
Title F1 / level accuracy harmonic mean
& FinTOC~\citep{maarouf2021fintoc}, adapted to page headings. Titles match at normalized Levenshtein similarity $\geq0.85$. The score combines title F1 and absolute level accuracy, without a page-number constraint. \\
MHS heading tree
& OpenDataLoader benchmark~\citep{opendataloader2026bench}. Flat heading-and-content trees. \\
STEDS
& OCRBench v2~\citep{fu2025ocrbenchv2}. Document-tree structure scoring. \\
Raw Markdown $1-$NED
& Raw-string edit similarity, as used in Nougat and READoc~\citep{blecher2023nougat,li2024readoc}. \\
Text $1-$NED (cleaned)
& OmniDocBench-inspired normalization~\citep{ouyang2024omnidocbench}, applied to the page text. \\
Style-token F1, Heading F1, exact level
& Our generic baselines over style--word pairs and heading-level--title pairs. \\

\RCband{2}{TABLES} \\*
\pbtwo{} Tables
& Our metric (\Cref{sec:tables}). \\
GriTS-Con (\pbone{}), TableRecordMatch, GTRM composite
& \pbone{}~\citep{zhang2026parsebench}, including its GriTS component~\citep{smock2023gritsgridtablesimilarity}. Local pairing uses the best predicted table and omits some title-handling retries. \\
T-LAG (cell-neighbour graph F1)
& PulseBench-Tab~\citep{pandey2026pulsebenchtab}. Weighted matching of neighbouring-cell relations. \\
Shifted content accuracy
& SCORE~\citep{li2025score}. Content comparison permits row and column offsets. \\
Position accuracy (text adaptation)
& GAP position accuracy~\citep{choi2026gap}, adapted to compare cell text at each grid position instead of bounding-box pointers. \\
TEDS, TEDS-S
& PubTabNet TEDS~\citep{zhong2020imagebasedtablerecognitiondata}. TEDS-S omits cell text. \\
TEDS (v1.6 normalization), TEDS (normalized), $1-$table edit distance
& OmniDocBench and DP-Bench normalization~\citep{ouyang2024omnidocbench,omnidocbench2026code,upstage2026dpbench}. The suite simplifies pairing and LaTeX handling. \\
GriTS-Con, GriTS-Top
& GriTS~\citep{smock2023gritsgridtablesimilarity}: content and grid-topology variants. \\
Adjacency F1
& SciTSR~\citep{chi2019scitsr} and the ICDAR table competition~\citep{gobel2013icdar}. F1 over neighbouring-cell relations. \\
Table test pass rate
& olmOCR-bench~\citep{allenai2026olmocrbench}. The suite generates cell-relation tests from annotations. \\
Alignment score
& RD-TableBench~\citep{reducto2026tablebench}. Hierarchical row and cell alignment, with an approximate aligned-row count. \\
Composite structure acc
& Unstructured table evaluation~\citep{unstructured2026eval}. Combines cell positions and content similarity. \\
AccCon (exact table)
& PubTables-1M~\citep{smock2022pubtables}. A table passes only when GriTS-Con is perfect. \\
Relation F1, exact / chrF
& Text-to-Table~\citep{wu2022texttotable} with exact or chrF matching~\citep{popovic2015chrf}. The local variant treats identical empty strings as a perfect match. \\
Cell accuracy (TCE), Row and column F1 (RCE), Merged-cell F1 (MCD), Table size match
& MMTab/Table-LLaVA~\citep{zheng2024mmtab}. Individual question-based probes are applied across the full table. \\
JSON accuracy
& OmniAI OCR benchmark~\citep{omniai2026benchmark}. Tables are converted into header-keyed row objects. \\
Structure BLEU-4
& TableBank~\citep{li2020tablebank}, using BLEU~\citep{papineni2002bleu} over table structure tokens. \\
Bigram Jaccard
& Text-based table detection (txt-TD)~\citep{soric2026benchmarking}. The local variant compares character-bigram multisets over serialized cells. \\

\RCband{2}{CHARTS} \\*
\pbtwo{} Charts
& Our metric (\Cref{sec:charts}). \\
ChartDataPoint pass rate
& \pbone{}~\citep{zhang2026parsebench}. The suite simplifies header chains and uses one tolerance per sample. \\
RMS F1, RNSS
& DePlot's mapping and number-set metrics~\citep{liu2023deplot}. The suite linearizes one output table for RMS. \\
SCRM strict / slight / high
& StructChart~\citep{xia2023structchart}, using the ChartX implementation~\citep{xia2024chartx}. Reports per-chart similarity at three tolerances. \\
Relaxed accuracy (QA by lookup)
& ChartQA~\citep{masry2022chartqa}. Table lookup supplies each answer instead of a question-answering model. \\
Chart parsing TEDS
& OCRBench v2~\citep{fu2025ocrbenchv2}. The suite converts table rows into a category-keyed dictionary before scoring. \\
$1-$A-MAPE
& ExChart-Bench~\citep{he2026exchart}. Positional numerical error, expressed as a similarity. \\
EpiCurveSimilarity
& EpiCurveBench~\citep{berkane2026epicurvebench}. Series alignment with edit distance with real penalty. \\
Row IoU (medium tolerance)
& ChartAnchor~\citep{li2025chartanchor}. Whole-row tuple matching at the suite's selected tolerance. \\
$0.5$ RMS + $0.5$ TEDS$'$
& Sci-ImageMiner~\citep{ahmed2026sciimageminer}. The suite combines numerical and table-structure similarity. \\
$(1-\mathrm{nMAE})\times$ coverage
& Our composite of the numerical error and completeness measures in ChartDensity-Bench~\citep{wu2026chartdensity}. \\
Metric 6b
& CHART-Infographics evaluation code~\citep{chartinfo2026metrics}. The suite applies discrete-point scoring to pie charts. \\
ChartOCR score
& ChartOCR~\citep{luo2021chartocr}. Pixel-box distances are replaced by value-space distances, and pie slices by numerical shares. \\
RMS-TBE-F1, RMS-TBE-F1-Sig
& FairChart2Table~\citep{song2026fairchart}. The suite uses the sample chart's tick gap. \\
Normalized mapping similarity
& PaperUnPlot NMS~\citep{bertillo2026paperunplot}. The local variant uses linear axis ranges and a 10\% cutoff for value error. \\
Point F1 (2\% box)
& Scatteract~\citep{cliche2017scatteract}, adapted to bar, line and pie data by assigning category indices to horizontal positions. \\

\RCband{2}{FORMS} \\*
\pbtwo{} Forms, \nolinkurl{forms_v2} proportional only, \nolinkurl{forms_v2} cap only
& Our metric and ablations (\Cref{sec:forms}). \\
\nolinkurl{form_kv_md_core_kv_f1}, \nolinkurl{rule_form_field_pass_rate}
& \pbone{}~\citep{zhang2026parsebench}. Scores are precomputed with its Python scorers. \\
\nolinkurl{form_bench_text_value_f1}, \nolinkurl{form_bench_text_attribution_f1}
& Our unpublished Form Bench draft: value matching and attribution to the field's section and table row. \\
Entity-linking F1
& FUNSD/XFUND-inspired link scoring~\citep{jaume2019funsd,xu2022xfund}. A generic Markdown extractor supplies caption--value pairs instead of OCR-box entities. \\
Field F1 (exact)
& Our generic exact-match field F1 baseline. Matches normalized field names and values. \\
ANLS per field
& DocVQA ANLS~\citep{mathew2021docvqa}. Each field becomes a lookup question with the usual 0.5 edit-distance threshold. \\
$1-$NED per field
& Our generic baseline using mean normalized Levenshtein similarity~\citep{levenshtein1966binary} over field values. \\
Checkbox accuracy
& CheckboxQA-inspired task~\citep{turski2025checkboxqa}. Our baseline scores each option's nearest checkbox state. \\
Value presence
& olmOCR-bench-inspired presence tests~\citep{allenai2026olmocrbench}. Checks whether each filled value occurs anywhere on the page. \\
Page text $1-$NED
& A local adaptation of text edit scoring~\citep{ouyang2024omnidocbench}, using whole-page words rather than matched character blocks. \\

\RCband{2}{VISUAL GROUNDING} \\*
\pbtwo{} Visual grounding
& Our metric (\Cref{sec:layout}). \\
EPR v3 draft
& Our internal design variant. \\
EPR v1, EPR v1, no text gate
& \pbone{}~\citep{zhang2026parsebench} and our ablation removing its text gate. \\
\end{longtable}
}

\clearpage
\subsection{Metric Responses to Test Scenarios}
\label{app:test-suites}

\Cref{tab:grid-text,tab:grid-formatting,tab:grid-tables,tab:grid-charts,tab:grid-forms,tab:grid-layout} show how the metrics respond to the perturbations in the six severity test suites.
Each row describes one perturbation, and each column is a metric.

\paragraph{Evaluation setup.}
We score the correct and perturbed outputs against the same page annotations, following \Cref{sec:lab-protocol}.
For each perturbation, we average the score drop over the pages it changes and compare this average with the policy target.
The targets are drops of at least 10 points for critical errors, at least 5 for major errors, and 1 to 25 for minor errors.
For benign variations, a drop below 0.5 points counts as no penalty.

\paragraph{Reading the results.}
A check mark (\textcolor{sevHarm}{\ding{51}}) means the average score drop meets the target.
A tilde (\textcolor{gridWeak}{$\sim$}) marks a partial response: a critical or major error lowers the score by at least 1 point but less than required, or a minor error lowers it by more than 25 points.
A cross (\textcolor{sevCrit}{\ding{55}}) means an error causes a drop below 1 point, or a benign variation loses at least 0.5 points.
A grey dash (\textcolor{LIMuted}{--}) marks an ungraded, debatable perturbation.
A dot (\labna{}) marks a cell with no scored cases.

The PCS row summarizes policy agreement across the individual test scenarios using \Cref{eq:pcs}.
Columns run from highest to lowest PCS.

The Sev.\ column shows a numeric rating for each severity level: 90 for critical, 50 for major, 25 for minor and 0 for benign.
PCS uses violation rates by severity level, not these numeric ratings.

\begin{landscape}
{\RClongsetup\scriptsize\setlength{\tabcolsep}{0pt}\renewcommand{\arraystretch}{0.92}
\begin{longtable}{>{\raggedright\arraybackslash}p{5.2cm}r@{\hspace{4pt}}*{31}{>{\centering\arraybackslash}p{0.54cm}}}
\caption{Text: responses of 31 metrics to 51 perturbations on 2 sample pages. \ding{51}: target met. $\sim$: partial response. \ding{55}: target missed. Sev.: severity rating.}\label{tab:grid-text}\\
\toprule
\RChead{Perturbation} & \RChead{Sev.} & \rotatebox{90}{\parbox{3.3cm}{\raggedright\sffamily 1.~\textbf{\pbtwo{} Text}}} & \rotatebox{90}{\parbox{3.3cm}{\raggedright\sffamily 2.~Order rule pass rate}} & \rotatebox{90}{\parbox{3.3cm}{\raggedright\sffamily 3.~Text 1 $-$ NED}} & \rotatebox{90}{\parbox{3.3cm}{\raggedright\sffamily 4.~Ordered critical tokens F1}} & \rotatebox{90}{\parbox{3.3cm}{\raggedright\sffamily 5.~Move-tolerant fidelity (component average)}} & \rotatebox{90}{\parbox{3.3cm}{\raggedright\sffamily 6.~Unit-test pass rate}} & \rotatebox{90}{\parbox{3.3cm}{\raggedright\sffamily 7.~Content Faithfulness}} & \rotatebox{90}{\parbox{3.3cm}{\raggedright\sffamily 8.~Text + reading order}} & \rotatebox{90}{\parbox{3.3cm}{\raggedright\sffamily 9.~CCT accuracy}} & \rotatebox{90}{\parbox{3.3cm}{\raggedright\sffamily 10.~Bag-of-words recall}} & \rotatebox{90}{\parbox{3.3cm}{\raggedright\sffamily 11.~Flexible character accuracy}} & \rotatebox{90}{\parbox{3.3cm}{\raggedright\sffamily 12.~ISRI word accuracy}} & \rotatebox{90}{\parbox{3.3cm}{\raggedright\sffamily 13.~NID (Indel similarity)}} & \rotatebox{90}{\parbox{3.3cm}{\raggedright\sffamily 14.~Text 1 $-$ NED (MGAM matching)}} & \rotatebox{90}{\parbox{3.3cm}{\raggedright\sffamily 15.~Next: reading-order component}} & \rotatebox{90}{\parbox{3.3cm}{\raggedright\sffamily 16.~Word order (Kendall)}} & \rotatebox{90}{\parbox{3.3cm}{\raggedright\sffamily 17.~Reading order 1 $-$ ED}} & \rotatebox{90}{\parbox{3.3cm}{\raggedright\sffamily 18.~BLEU-4}} & \rotatebox{90}{\parbox{3.3cm}{\raggedright\sffamily 19.~Heuristic score}} & \rotatebox{90}{\parbox{3.3cm}{\raggedright\sffamily 20.~Word accuracy (1 $-$ WER)}} & \rotatebox{90}{\parbox{3.3cm}{\raggedright\sffamily 21.~Word edit distance}} & \rotatebox{90}{\parbox{3.3cm}{\raggedright\sffamily 22.~METEOR}} & \rotatebox{90}{\parbox{3.3cm}{\raggedright\sffamily 23.~chrF (character n-gram F)}} & \rotatebox{90}{\parbox{3.3cm}{\raggedright\sffamily 24.~Unique-word F1}} & \rotatebox{90}{\parbox{3.3cm}{\raggedright\sffamily 25.~Tokens found}} & \rotatebox{90}{\parbox{3.3cm}{\raggedright\sffamily 26.~Word F1 (multiset)}} & \rotatebox{90}{\parbox{3.3cm}{\raggedright\sffamily 27.~Full-page OCR score}} & \rotatebox{90}{\parbox{3.3cm}{\raggedright\sffamily 28.~1 $-$ tokens added}} & \rotatebox{90}{\parbox{3.3cm}{\raggedright\sffamily 29.~1 $-$ edit distance}} & \rotatebox{90}{\parbox{3.3cm}{\raggedright\sffamily 30.~Character accuracy (1 $-$ CER)}} & \rotatebox{90}{\parbox{3.3cm}{\raggedright\sffamily 31.~Document parsing STEDS}} \\
\midrule
\endfirsthead
\multicolumn{33}{l}{\scriptsize\sffamily\color{LIMuted}Text test suite, continued}\\
\toprule
\RChead{Perturbation} & \RChead{Sev.} & \rotatebox{90}{\parbox{3.3cm}{\raggedright\sffamily 1.~\textbf{\pbtwo{} Text}}} & \rotatebox{90}{\parbox{3.3cm}{\raggedright\sffamily 2.~Order rule pass rate}} & \rotatebox{90}{\parbox{3.3cm}{\raggedright\sffamily 3.~Text 1 $-$ NED}} & \rotatebox{90}{\parbox{3.3cm}{\raggedright\sffamily 4.~Ordered critical tokens F1}} & \rotatebox{90}{\parbox{3.3cm}{\raggedright\sffamily 5.~Move-tolerant fidelity (component average)}} & \rotatebox{90}{\parbox{3.3cm}{\raggedright\sffamily 6.~Unit-test pass rate}} & \rotatebox{90}{\parbox{3.3cm}{\raggedright\sffamily 7.~Content Faithfulness}} & \rotatebox{90}{\parbox{3.3cm}{\raggedright\sffamily 8.~Text + reading order}} & \rotatebox{90}{\parbox{3.3cm}{\raggedright\sffamily 9.~CCT accuracy}} & \rotatebox{90}{\parbox{3.3cm}{\raggedright\sffamily 10.~Bag-of-words recall}} & \rotatebox{90}{\parbox{3.3cm}{\raggedright\sffamily 11.~Flexible character accuracy}} & \rotatebox{90}{\parbox{3.3cm}{\raggedright\sffamily 12.~ISRI word accuracy}} & \rotatebox{90}{\parbox{3.3cm}{\raggedright\sffamily 13.~NID (Indel similarity)}} & \rotatebox{90}{\parbox{3.3cm}{\raggedright\sffamily 14.~Text 1 $-$ NED (MGAM matching)}} & \rotatebox{90}{\parbox{3.3cm}{\raggedright\sffamily 15.~Next: reading-order component}} & \rotatebox{90}{\parbox{3.3cm}{\raggedright\sffamily 16.~Word order (Kendall)}} & \rotatebox{90}{\parbox{3.3cm}{\raggedright\sffamily 17.~Reading order 1 $-$ ED}} & \rotatebox{90}{\parbox{3.3cm}{\raggedright\sffamily 18.~BLEU-4}} & \rotatebox{90}{\parbox{3.3cm}{\raggedright\sffamily 19.~Heuristic score}} & \rotatebox{90}{\parbox{3.3cm}{\raggedright\sffamily 20.~Word accuracy (1 $-$ WER)}} & \rotatebox{90}{\parbox{3.3cm}{\raggedright\sffamily 21.~Word edit distance}} & \rotatebox{90}{\parbox{3.3cm}{\raggedright\sffamily 22.~METEOR}} & \rotatebox{90}{\parbox{3.3cm}{\raggedright\sffamily 23.~chrF (character n-gram F)}} & \rotatebox{90}{\parbox{3.3cm}{\raggedright\sffamily 24.~Unique-word F1}} & \rotatebox{90}{\parbox{3.3cm}{\raggedright\sffamily 25.~Tokens found}} & \rotatebox{90}{\parbox{3.3cm}{\raggedright\sffamily 26.~Word F1 (multiset)}} & \rotatebox{90}{\parbox{3.3cm}{\raggedright\sffamily 27.~Full-page OCR score}} & \rotatebox{90}{\parbox{3.3cm}{\raggedright\sffamily 28.~1 $-$ tokens added}} & \rotatebox{90}{\parbox{3.3cm}{\raggedright\sffamily 29.~1 $-$ edit distance}} & \rotatebox{90}{\parbox{3.3cm}{\raggedright\sffamily 30.~Character accuracy (1 $-$ CER)}} & \rotatebox{90}{\parbox{3.3cm}{\raggedright\sffamily 31.~Document parsing STEDS}} \\
\midrule
\endhead
\bottomrule
\endlastfoot
\RCband{33}{CRITICAL} \\
\end{longtable}}
\end{landscape}

\begin{landscape}
{\RClongsetup\scriptsize\setlength{\tabcolsep}{0pt}\renewcommand{\arraystretch}{0.92}
\begin{longtable}{>{\raggedright\arraybackslash}p{5.2cm}r@{\hspace{4pt}}*{17}{>{\centering\arraybackslash}p{0.62cm}}}
\caption{Formatting: responses of 17 metrics to 55 perturbations on 4 sample pages. \ding{51}: target met. $\sim$: partial response. \ding{55}: target missed. Sev.: severity rating.}\label{tab:grid-formatting}\\
\toprule
\RChead{Perturbation} & \RChead{Sev.} & \rotatebox{90}{\parbox{3.3cm}{\raggedright\sffamily 1.~\textbf{\pbtwo{} Formatting}}} & \rotatebox{90}{\parbox{3.3cm}{\raggedright\sffamily 2.~formatting\_\allowbreak{}v2 cap only}} & \rotatebox{90}{\parbox{3.3cm}{\raggedright\sffamily 3.~formatting\_\allowbreak{}v2 proportional only}} & \rotatebox{90}{\parbox{3.3cm}{\raggedright\sffamily 4.~Style-token F1}} & \rotatebox{90}{\parbox{3.3cm}{\raggedright\sffamily 5.~Heading F1, exact level}} & \rotatebox{90}{\parbox{3.3cm}{\raggedright\sffamily 6.~Normalized title accuracy}} & \rotatebox{90}{\parbox{3.3cm}{\raggedright\sffamily 7.~FormatTest pass rate}} & \rotatebox{90}{\parbox{3.3cm}{\raggedright\sffamily 8.~STEDS}} & \rotatebox{90}{\parbox{3.3cm}{\raggedright\sffamily 9.~Normalized text styling}} & \rotatebox{90}{\parbox{3.3cm}{\raggedright\sffamily 10.~Title F1 / level accuracy harmonic mean}} & \rotatebox{90}{\parbox{3.3cm}{\raggedright\sffamily 11.~All formatting rules pass rate}} & \rotatebox{90}{\parbox{3.3cm}{\raggedright\sffamily 12.~Heading TEDS}} & \rotatebox{90}{\parbox{3.3cm}{\raggedright\sffamily 13.~Semantic Formatting}} & \rotatebox{90}{\parbox{3.3cm}{\raggedright\sffamily 14.~MHS heading tree}} & \rotatebox{90}{\parbox{3.3cm}{\raggedright\sffamily 15.~Text 1 $-$ NED (cleaned)}} & \rotatebox{90}{\parbox{3.3cm}{\raggedright\sffamily 16.~Tag-sequence 1 $-$ NED}} & \rotatebox{90}{\parbox{3.3cm}{\raggedright\sffamily 17.~Raw Markdown 1 $-$ NED}} \\
\midrule
\endfirsthead
\multicolumn{19}{l}{\scriptsize\sffamily\color{LIMuted}Formatting test suite, continued}\\
\toprule
\RChead{Perturbation} & \RChead{Sev.} & \rotatebox{90}{\parbox{3.3cm}{\raggedright\sffamily 1.~\textbf{\pbtwo{} Formatting}}} & \rotatebox{90}{\parbox{3.3cm}{\raggedright\sffamily 2.~formatting\_\allowbreak{}v2 cap only}} & \rotatebox{90}{\parbox{3.3cm}{\raggedright\sffamily 3.~formatting\_\allowbreak{}v2 proportional only}} & \rotatebox{90}{\parbox{3.3cm}{\raggedright\sffamily 4.~Style-token F1}} & \rotatebox{90}{\parbox{3.3cm}{\raggedright\sffamily 5.~Heading F1, exact level}} & \rotatebox{90}{\parbox{3.3cm}{\raggedright\sffamily 6.~Normalized title accuracy}} & \rotatebox{90}{\parbox{3.3cm}{\raggedright\sffamily 7.~FormatTest pass rate}} & \rotatebox{90}{\parbox{3.3cm}{\raggedright\sffamily 8.~STEDS}} & \rotatebox{90}{\parbox{3.3cm}{\raggedright\sffamily 9.~Normalized text styling}} & \rotatebox{90}{\parbox{3.3cm}{\raggedright\sffamily 10.~Title F1 / level accuracy harmonic mean}} & \rotatebox{90}{\parbox{3.3cm}{\raggedright\sffamily 11.~All formatting rules pass rate}} & \rotatebox{90}{\parbox{3.3cm}{\raggedright\sffamily 12.~Heading TEDS}} & \rotatebox{90}{\parbox{3.3cm}{\raggedright\sffamily 13.~Semantic Formatting}} & \rotatebox{90}{\parbox{3.3cm}{\raggedright\sffamily 14.~MHS heading tree}} & \rotatebox{90}{\parbox{3.3cm}{\raggedright\sffamily 15.~Text 1 $-$ NED (cleaned)}} & \rotatebox{90}{\parbox{3.3cm}{\raggedright\sffamily 16.~Tag-sequence 1 $-$ NED}} & \rotatebox{90}{\parbox{3.3cm}{\raggedright\sffamily 17.~Raw Markdown 1 $-$ NED}} \\
\midrule
\endhead
\bottomrule
\endlastfoot
\RCband{19}{CRITICAL} \\
Strikethrough lost & 90 & \textcolor{sevHarm}{\ding{51}} & \textcolor{sevHarm}{\ding{51}} & \textcolor{gridWeak}{$\sim$} & \textcolor{gridWeak}{$\sim$} & \textcolor{sevCrit}{\ding{55}} & \textcolor{sevCrit}{\ding{55}} & \textcolor{sevCrit}{\ding{55}} & \textcolor{sevCrit}{\ding{55}} & \textcolor{sevHarm}{\ding{51}} & \textcolor{sevCrit}{\ding{55}} & \textcolor{sevHarm}{\ding{51}} & \textcolor{sevCrit}{\ding{55}} & \textcolor{gridWeak}{$\sim$} & \textcolor{sevCrit}{\ding{55}} & \textcolor{sevCrit}{\ding{55}} & \textcolor{sevCrit}{\ding{55}} & \textcolor{gridWeak}{$\sim$} \\
Exponent lost (10$^9$ $\to$ 109) & 90 & \textcolor{sevHarm}{\ding{51}} & \textcolor{sevHarm}{\ding{51}} & \textcolor{gridWeak}{$\sim$} & \textcolor{gridWeak}{$\sim$} & \textcolor{sevCrit}{\ding{55}} & \textcolor{sevCrit}{\ding{55}} & \textcolor{sevCrit}{\ding{55}} & \textcolor{sevCrit}{\ding{55}} & \textcolor{gridWeak}{$\sim$} & \textcolor{sevCrit}{\ding{55}} & \textcolor{gridWeak}{$\sim$} & \textcolor{sevCrit}{\ding{55}} & \textcolor{gridWeak}{$\sim$} & \textcolor{gridWeak}{$\sim$} & \textcolor{gridWeak}{$\sim$} & \textcolor{gridWeak}{$\sim$} & \textcolor{gridWeak}{$\sim$} \\
Live text struck through & 90 & \textcolor{sevHarm}{\ding{51}} & \textcolor{sevHarm}{\ding{51}} & \textcolor{gridWeak}{$\sim$} & \textcolor{gridWeak}{$\sim$} & \textcolor{sevCrit}{\ding{55}} & \textcolor{sevCrit}{\ding{55}} & \textcolor{sevCrit}{\ding{55}} & \textcolor{sevCrit}{\ding{55}} & \textcolor{sevCrit}{\ding{55}} & \textcolor{sevCrit}{\ding{55}} & \textcolor{sevCrit}{\ding{55}} & \textcolor{sevCrit}{\ding{55}} & \textcolor{sevCrit}{\ding{55}} & \textcolor{sevCrit}{\ding{55}} & \textcolor{sevCrit}{\ding{55}} & \textcolor{sevCrit}{\ding{55}} & \textcolor{sevCrit}{\ding{55}} \\
\end{longtable}}
\end{landscape}

\begin{landscape}
{\RClongsetup\scriptsize\setlength{\tabcolsep}{0pt}\renewcommand{\arraystretch}{0.92}
\begin{longtable}{>{\raggedright\arraybackslash}p{5.2cm}r@{\hspace{4pt}}*{28}{>{\centering\arraybackslash}p{0.54cm}}}
\caption{Tables: responses of 28 metrics to 62 perturbations on 5 sample pages. \ding{51}: target met. $\sim$: partial response. \ding{55}: target missed. Sev.: severity rating.}\label{tab:grid-tables}\\
\toprule
\RChead{Perturbation} & \RChead{Sev.} & \rotatebox{90}{\parbox{3.3cm}{\raggedright\sffamily 1.~\textbf{\pbtwo{} Tables}}} & \rotatebox{90}{\parbox{3.3cm}{\raggedright\sffamily 2.~Table test pass rate}} & \rotatebox{90}{\parbox{3.3cm}{\raggedright\sffamily 3.~TableRecordMatch}} & \rotatebox{90}{\parbox{3.3cm}{\raggedright\sffamily 4.~TEDS-S}} & \rotatebox{90}{\parbox{3.3cm}{\raggedright\sffamily 5.~GriTS-Top}} & \rotatebox{90}{\parbox{3.3cm}{\raggedright\sffamily 6.~GriTS-Con}} & \rotatebox{90}{\parbox{3.3cm}{\raggedright\sffamily 7.~Table size match}} & \rotatebox{90}{\parbox{3.3cm}{\raggedright\sffamily 8.~T-LAG (cell-neighbour graph F1)}} & \rotatebox{90}{\parbox{3.3cm}{\raggedright\sffamily 9.~GTRM composite}} & \rotatebox{90}{\parbox{3.3cm}{\raggedright\sffamily 10.~Position accuracy (text adaptation)}} & \rotatebox{90}{\parbox{3.3cm}{\raggedright\sffamily 11.~Structure BLEU-4}} & \rotatebox{90}{\parbox{3.3cm}{\raggedright\sffamily 12.~Merged-cell F1 (MCD)}} & \rotatebox{90}{\parbox{3.3cm}{\raggedright\sffamily 13.~Composite structure acc}} & \rotatebox{90}{\parbox{3.3cm}{\raggedright\sffamily 14.~AccCon (exact table)}} & \rotatebox{90}{\parbox{3.3cm}{\raggedright\sffamily 15.~Adjacency F1}} & \rotatebox{90}{\parbox{3.3cm}{\raggedright\sffamily 16.~JSON accuracy}} & \rotatebox{90}{\parbox{3.3cm}{\raggedright\sffamily 17.~Cell accuracy (TCE)}} & \rotatebox{90}{\parbox{3.3cm}{\raggedright\sffamily 18.~Row and column F1 (RCE)}} & \rotatebox{90}{\parbox{3.3cm}{\raggedright\sffamily 19.~Relation F1, exact}} & \rotatebox{90}{\parbox{3.3cm}{\raggedright\sffamily 20.~TEDS (v1.6 normalization)}} & \rotatebox{90}{\parbox{3.3cm}{\raggedright\sffamily 21.~TEDS (normalized)}} & \rotatebox{90}{\parbox{3.3cm}{\raggedright\sffamily 22.~Shifted content accuracy}} & \rotatebox{90}{\parbox{3.3cm}{\raggedright\sffamily 23.~GriTS-Con}} & \rotatebox{90}{\parbox{3.3cm}{\raggedright\sffamily 24.~TEDS}} & \rotatebox{90}{\parbox{3.3cm}{\raggedright\sffamily 25.~Relation F1, chrF}} & \rotatebox{90}{\parbox{3.3cm}{\raggedright\sffamily 26.~1 $-$ table edit distance}} & \rotatebox{90}{\parbox{3.3cm}{\raggedright\sffamily 27.~Alignment score}} & \rotatebox{90}{\parbox{3.3cm}{\raggedright\sffamily 28.~Bigram Jaccard}} \\
\midrule
\endfirsthead
\multicolumn{30}{l}{\scriptsize\sffamily\color{LIMuted}Tables test suite, continued}\\
\toprule
\RChead{Perturbation} & \RChead{Sev.} & \rotatebox{90}{\parbox{3.3cm}{\raggedright\sffamily 1.~\textbf{\pbtwo{} Tables}}} & \rotatebox{90}{\parbox{3.3cm}{\raggedright\sffamily 2.~Table test pass rate}} & \rotatebox{90}{\parbox{3.3cm}{\raggedright\sffamily 3.~TableRecordMatch}} & \rotatebox{90}{\parbox{3.3cm}{\raggedright\sffamily 4.~TEDS-S}} & \rotatebox{90}{\parbox{3.3cm}{\raggedright\sffamily 5.~GriTS-Top}} & \rotatebox{90}{\parbox{3.3cm}{\raggedright\sffamily 6.~GriTS-Con}} & \rotatebox{90}{\parbox{3.3cm}{\raggedright\sffamily 7.~Table size match}} & \rotatebox{90}{\parbox{3.3cm}{\raggedright\sffamily 8.~T-LAG (cell-neighbour graph F1)}} & \rotatebox{90}{\parbox{3.3cm}{\raggedright\sffamily 9.~GTRM composite}} & \rotatebox{90}{\parbox{3.3cm}{\raggedright\sffamily 10.~Position accuracy (text adaptation)}} & \rotatebox{90}{\parbox{3.3cm}{\raggedright\sffamily 11.~Structure BLEU-4}} & \rotatebox{90}{\parbox{3.3cm}{\raggedright\sffamily 12.~Merged-cell F1 (MCD)}} & \rotatebox{90}{\parbox{3.3cm}{\raggedright\sffamily 13.~Composite structure acc}} & \rotatebox{90}{\parbox{3.3cm}{\raggedright\sffamily 14.~AccCon (exact table)}} & \rotatebox{90}{\parbox{3.3cm}{\raggedright\sffamily 15.~Adjacency F1}} & \rotatebox{90}{\parbox{3.3cm}{\raggedright\sffamily 16.~JSON accuracy}} & \rotatebox{90}{\parbox{3.3cm}{\raggedright\sffamily 17.~Cell accuracy (TCE)}} & \rotatebox{90}{\parbox{3.3cm}{\raggedright\sffamily 18.~Row and column F1 (RCE)}} & \rotatebox{90}{\parbox{3.3cm}{\raggedright\sffamily 19.~Relation F1, exact}} & \rotatebox{90}{\parbox{3.3cm}{\raggedright\sffamily 20.~TEDS (v1.6 normalization)}} & \rotatebox{90}{\parbox{3.3cm}{\raggedright\sffamily 21.~TEDS (normalized)}} & \rotatebox{90}{\parbox{3.3cm}{\raggedright\sffamily 22.~Shifted content accuracy}} & \rotatebox{90}{\parbox{3.3cm}{\raggedright\sffamily 23.~GriTS-Con}} & \rotatebox{90}{\parbox{3.3cm}{\raggedright\sffamily 24.~TEDS}} & \rotatebox{90}{\parbox{3.3cm}{\raggedright\sffamily 25.~Relation F1, chrF}} & \rotatebox{90}{\parbox{3.3cm}{\raggedright\sffamily 26.~1 $-$ table edit distance}} & \rotatebox{90}{\parbox{3.3cm}{\raggedright\sffamily 27.~Alignment score}} & \rotatebox{90}{\parbox{3.3cm}{\raggedright\sffamily 28.~Bigram Jaccard}} \\
\midrule
\endhead
\bottomrule
\endlastfoot
\RCband{30}{CRITICAL} \\
Table missing entirely & 90 & \textcolor{sevHarm}{\ding{51}} & \textcolor{sevHarm}{\ding{51}} & \textcolor{sevHarm}{\ding{51}} & \textcolor{sevHarm}{\ding{51}} & \textcolor{sevHarm}{\ding{51}} & \textcolor{sevHarm}{\ding{51}} & \textcolor{sevHarm}{\ding{51}} & \textcolor{sevHarm}{\ding{51}} & \textcolor{sevHarm}{\ding{51}} & \textcolor{sevHarm}{\ding{51}} & \textcolor{sevHarm}{\ding{51}} & \textcolor{sevHarm}{\ding{51}} & \textcolor{sevHarm}{\ding{51}} & \textcolor{sevHarm}{\ding{51}} & \textcolor{sevHarm}{\ding{51}} & \textcolor{sevHarm}{\ding{51}} & \textcolor{sevHarm}{\ding{51}} & \textcolor{sevHarm}{\ding{51}} & \textcolor{sevHarm}{\ding{51}} & \textcolor{sevHarm}{\ding{51}} & \textcolor{sevHarm}{\ding{51}} & \textcolor{sevHarm}{\ding{51}} & \textcolor{sevHarm}{\ding{51}} & \textcolor{sevHarm}{\ding{51}} & \textcolor{sevHarm}{\ding{51}} & \textcolor{sevHarm}{\ding{51}} & \textcolor{sevHarm}{\ding{51}} & \textcolor{sevHarm}{\ding{51}} \\
\end{longtable}}
\end{landscape}

\begin{landscape}
{\RClongsetup\scriptsize\setlength{\tabcolsep}{0pt}\renewcommand{\arraystretch}{0.92}
\begin{longtable}{>{\raggedright\arraybackslash}p{5.2cm}r@{\hspace{4pt}}*{20}{>{\centering\arraybackslash}p{0.62cm}}}
\caption{Charts: responses of 20 metrics to 53 perturbations on 5 sample pages. \ding{51}: target met. $\sim$: partial response. \ding{55}: target missed. Sev.: severity rating.}\label{tab:grid-charts}\\
\toprule
\RChead{Perturbation} & \RChead{Sev.} & \rotatebox{90}{\parbox{3.3cm}{\raggedright\sffamily 1.~\textbf{\pbtwo{} Charts}}} & \rotatebox{90}{\parbox{3.3cm}{\raggedright\sffamily 2.~ChartDataPoint pass rate}} & \rotatebox{90}{\parbox{3.3cm}{\raggedright\sffamily 3.~EpiCurveSimilarity}} & \rotatebox{90}{\parbox{3.3cm}{\raggedright\sffamily 4.~1 $-$ A-MAPE}} & \rotatebox{90}{\parbox{3.3cm}{\raggedright\sffamily 5.~SCRM slight}} & \rotatebox{90}{\parbox{3.3cm}{\raggedright\sffamily 6.~SCRM strict}} & \rotatebox{90}{\parbox{3.3cm}{\raggedright\sffamily 7.~SCRM high}} & \rotatebox{90}{\parbox{3.3cm}{\raggedright\sffamily 8.~ChartOCR score}} & \rotatebox{90}{\parbox{3.3cm}{\raggedright\sffamily 9.~RMS-TBE-F1-Sig}} & \rotatebox{90}{\parbox{3.3cm}{\raggedright\sffamily 10.~RMS-TBE-F1}} & \rotatebox{90}{\parbox{3.3cm}{\raggedright\sffamily 11.~Metric 6b}} & \rotatebox{90}{\parbox{3.3cm}{\raggedright\sffamily 12.~Point F1 (2\% box)}} & \rotatebox{90}{\parbox{3.3cm}{\raggedright\sffamily 13.~Normalized mapping similarity}} & \rotatebox{90}{\parbox{3.3cm}{\raggedright\sffamily 14.~RNSS}} & \rotatebox{90}{\parbox{3.3cm}{\raggedright\sffamily 15.~(1 $-$ nMAE) $\times$ coverage}} & \rotatebox{90}{\parbox{3.3cm}{\raggedright\sffamily 16.~Relaxed accuracy (QA by lookup)}} & \rotatebox{90}{\parbox{3.3cm}{\raggedright\sffamily 17.~RMS F1}} & \rotatebox{90}{\parbox{3.3cm}{\raggedright\sffamily 18.~Row IoU (medium tolerance)}} & \rotatebox{90}{\parbox{3.3cm}{\raggedright\sffamily 19.~0.5 RMS + 0.5 TEDS$'$}} & \rotatebox{90}{\parbox{3.3cm}{\raggedright\sffamily 20.~Chart parsing TEDS}} \\
\midrule
\endfirsthead
\multicolumn{22}{l}{\scriptsize\sffamily\color{LIMuted}Charts test suite, continued}\\
\toprule
\RChead{Perturbation} & \RChead{Sev.} & \rotatebox{90}{\parbox{3.3cm}{\raggedright\sffamily 1.~\textbf{\pbtwo{} Charts}}} & \rotatebox{90}{\parbox{3.3cm}{\raggedright\sffamily 2.~ChartDataPoint pass rate}} & \rotatebox{90}{\parbox{3.3cm}{\raggedright\sffamily 3.~EpiCurveSimilarity}} & \rotatebox{90}{\parbox{3.3cm}{\raggedright\sffamily 4.~1 $-$ A-MAPE}} & \rotatebox{90}{\parbox{3.3cm}{\raggedright\sffamily 5.~SCRM slight}} & \rotatebox{90}{\parbox{3.3cm}{\raggedright\sffamily 6.~SCRM strict}} & \rotatebox{90}{\parbox{3.3cm}{\raggedright\sffamily 7.~SCRM high}} & \rotatebox{90}{\parbox{3.3cm}{\raggedright\sffamily 8.~ChartOCR score}} & \rotatebox{90}{\parbox{3.3cm}{\raggedright\sffamily 9.~RMS-TBE-F1-Sig}} & \rotatebox{90}{\parbox{3.3cm}{\raggedright\sffamily 10.~RMS-TBE-F1}} & \rotatebox{90}{\parbox{3.3cm}{\raggedright\sffamily 11.~Metric 6b}} & \rotatebox{90}{\parbox{3.3cm}{\raggedright\sffamily 12.~Point F1 (2\% box)}} & \rotatebox{90}{\parbox{3.3cm}{\raggedright\sffamily 13.~Normalized mapping similarity}} & \rotatebox{90}{\parbox{3.3cm}{\raggedright\sffamily 14.~RNSS}} & \rotatebox{90}{\parbox{3.3cm}{\raggedright\sffamily 15.~(1 $-$ nMAE) $\times$ coverage}} & \rotatebox{90}{\parbox{3.3cm}{\raggedright\sffamily 16.~Relaxed accuracy (QA by lookup)}} & \rotatebox{90}{\parbox{3.3cm}{\raggedright\sffamily 17.~RMS F1}} & \rotatebox{90}{\parbox{3.3cm}{\raggedright\sffamily 18.~Row IoU (medium tolerance)}} & \rotatebox{90}{\parbox{3.3cm}{\raggedright\sffamily 19.~0.5 RMS + 0.5 TEDS$'$}} & \rotatebox{90}{\parbox{3.3cm}{\raggedright\sffamily 20.~Chart parsing TEDS}} \\
\midrule
\endhead
\bottomrule
\endlastfoot
\RCband{22}{CRITICAL} \\
Scale lost & 90 & \textcolor{sevHarm}{\ding{51}} & \textcolor{sevHarm}{\ding{51}} & \textcolor{sevHarm}{\ding{51}} & \textcolor{sevHarm}{\ding{51}} & \textcolor{sevHarm}{\ding{51}} & \textcolor{sevHarm}{\ding{51}} & \textcolor{sevHarm}{\ding{51}} & \textcolor{sevHarm}{\ding{51}} & \textcolor{sevHarm}{\ding{51}} & \textcolor{sevHarm}{\ding{51}} & \textcolor{sevHarm}{\ding{51}} & \textcolor{sevHarm}{\ding{51}} & \textcolor{sevHarm}{\ding{51}} & \textcolor{sevHarm}{\ding{51}} & \textcolor{sevHarm}{\ding{51}} & \textcolor{sevHarm}{\ding{51}} & \textcolor{sevHarm}{\ding{51}} & \textcolor{sevHarm}{\ding{51}} & \textcolor{sevHarm}{\ding{51}} & \textcolor{sevHarm}{\ding{51}} \\
Order of magnitude off & 90 & \textcolor{sevHarm}{\ding{51}} & \textcolor{sevHarm}{\ding{51}} & \textcolor{sevHarm}{\ding{51}} & \textcolor{sevHarm}{\ding{51}} & \textcolor{sevHarm}{\ding{51}} & \textcolor{sevHarm}{\ding{51}} & \textcolor{sevHarm}{\ding{51}} & \textcolor{sevHarm}{\ding{51}} & \textcolor{sevHarm}{\ding{51}} & \textcolor{sevHarm}{\ding{51}} & \textcolor{gridWeak}{$\sim$} & \textcolor{sevHarm}{\ding{51}} & \textcolor{sevHarm}{\ding{51}} & \textcolor{sevHarm}{\ding{51}} & \textcolor{sevHarm}{\ding{51}} & \textcolor{sevHarm}{\ding{51}} & \textcolor{sevHarm}{\ding{51}} & \textcolor{sevHarm}{\ding{51}} & \textcolor{sevHarm}{\ding{51}} & \textcolor{gridWeak}{$\sim$} \\
\end{longtable}}
\end{landscape}

\begin{landscape}
{\RClongsetup\scriptsize\setlength{\tabcolsep}{0pt}\renewcommand{\arraystretch}{0.92}
\begin{longtable}{>{\raggedright\arraybackslash}p{5.2cm}r@{\hspace{4pt}}*{14}{>{\centering\arraybackslash}p{0.62cm}}}
\caption{Forms: responses of 14 metrics to 71 perturbations on 6 sample pages. \ding{51}: target met. $\sim$: partial response. \ding{55}: target missed. Sev.: severity rating.}\label{tab:grid-forms}\\
\toprule
\RChead{Perturbation} & \RChead{Sev.} & \rotatebox{90}{\parbox{3.3cm}{\raggedright\sffamily 1.~\textbf{\pbtwo{} Forms}}} & \rotatebox{90}{\parbox{3.3cm}{\raggedright\sffamily 2.~forms\_\allowbreak{}v2 cap only}} & \rotatebox{90}{\parbox{3.3cm}{\raggedright\sffamily 3.~forms\_\allowbreak{}v2 proportional only}} & \rotatebox{90}{\parbox{3.3cm}{\raggedright\sffamily 4.~Value presence}} & \rotatebox{90}{\parbox{3.3cm}{\raggedright\sffamily 5.~Field F1 (exact)}} & \rotatebox{90}{\parbox{3.3cm}{\raggedright\sffamily 6.~ANLS per field}} & \rotatebox{90}{\parbox{3.3cm}{\raggedright\sffamily 7.~1 - NED per field}} & \rotatebox{90}{\parbox{3.3cm}{\raggedright\sffamily 8.~Entity-linking F1}} & \rotatebox{90}{\parbox{3.3cm}{\raggedright\sffamily 9.~Checkbox accuracy}} & \rotatebox{90}{\parbox{3.3cm}{\raggedright\sffamily 10.~Page text 1 - NED}} & \rotatebox{90}{\parbox{3.3cm}{\raggedright\sffamily 11.~rule\_\allowbreak{}form\_\allowbreak{}field\_\allowbreak{}pass\_\allowbreak{}rate}} & \rotatebox{90}{\parbox{3.3cm}{\raggedright\sffamily 12.~form\_\allowbreak{}kv\_\allowbreak{}md\_\allowbreak{}core\_\allowbreak{}kv\_\allowbreak{}f1}} & \rotatebox{90}{\parbox{3.3cm}{\raggedright\sffamily 13.~form\_\allowbreak{}bench\_\allowbreak{}text\_\allowbreak{}value\_\allowbreak{}f1}} & \rotatebox{90}{\parbox{3.3cm}{\raggedright\sffamily 14.~form\_\allowbreak{}bench\_\allowbreak{}text\_\allowbreak{}attribution\_\allowbreak{}f1}} \\
\midrule
\endfirsthead
\multicolumn{16}{l}{\scriptsize\sffamily\color{LIMuted}Forms test suite, continued}\\
\toprule
\RChead{Perturbation} & \RChead{Sev.} & \rotatebox{90}{\parbox{3.3cm}{\raggedright\sffamily 1.~\textbf{\pbtwo{} Forms}}} & \rotatebox{90}{\parbox{3.3cm}{\raggedright\sffamily 2.~forms\_\allowbreak{}v2 cap only}} & \rotatebox{90}{\parbox{3.3cm}{\raggedright\sffamily 3.~forms\_\allowbreak{}v2 proportional only}} & \rotatebox{90}{\parbox{3.3cm}{\raggedright\sffamily 4.~Value presence}} & \rotatebox{90}{\parbox{3.3cm}{\raggedright\sffamily 5.~Field F1 (exact)}} & \rotatebox{90}{\parbox{3.3cm}{\raggedright\sffamily 6.~ANLS per field}} & \rotatebox{90}{\parbox{3.3cm}{\raggedright\sffamily 7.~1 - NED per field}} & \rotatebox{90}{\parbox{3.3cm}{\raggedright\sffamily 8.~Entity-linking F1}} & \rotatebox{90}{\parbox{3.3cm}{\raggedright\sffamily 9.~Checkbox accuracy}} & \rotatebox{90}{\parbox{3.3cm}{\raggedright\sffamily 10.~Page text 1 - NED}} & \rotatebox{90}{\parbox{3.3cm}{\raggedright\sffamily 11.~rule\_\allowbreak{}form\_\allowbreak{}field\_\allowbreak{}pass\_\allowbreak{}rate}} & \rotatebox{90}{\parbox{3.3cm}{\raggedright\sffamily 12.~form\_\allowbreak{}kv\_\allowbreak{}md\_\allowbreak{}core\_\allowbreak{}kv\_\allowbreak{}f1}} & \rotatebox{90}{\parbox{3.3cm}{\raggedright\sffamily 13.~form\_\allowbreak{}bench\_\allowbreak{}text\_\allowbreak{}value\_\allowbreak{}f1}} & \rotatebox{90}{\parbox{3.3cm}{\raggedright\sffamily 14.~form\_\allowbreak{}bench\_\allowbreak{}text\_\allowbreak{}attribution\_\allowbreak{}f1}} \\
\midrule
\endhead
\bottomrule
\endlastfoot
\RCband{16}{CRITICAL} \\
Digit misread in a value & 90 & \textcolor{sevHarm}{\ding{51}} & \textcolor{sevHarm}{\ding{51}} & \textcolor{gridWeak}{$\sim$} & \textcolor{sevHarm}{\ding{51}} & \textcolor{gridWeak}{$\sim$} & \textcolor{sevCrit}{\ding{55}} & \textcolor{sevCrit}{\ding{55}} & \textcolor{gridWeak}{$\sim$} & \textcolor{sevCrit}{\ding{55}} & \textcolor{sevCrit}{\ding{55}} & \textcolor{gridWeak}{$\sim$} & \textcolor{gridWeak}{$\sim$} & \textcolor{gridWeak}{$\sim$} & \textcolor{gridWeak}{$\sim$} \\
Decimal point lost in an amount & 90 & \textcolor{sevHarm}{\ding{51}} & \textcolor{sevHarm}{\ding{51}} & \textcolor{gridWeak}{$\sim$} & \textcolor{gridWeak}{$\sim$} & \textcolor{gridWeak}{$\sim$} & \textcolor{sevCrit}{\ding{55}} & \textcolor{sevCrit}{\ding{55}} & \textcolor{gridWeak}{$\sim$} & \textcolor{sevCrit}{\ding{55}} & \textcolor{gridWeak}{$\sim$} & \textcolor{sevCrit}{\ding{55}} & \textcolor{gridWeak}{$\sim$} & \textcolor{gridWeak}{$\sim$} & \textcolor{gridWeak}{$\sim$} \\
Two digits of an ID transposed & 90 & \textcolor{sevHarm}{\ding{51}} & \textcolor{sevHarm}{\ding{51}} & \textcolor{gridWeak}{$\sim$} & \textcolor{gridWeak}{$\sim$} & \textcolor{gridWeak}{$\sim$} & \textcolor{sevCrit}{\ding{55}} & \textcolor{sevCrit}{\ding{55}} & \textcolor{gridWeak}{$\sim$} & \textcolor{sevCrit}{\ding{55}} & \textcolor{sevCrit}{\ding{55}} & \textcolor{gridWeak}{$\sim$} & \textcolor{gridWeak}{$\sim$} & \textcolor{gridWeak}{$\sim$} & \textcolor{gridWeak}{$\sim$} \\
Values of two fields swapped & 90 & \textcolor{sevHarm}{\ding{51}} & \textcolor{sevHarm}{\ding{51}} & \textcolor{gridWeak}{$\sim$} & \textcolor{sevCrit}{\ding{55}} & \textcolor{sevHarm}{\ding{51}} & \textcolor{gridWeak}{$\sim$} & \textcolor{gridWeak}{$\sim$} & \textcolor{gridWeak}{$\sim$} & \textcolor{sevCrit}{\ding{55}} & \textcolor{gridWeak}{$\sim$} & \textcolor{gridWeak}{$\sim$} & \textcolor{gridWeak}{$\sim$} & \textcolor{gridWeak}{$\sim$} & \textcolor{gridWeak}{$\sim$} \\
\end{longtable}}
\end{landscape}

{\RClongsetup\scriptsize\setlength{\tabcolsep}{0pt}\renewcommand{\arraystretch}{0.92}
\begin{longtable}{>{\raggedright\arraybackslash}p{6.5cm}r@{\hspace{4pt}}*{4}{>{\centering\arraybackslash}p{1.30cm}}}
\caption{Visual grounding: responses of 4 metrics to 38 perturbations on one synthetic page. \ding{51}: target met. $\sim$: partial response. \ding{55}: target missed. Sev.: severity rating.}\label{tab:grid-layout}\\
\toprule
\RChead{Perturbation} & \RChead{Sev.} & \rotatebox{90}{\parbox{2.2cm}{\raggedright\sffamily 1.~\textbf{\pbtwo{} Visual grounding}}} & \rotatebox{90}{\parbox{2.2cm}{\raggedright\sffamily 2.~EPR v3 draft}} & \rotatebox{90}{\parbox{2.2cm}{\raggedright\sffamily 3.~EPR v1, no text gate}} & \rotatebox{90}{\parbox{2.2cm}{\raggedright\sffamily 4.~EPR v1}} \\
\midrule
\endfirsthead
\multicolumn{6}{l}{\scriptsize\sffamily\color{LIMuted}Visual grounding test suite, continued}\\
\toprule
\RChead{Perturbation} & \RChead{Sev.} & \rotatebox{90}{\parbox{2.2cm}{\raggedright\sffamily 1.~\textbf{\pbtwo{} Visual grounding}}} & \rotatebox{90}{\parbox{2.2cm}{\raggedright\sffamily 2.~EPR v3 draft}} & \rotatebox{90}{\parbox{2.2cm}{\raggedright\sffamily 3.~EPR v1, no text gate}} & \rotatebox{90}{\parbox{2.2cm}{\raggedright\sffamily 4.~EPR v1}} \\
\midrule
\endhead
\bottomrule
\endlastfoot
\RCband{6}{CRITICAL} \\
One page-wide box & 90 & \textcolor{sevHarm}{\ding{51}} & \textcolor{sevHarm}{\ding{51}} & \textcolor{sevHarm}{\ding{51}} & \textcolor{sevHarm}{\ding{51}} \\
Biggest paragraph missed & 90 & \textcolor{sevHarm}{\ding{51}} & \textcolor{sevHarm}{\ding{51}} & \textcolor{gridWeak}{$\sim$} & \textcolor{gridWeak}{$\sim$} \\
Five paragraphs in one box & 90 & \textcolor{sevHarm}{\ding{51}} & \textcolor{sevHarm}{\ding{51}} & \textcolor{sevHarm}{\ding{51}} & \textcolor{sevHarm}{\ding{51}} \\
Right box, wrong text & 90 & \textcolor{sevHarm}{\ding{51}} & \textcolor{sevHarm}{\ding{51}} & \textcolor{sevCrit}{\ding{55}} & \textcolor{gridWeak}{$\sim$} \\
Coarse blocks & 90 & \textcolor{sevHarm}{\ding{51}} & \textcolor{sevHarm}{\ding{51}} & \textcolor{sevHarm}{\ding{51}} & \textcolor{sevHarm}{\ding{51}} \\
Loose boxes (IoU about 0.45) & 90 & \textcolor{sevHarm}{\ding{51}} & \textcolor{sevHarm}{\ding{51}} & \textcolor{sevCrit}{\ding{55}} & \textcolor{sevCrit}{\ding{55}} \\
Japanese paragraph garbled & 90 & \textcolor{gridWeak}{$\sim$} & \textcolor{gridWeak}{$\sim$} & \textcolor{sevCrit}{\ding{55}} & \textcolor{gridWeak}{$\sim$} \\
\end{longtable}}

\FloatBarrier

\clearpage
\section{Parser Evaluation}
\label{app:parser-evaluation}
\subsection{Experimental Setup}
\label{app:experimental-setup}
\label{sec:setup}
\label{sec:protocol}

We use the same evaluation pipeline and scoring code for all systems.

\paragraph{Configurations.}
\label{sec:options}
\label{sec:options-rule}
\label{sec:options-list}
We compute overall scores using the default configuration of each parser tier or mode.
We evaluate optional features, such as chart and form parsing, separately and report their dimension scores alongside the defaults.
We ask frontier VLMs to transcribe each page using the same prompt.
We run open-weight models with their own prompts or pipelines.

\paragraph{Scoring protocol.}
\label{sec:setup-normalization}
\label{sec:setup-scoring}
Providers return outputs in different formats, so we convert them into a common representation before scoring.
We score saved outputs against the page annotations using the \pbtwo{} metrics, then combine page scores into dimension and overall scores (\Cref{sec:score-compose}).
We retry operational failures and flag runs where these failures remain on at least 1\% of the pages.

\paragraph{Cost.}
Cost is the average cost per page over the run, at list prices or usage-based billing as the provider's API reports it.
Document parsers use per-page prices, while frontier VLM costs depend on billed token usage.
The reported VLM costs therefore depend on the benchmark documents and transcription prompt.
We do not report per-page serving costs for self-hosted open-weight models.

\paragraph{Uncertainty.}
We report 95\% confidence intervals using 2{,}000 bootstrap resamples.
We group test cases from the same source document and resample these groups together, using the same draws for every system.
This accounts for related pages when comparing scores.
The intervals reflect variation across sampled documents; because we run each configuration once, they do not capture variation between repeated runs.

\clearpage
\subsection{Severity Caps and Detected Errors}
\label{app:cap-per-system}

\Cref{tab:cap-binds} shows how often the severity cap lowers each system's score below its parsing score (\Cref{sec:results-errors}).
This occurs when $S = c < m(y,r)$.
We compute this rate over scored core pages with an output, across the five dimensions that use a cap.
The table also reports the average number of penalized errors per scored page, grouped by severity.

\begin{table}[!htbp]
\RCsetup
\renewcommand{\RCsep}{\noalign{\vspace{1pt}{\color{LIPanelLine}\hrule height 0.4pt}\vspace{1pt}}}
\begin{tabularx}{\linewidth}{Xcccccc@{\hspace{10pt}}ccc}
\toprule
& \multicolumn{6}{c}{\sffamily\bfseries Pages limited by the cap (\%)} & \multicolumn{3}{c}{\sffamily\bfseries Errors per page} \\
\cmidrule(lr){2-7}\cmidrule(lr){8-10}
\RChead{Pipeline} & \RChead{Text} & \RChead{Fmt} & \RChead{Tab.} & \RChead{Chart} & \RChead{Form} & \RChead{All} & \RChead{Crit.} & \RChead{Maj.} & \RChead{Min.} \\
\midrule
\RCgroup{Document parsers}{}
LlamaParse (Agentic Plus) & 47 & 37 & 21 & 34 & 75 & 42 & 0.63 & 1.85 & 3.84\RCnext
LlamaParse (Agentic) & 54 & 33 & 27 & 59 & 79 & 51 & 1.92 & 4.25 & 4.23\RCnext
LlamaParse (Cost Effective) & 56 & 35 & 31 & 56 & 79 & 52 & 2.65 & 4.77 & 4.24\RCnext
Reducto & 59 & 33 & 39 & 46 & 78 & 51 & 1.36 & 2.75 & 5.28\RCnext
Extend (Parse 2) & 61 & 32 & 41 & 43 & 76 & 51 & 1.50 & 3.32 & 4.95\RCnext
Pulse (Ultra 2) & 51 & 33 & 33 & 49 & 82 & 50 & 1.64 & 3.06 & 4.83\RCnext
Extend (Parse Light) & 60 & 26 & 42 & 43 & 77 & 50 & 1.58 & 3.57 & 5.09\RCnext
Datalab (Accurate) & 52 & 36 & 32 & 42 & 83 & 49 & 1.84 & 3.66 & 4.32\RCnext
Datalab (Balanced) & 51 & 36 & 34 & 45 & 84 & 50 & 1.99 & 3.85 & 4.44\RCnext
Datalab (Fast) & 56 & 38 & 40 & 48 & 84 & 53 & 2.94 & 4.50 & 4.57\RCnext
Databricks AI Parse & 58 & 27 & 44 & 57 & 83 & 54 & 2.83 & 5.66 & 5.58\RCnext
Reducto (r-1) & 50 & 39 & 35 & 9 & 77 & 41 & 1.22 & 1.58 & 4.75\RCnext
Mistral OCR 4.1 (annotation) & 54 & 24 & 40 & 60 & 83 & 53 & 7.13 & 5.13 & 5.01\RCnext
Reducto (r-1-mini) & 54 & 40 & 35 & 0 & 78 & 40 & 1.33 & 1.61 & 4.93\RCnext
LandingAI & 53 & 33 & 40 & 17 & 82 & 44 & 1.55 & 2.30 & 4.67\RCnext
Extend (Parse) & 60 & 26 & 35 & 1 & 77 & 39 & 1.17 & 2.04 & 4.89\RCnext
Mistral OCR 4.1 & 54 & 24 & 40 & 0 & 82 & 39 & 2.02 & 2.61 & 4.97\RCnext
Reducto (non-agentic) & 63 & 27 & 49 & 0 & 78 & 42 & 2.12 & 2.27 & 5.28\RCnext
Azure Document Intelligence (layout) & 61 & 20 & 43 & 0 & 84 & 41 & 1.64 & 2.77 & 5.28\RClast
\RCgroup{Direct vision models}{}
Claude Opus 5.5 & 42 & 30 & 17 & 15 & 78 & 36 & 0.75 & 1.45 & 4.02\RCnext
Gemini 4 argon (low) & 45 & 31 & 23 & 37 & 79 & 43 & 1.26 & 2.38 & 4.12\RCnext
GPT-5.6 terra (low) & 64 & 36 & 55 & 46 & 83 & 57 & 2.28 & 3.84 & 4.59\RCnext
Gemini 3 Flash (minimal) & 42 & 29 & 33 & 50 & 78 & 46 & 2.90 & 3.96 & 4.50\RCnext
Gemini 3.7 Flash (low) & 47 & 30 & 31 & 10 & 82 & 39 & 1.09 & 1.74 & 4.45\RCnext
Gemini 3.8 Flash (low) & 45 & 30 & 32 & 10 & 82 & 39 & 1.02 & 1.75 & 4.43\RCnext
Gemini 3.5 Flash & 41 & 28 & 28 & 11 & 79 & 37 & 1.28 & 1.56 & 4.40\RCnext
Gemini 3.6 Flash & 42 & 29 & 31 & 9 & 79 & 37 & 1.30 & 1.52 & 4.41\RCnext
Muse Ember 1.0 (low) & 48 & 30 & 34 & 29 & 80 & 44 & 1.20 & 2.59 & 4.37\RCnext
GPT-6.1 terra (low) & 48 & 29 & 22 & 13 & 76 & 37 & 0.78 & 1.50 & 4.29\RCnext
Claude Sonnet 5.5 & 47 & 29 & 26 & 7 & 80 & 37 & 0.90 & 1.29 & 4.44\RCnext
GPT-6 luna (low) & 57 & 33 & 40 & 17 & 80 & 45 & 1.38 & 2.09 & 4.44\RCnext
Claude Haiku 5.5 & 47 & 30 & 39 & 2 & 82 & 39 & 1.42 & 1.52 & 4.61\RCnext
GPT-5.6 luna (low) & 59 & 33 & 53 & 14 & 80 & 47 & 2.10 & 2.39 & 4.72\RCnext
Gemini 3.1 Flash-Lite & 48 & 23 & 40 & 3 & 80 & 38 & 1.54 & 2.09 & 5.11\RCnext
Gemini 3.5 Flash-Lite & 56 & 26 & 42 & 3 & 84 & 41 & 1.47 & 2.93 & 5.18\RClast

\bottomrule
\end{tabularx}
\caption{Pages where the cap determines the final score, and mean penalized errors per scored page. All combines the five dimensions with a cap. Fmt scores Formatting on the Text core pages. Error counts are pooled across these five core collections. Visual grounding has no cap.}
\label{tab:cap-binds}
\end{table}

\clearpage
\subsection{Full Results}
\label{app:full-results}

\Cref{tab:main-results-v2,tab:main-results-models} rank document parsers, frontier VLMs and open-weight models using their default configurations.
\Cref{tab:app-datasets} gives the mean page score for each collection.

\paragraph{Reading the scores.}
The overall score is the mean of the six dimension scores.
Each dimension combines collection means using the weights in \Cref{eq:slots}.
We rank configurations using unrounded scores.
A dash marks an unevaluated collection or a dimension or overall score that cannot be computed because a required collection is missing.
Scores in parentheses show an optional feature's dimension score and do not enter the default overall score.
These scores can be lower than the defaults.

\paragraph{Comparing systems.}
\bestoverallname{} leads the document parsers at \bestoverall{}, compared with \bestvlm{} for the best frontier VLM, \bestvlmname{}.
Charts shows the widest spread among document parsers.
Scores near zero usually indicate that the default configuration extracts no chart data.
For \resAdjZeroParser{} of the \resAdjParser{} adjacent parser pairs, the 95\% confidence interval for the score difference includes zero.
These pairs are not clearly separated by the sampled documents.
See \Cref{sec:leaderboard} for the main findings.

Frontier VLMs use our shared transcription prompt, while open-weight models use their own prompts or pipelines.
Differences between these groups therefore reflect both the models and how we run them.
The labels (low) and (minimal) specify reasoning settings.
We omit olmOCR 2 and Xiaomi-OCR-0 from the ranked table because they return no element boxes and have no overall score.
We do not report serving costs for self-hosted models.

\begin{table}[!htbp]
\RCsetup\setlength{\tabcolsep}{3pt}%
\renewcommand{\RCsep}{\noalign{\vspace{1.4pt}{\color{LIPanelLine}\hrule height 0.4pt}\vspace{1.4pt}}}%
\begin{tabularx}{\linewidth}{r>{\raggedright\arraybackslash}Xcccccccc}
\toprule
\RChead{\#} & \RChead{Pipeline} & \RChead{\pbtwo{} score} & \RChead{Text} & \RChead{Format.} & \RChead{Tables} & \RChead{Charts} & \RChead{\shortstack{Visual\\grounding}} & \RChead{Forms} & \RChead{\$/page} \\
\midrule
\RCgroup{Document parsers}{ranked by \pbtwo{} score; best score per column in bold}
1 & LlamaParse (Agentic Plus) & \textbf{90.1} & 90.3 & \textbf{93.6} & \textbf{94.6} & \textbf{92.6} & 87.5 & \textbf{81.8}\,{\scriptsize(82.9)} & 0.028\RCnext
2 & LlamaParse (Agentic) & 86.0 & 87.0 & 92.7 & 89.9 & 77.4\,{\scriptsize(87.1)} & 88.3 & 80.9\,{\scriptsize(82.7)} & 0.013\RCnext
3 & LlamaParse (Cost Effective) & 84.6 & 85.5 & 91.9 & 87.2 & 73.1\,{\scriptsize(87.2)} & \textbf{88.6} & 81.2\,{\scriptsize(82.5)} & 0.004\RCnext
4 & Reducto & 83.9 & 81.9 & 82.8 & 84.9 & 89.1 & 84.1 & 80.9 & 0.030\RCnext
5 & Extend (Parse 2) & 83.2 & 80.2 & 83.3 & 84.2 & 85.1 & 84.8 & 81.5 & 0.025\RCnext
6 & Pulse (Ultra 2) & 82.7 & 86.1 & 88.7 & 84.7 & 86.8 & 73.5 & 76.4 & 0.015\RCnext
7 & Extend (Parse Light) & 81.6 & 78.3 & 78.7 & 81.4 & 85.4 & 84.7 & 81.2 & 0.006\RCnext
8 & Datalab (Accurate) & 81.3 & \textbf{91.1} & 92.5 & 90.3 & 67.8\,{\scriptsize(61.8)} & 70.1 & 75.8 & 0.010\RCnext
9 & Datalab (Balanced) & 81.1 & 90.3 & 92.6 & 89.5 & 67.6 & 70.3 & 76.0 & 0.004\RCnext
10 & Datalab (Fast) & 78.3 & 87.2 & 92.0 & 84.1 & 61.5 & 69.4 & 75.6 & 0.004\RCnext
11 & Databricks AI Parse & 76.9 & 79.3 & 79.4 & 81.3 & 64.2 & 83.8 & 73.5 & 0.004\RCnext
12 & Reducto (r-1) & 76.8 & 87.8 & 92.3 & 89.3 & 25.2\,{\scriptsize(\incomplete{73.6})} & 87.7 & 78.2 & 0.010\RCnext
13 & Mistral OCR 4.1 (annotation) & 72.5 & 84.5 & 82.0 & 81.8 & 29.8 & 81.8 & 75.3 & 0.005\RCnext
14 & Reducto (r-1-mini) & 71.6 & 85.1 & 91.7 & 86.1 & 1.4 & 87.9 & 77.7 & 0.001\RCnext
15 & LandingAI & 70.8 & 86.2 & 79.9 & 84.7 & 38.8 & 58.4 & 76.9 & 0.030\RCnext
16 & Extend (Parse) & 69.4 & 80.0 & 78.8 & 88.6 & 2.6 & 84.7 & 81.7 & 0.025\RCnext
17 & Mistral OCR 4.1 & 67.8 & 85.1 & 82.0 & 81.8 & 1.2 & 81.8 & 75.2 & 0.004\RCnext
18 & Reducto (non-agentic) & 66.3 & 79.1 & 79.8 & 73.4 & 1.5 & 83.8 & 80.0 & 0.022\RCnext
19 & Azure Document Intelligence (layout) & 66.1 & 79.8 & 77.9 & 77.6 & 1.5 & 83.8 & 76.3\,{\scriptsize(75.4)} & 0.010\RClast

\bottomrule
\end{tabularx}
\caption{Document parsers ranked by overall score using default configurations. Bold marks the best score in each column. Parentheses show dimension scores with an optional chart or form feature enabled. \$/page is the average cost, using the pricing basis in \Cref{app:experimental-setup}. \incompletenote}
\label{tab:main-results-v2}
\end{table}

\begin{table}[!htbp]
\RCsetup\setlength{\tabcolsep}{3pt}%
\renewcommand{\RCsep}{\noalign{\vspace{1.4pt}{\color{LIPanelLine}\hrule height 0.4pt}\vspace{1.4pt}}}%
\begin{tabularx}{\linewidth}{r>{\raggedright\arraybackslash}Xcccccccc}
\toprule
\RChead{\#} & \RChead{Pipeline} & \RChead{\pbtwo{} score} & \RChead{Text} & \RChead{Format.} & \RChead{Tables} & \RChead{Charts} & \RChead{\shortstack{Visual\\grounding}} & \RChead{Forms} & \RChead{\$/page} \\
\midrule
\RCgroup{Direct vision models}{the raw model called with our prompt}
1 & Claude Opus 5.5 & 85.7 & 92.7 & 89.4 & 94.1 & 74.1 & 86.2 & 77.8 & 0.065\RCnext
2 & Gemini 4 argon (low) & 84.2 & 89.7 & 89.2 & 92.8 & 68.2 & 88.1 & 77.2 & 0.027\RCnext
3 & GPT-5.6 terra (low) & 77.1 & 81.0 & 87.9 & 79.6 & 76.3 & 62.5 & 75.1 & 0.031\RCnext
4 & Gemini 3 Flash (minimal) & 76.6 & 86.7 & 86.1 & 88.9 & 58.0 & 66.3 & 73.8 & 0.009\RCnext
5 & Gemini 3.7 Flash (low) & 75.6 & 86.7 & 86.6 & 89.4 & 33.9 & 80.9 & 75.9 & 0.008\RCnext
6 & Gemini 3.8 Flash (low) & 75.5 & 87.1 & 86.7 & 88.8 & 33.3 & 81.3 & 76.0 & 0.008\RCnext
7 & Gemini 3.5 Flash & 74.7 & 88.5 & 86.2 & 90.0 & 31.1 & 75.3 & 77.4 & 0.050\RCnext
8 & Gemini 3.6 Flash & 74.7 & 88.2 & 85.8 & 89.8 & 30.4 & 76.0 & 77.8 & 0.042\RCnext
9 & Muse Ember 1.0 (low) & 74.5 & 90.2 & 86.0 & 89.0 & 60.5 & 43.8 & 77.5 & 0.002\RCnext
10 & GPT-6.1 terra (low) & 74.3 & 87.4 & 88.4 & 92.2 & 46.7 & 51.9 & 79.0 & 0.012\RCnext
11 & Claude Sonnet 5.5 & 73.8 & 91.4 & 86.3 & 90.9 & 38.6 & 57.9 & 77.6 & 0.031\RCnext
12 & GPT-6 luna (low) & 73.5 & 84.7 & 89.2 & 81.5 & 47.9 & 61.7 & 75.8 & 0.001\RCnext
13 & Claude Haiku 5.5 & 70.1 & 89.2 & 87.4 & 83.4 & 12.1 & 73.0 & 75.4 & 0.002\RCnext
14 & GPT-5.6 luna (low) & 69.1 & 80.7 & 88.6 & 76.7 & 40.7 & 54.0 & 74.0 & 0.003\RCnext
15 & Gemini 3.1 Flash-Lite & 62.7 & 84.0 & 79.3 & 79.7 & 12.2 & 48.1 & 73.0 & 0.004\RCnext
16 & Gemini 3.5 Flash-Lite & 61.4 & 79.2 & 78.7 & 72.2 & 5.1 & 62.2 & 71.3 & 0.007\RClast
\RCgroup{Open-weight models}{self-hosted, each with its own prompt or pipeline}
1 & Infinity-Parser2 Pro & 77.1 & 85.6 & 78.8 & 86.2 & 55.6 & 82.4 & 74.0 & --\RCnext
2 & LightOnOCR-3 4B & 77.0 & 84.1 & 81.6 & 78.2 & 62.9 & 80.5 & 75.0 & --\RCnext
3 & Chandra 2 & \incomplete{76.2} & 77.2 & 88.8 & 85.5 & 62.5 & 69.6 & \incomplete{73.7} & --\RCnext
4 & LightOnOCR-3 0.8B & 75.9 & 81.0 & 80.9 & 78.6 & 61.2 & 79.2 & 74.4 & --\RCnext
5 & MinerU2.5-Pro & \incomplete{75.6} & 76.6 & 79.0 & \incomplete{80.6} & 64.0 & 82.2 & 71.1 & --\RCnext
6 & Infinity-Parser2 Flash & 74.4 & 80.8 & 78.3 & 75.4 & 56.6 & 81.4 & 73.8 & --\RCnext
7 & Nemotron Parse 3.5 & 73.5 & 81.2 & 82.1 & 69.5 & 60.1 & 76.4 & 71.4 & --\RCnext
8 & PaddleOCR-VL 1.6 & 70.3 & 73.5 & 79.4 & 73.0 & 57.5 & 70.1 & 68.1 & --\RCnext
9 & Qwen3.8 27B & 61.6 & 81.6 & 77.3 & 36.7 & 41.3 & 63.5 & 69.0 & --\RClast

\bottomrule
\end{tabularx}
\caption{Frontier VLMs and open-weight models, ranked within each group by overall score. Columns follow \Cref{tab:main-results-v2}. Reasoning settings appear in parentheses. A dash in the cost column indicates that serving cost was not measured.}
\label{tab:main-results-models}
\end{table}

\clearpage
\begingroup
\RClongsetup\scriptsize\setlength{\tabcolsep}{1.8pt}
\begin{longtable}{@{}>{\raggedright\arraybackslash}p{4.6cm}cccccccccccccc@{}}
\caption{Mean page scores for the 13 collections used in the overall score, all at v2.0. The first five are core collections: Text, Tables, Charts, Layout and Forms. The remaining eight are targeted collections listed in \Cref{tab:app-dataset-composition}. Strikethrough and underline share one weighted group. $n$ counts evaluated collections. \incompletenote}\label{tab:app-datasets}\label{tab:dataset-results}\\
\toprule
\RChead{Pipeline} & \RChead{Text} & \RChead{Tab.} & \RChead{Chart} & \RChead{Lay.} & \RChead{Form} & \RChead{Hdr.} & \RChead{SERFF} & \RChead{Fin.} & \RChead{Dense} & \RChead{Head.} & \RChead{Strike} & \RChead{Under.} & \RChead{Blank} & \RChead{$n$} \\
\midrule
\endfirsthead
\toprule
\RChead{Pipeline} & \RChead{Text} & \RChead{Tab.} & \RChead{Chart} & \RChead{Lay.} & \RChead{Form} & \RChead{Hdr.} & \RChead{SERFF} & \RChead{Fin.} & \RChead{Dense} & \RChead{Head.} & \RChead{Strike} & \RChead{Under.} & \RChead{Blank} & \RChead{$n$} \\
\midrule
\endhead
\bottomrule
\endfoot
\multicolumn{15}{l}{\rule{0pt}{2.6ex}\sffamily\bfseries\scriptsize Document parsers, ranked}\\
LlamaParse (Agentic Plus) & 90.3 & 96.1 & 92.6 & 87.5 & 79.8 & 90.3 & 93.5 & 93.7 & 94.8 & 97.1 & 94.5 & 92.4 & 95.7 & 13\\
LlamaParse (Agentic) & 87.0 & 94.9 & 77.4 & 88.3 & 78.9 & 75.2 & 85.3 & 92.2 & 86.5 & 97.2 & 82.7 & 94.1 & 95.5 & 13\\
LlamaParse (Cost Effective) & 85.5 & 91.2 & 73.1 & 88.6 & 79.2 & 73.3 & 84.1 & 90.8 & 84.5 & 97.6 & 78.1 & 91.7 & 95.0 & 13\\
Reducto & 81.9 & 88.6 & 89.1 & 84.1 & 79.0 & 70.2 & 86.4 & 88.5 & 80.2 & 95.3 & 32.6 & 49.5 & 94.0 & 13\\
Extend (Parse 2) & 80.2 & 87.3 & 85.1 & 84.8 & 79.6 & 74.3 & 84.8 & 88.8 & 76.5 & 95.8 & 85.5 & 34.5 & 94.8 & 13\\
Pulse (Ultra 2) & 86.1 & 90.7 & 86.8 & 73.5 & 74.3 & 65.7 & 85.3 & 92.4 & 71.5 & 96.4 & 67.9 & 74.5 & 91.8 & 13\\
Extend (Parse Light) & 78.3 & 85.5 & 85.4 & 84.7 & 79.2 & 70.3 & 82.9 & 86.1 & 69.9 & 96.0 & 13.9 & 20.0 & 95.2 & 13\\
Datalab (Accurate) & 91.1 & 93.3 & 67.8 & 70.1 & 73.2 & 82.9 & 86.6 & 93.2 & 86.2 & 96.6 & 89.4 & 93.0 & 93.9 & 13\\
Datalab (Balanced) & 90.3 & 92.7 & 67.6 & 70.3 & 73.5 & 82.2 & 86.0 & 92.7 & 84.8 & 96.6 & 91.0 & 92.8 & 93.9 & 13\\
Datalab (Fast) & 87.2 & 86.9 & 61.5 & 69.4 & 73.0 & 79.1 & 84.2 & 91.3 & 70.5 & 96.5 & 87.6 & 91.3 & 93.8 & 13\\
Databricks AI Parse & 79.3 & 84.2 & 64.2 & 83.8 & 70.9 & 66.1 & 81.2 & 90.0 & 76.4 & 94.1 & 22.4 & 34.2 & 91.6 & 13\\
Reducto (r-1) & 87.8 & 92.9 & 25.2 & 87.7 & 76.1 & 75.3 & 88.8 & 92.3 & 86.8 & 95.9 & 95.5 & 94.6 & 93.0 & 13\\
Mistral OCR 4.1 (annotation) & 84.5 & 86.4 & 29.8 & 81.8 & 72.9 & 77.2 & 79.0 & 88.4 & 64.2 & 96.4 & 25.4 & 20.0 & 92.2 & 13\\
Reducto (r-1-mini) & 85.1 & 90.2 & 1.4 & 87.9 & 75.6 & 68.6 & 85.6 & 90.8 & 82.9 & 95.6 & 94.1 & 96.8 & 92.5 & 13\\
LandingAI & 86.2 & 89.3 & 38.8 & 58.4 & 74.8 & 69.3 & 80.7 & 90.6 & 80.0 & 90.9 & 32.1 & 21.4 & 92.2 & 13\\
Extend (Parse) & 80.0 & 91.4 & 2.6 & 84.7 & 79.8 & 77.2 & 87.0 & 92.9 & 85.7 & 96.0 & 13.9 & 20.0 & 95.1 & 13\\
Mistral OCR 4.1 & 85.1 & 86.3 & 1.2 & 81.8 & 72.8 & 77.1 & 79.4 & 88.4 & 64.0 & 96.5 & 25.3 & 20.0 & 92.2 & 13\\
Reducto (non-agentic) & 79.1 & 80.5 & 1.5 & 83.8 & 78.1 & 38.1 & 78.6 & 81.9 & 67.1 & 95.3 & 15.7 & 22.8 & 93.7 & 13\\
Azure Document Intelligence (layout) & 79.8 & 86.2 & 1.5 & 83.8 & 73.7 & 34.5 & 77.7 & 86.9 & 76.7 & 96.1 & 13.9 & 20.0 & 94.0 & 13\\
\multicolumn{15}{l}{\rule{0pt}{2.6ex}\sffamily\bfseries\scriptsize Direct vision models}\\
Claude Opus 5.5 & 92.7 & 96.2 & 74.1 & 86.2 & 75.7 & 84.5 & 93.9 & 96.3 & 93.4 & 96.0 & 93.7 & 36.0 & 93.2 & 13\\
Gemini 4 argon (low) & 89.7 & 94.3 & 68.2 & 88.1 & 74.9 & 87.4 & 90.9 & 94.2 & 92.7 & 95.3 & 89.2 & 52.0 & 93.3 & 13\\
GPT-5.6 terra (low) & 81.0 & 83.7 & 76.3 & 62.5 & 72.7 & 57.6 & 83.1 & 89.1 & 71.8 & 95.4 & 70.9 & 61.1 & 92.1 & 13\\
Gemini 3 Flash (minimal) & 86.7 & 91.8 & 58.0 & 66.3 & 72.1 & 78.3 & 87.7 & 92.2 & 86.1 & 95.8 & 67.1 & 43.3 & 85.1 & 13\\
Gemini 3.7 Flash (low) & 86.7 & 92.6 & 33.9 & 80.9 & 73.5 & 75.2 & 89.4 & 92.2 & 88.2 & 96.1 & 74.1 & 49.2 & 93.1 & 13\\
Gemini 3.8 Flash (low) & 87.1 & 91.9 & 33.3 & 81.3 & 73.5 & 75.6 & 88.2 & 91.6 & 87.4 & 95.9 & 72.9 & 55.9 & 93.1 & 13\\
Gemini 3.5 Flash & 88.5 & 92.2 & 31.1 & 75.3 & 75.2 & 80.9 & 89.9 & 93.3 & 86.7 & 95.8 & 71.0 & 42.2 & 92.5 & 13\\
Gemini 3.6 Flash & 88.2 & 92.2 & 30.4 & 76.0 & 75.7 & 81.8 & 89.1 & 92.7 & 86.3 & 95.9 & 69.4 & 39.7 & 93.1 & 13\\
Muse Ember 1.0 (low) & 90.2 & 92.6 & 60.5 & 43.8 & 75.2 & 74.2 & 87.4 & 92.7 & 87.1 & 95.9 & 43.0 & 43.5 & 93.3 & 13\\
GPT-6.1 terra (low) & 87.4 & 96.5 & 46.7 & 51.9 & 77.7 & 72.0 & 93.8 & 95.4 & 90.8 & 96.3 & 87.1 & 42.7 & 88.7 & 13\\
Claude Sonnet 5.5 & 91.4 & 93.6 & 38.6 & 57.9 & 75.5 & 82.4 & 89.6 & 94.0 & 86.7 & 95.8 & 61.0 & 30.6 & 92.2 & 13\\
GPT-6 luna (low) & 84.7 & 87.1 & 47.9 & 61.7 & 73.6 & 63.6 & 80.4 & 90.4 & 69.5 & 95.2 & 83.4 & 62.2 & 91.4 & 13\\
Claude Haiku 5.5 & 89.2 & 88.0 & 12.1 & 73.0 & 73.2 & 67.0 & 85.4 & 89.1 & 73.8 & 95.3 & 69.6 & 45.5 & 90.7 & 13\\
GPT-5.6 luna (low) & 80.7 & 83.2 & 40.7 & 54.0 & 71.8 & 52.4 & 79.6 & 87.8 & 61.3 & 96.1 & 77.4 & 64.2 & 89.7 & 13\\
Gemini 3.1 Flash-Lite & 84.0 & 84.2 & 12.2 & 48.1 & 70.7 & 61.2 & 81.2 & 84.5 & 74.0 & 95.3 & 16.5 & 20.0 & 89.1 & 13\\
Gemini 3.5 Flash-Lite & 79.2 & 78.3 & 5.1 & 62.2 & 68.8 & 58.1 & 69.5 & 68.8 & 67.9 & 93.1 & 27.3 & 22.2 & 88.8 & 13\\
\multicolumn{15}{l}{\rule{0pt}{2.6ex}\sffamily\bfseries\scriptsize Open-weight models}\\
Infinity-Parser2 Pro & 85.6 & 88.9 & 55.6 & 82.4 & 71.4 & 74.1 & 86.7 & 89.7 & 83.4 & 94.4 & 13.0 & 20.7 & 91.8 & 13\\
LightOnOCR-3 4B & 84.1 & 83.9 & 62.9 & 80.5 & 72.8 & 65.5 & 81.4 & 80.7 & 62.2 & 96.7 & 14.4 & 20.8 & 90.2 & 13\\
Chandra 2 & 77.2 & 89.1 & 62.5 & 69.6 & 72.2 & 72.4 & 85.4 & 91.7 & 78.0 & 95.9 & 57.1 & 78.2 & \incomplete{83.7} & 13\\
LightOnOCR-3 0.8B & 81.0 & 84.3 & 61.2 & 79.2 & 72.3 & 61.4 & 80.5 & 82.4 & 67.0 & 96.7 & 13.8 & 20.8 & 89.4 & 13\\
MinerU2.5-Pro & 76.6 & 86.0 & 64.0 & 82.2 & 68.3 & \incomplete{52.0} & 82.6 & 90.0 & 76.1 & 94.6 & 13.6 & 20.0 & 90.7 & 13\\
Infinity-Parser2 Flash & 80.8 & 79.7 & 56.6 & 81.4 & 71.1 & 57.7 & 74.4 & 82.6 & 70.2 & 94.9 & 13.9 & 20.6 & 92.3 & 13\\
Nemotron Parse 3.5 & 81.2 & 75.8 & 60.1 & 76.4 & 69.3 & 55.3 & 67.4 & 77.2 & 52.8 & 95.8 & 16.9 & 26.7 & 86.4 & 13\\
PaddleOCR-VL 1.6 & 73.5 & 76.2 & 57.5 & 70.1 & 65.7 & 69.7 & 71.7 & 80.0 & 58.3 & 93.3 & 13.8 & 36.1 & 85.5 & 13\\
Qwen3.8 27B & 81.6 & 38.6 & 41.3 & 63.5 & 66.7 & 38.1 & 35.7 & 37.1 & 28.3 & 93.2 & 13.9 & 18.4 & 85.2 & 13\\
\multicolumn{15}{l}{\rule{0pt}{2.6ex}\sffamily\bfseries\scriptsize Option pipelines (base configuration plus one option)}\\
LlamaParse (Agentic) + specialized charts & -- & -- & 87.1 & -- & -- & -- & -- & -- & -- & -- & -- & -- & -- & 1\\
LlamaParse (Cost Effective) + specialized charts & -- & -- & 87.2 & -- & -- & -- & -- & -- & -- & -- & -- & -- & -- & 1\\
Reducto (r-1) + chart agent & -- & -- & \incomplete{73.6} & -- & -- & -- & -- & -- & -- & -- & -- & -- & -- & 1\\
Datalab (Accurate) + chart understanding & -- & -- & 61.8 & -- & -- & -- & -- & -- & -- & -- & -- & -- & -- & 1\\
LlamaParse (Agentic Plus) + enriched forms & -- & -- & -- & -- & 81.2 & -- & -- & -- & -- & -- & -- & -- & 95.0 & 2\\
LlamaParse (Agentic) + enriched forms & -- & -- & -- & -- & 81.0 & -- & -- & -- & -- & -- & -- & -- & 94.8 & 2\\
LlamaParse (Cost Effective) + enriched forms & -- & -- & -- & -- & 80.7 & -- & -- & -- & -- & -- & -- & -- & 94.7 & 2\\
Azure Document Intelligence (layout) + key--value pairs & -- & -- & -- & -- & 72.9 & -- & -- & -- & -- & -- & -- & -- & 93.0 & 2\\

\end{longtable}
\endgroup

\subsubsection{Opt-in chart and form features}
\label{sec:options-results}

Several parsers offer optional chart or form features.
\Cref{tab:options} compares each feature with its base configuration on the dimension it targets.
The default overall scores exclude these features.

\paragraph{Chart extraction.}
Enabling a chart feature can substantially improve extraction.
Reducto r-1's chart agent raises its Charts score from \optReductoChartsBase{} to \optReductoChartsOpt{}.
LlamaParse's specialized chart parsing raises the score from \optAgenticChartsBase{} to \optAgenticChartsOpt{} on Agentic and from \optCeChartsBase{} to \optCeChartsOpt{} on Cost Effective.
The gains are not universal.
Datalab Accurate's chart understanding lowers its score from \optDatalabChartsBase{} to \optDatalabChartsOpt{}.

\paragraph{Cost and latency.}
Reducto's chart agent raises cost from \$\optReductoChartsCostBase{} to \$\optReductoChartsCostOpt{} per page.
It also increases median latency from \optReductoChartsLatBase\,s to \optReductoChartsLatOpt\,s per chart page.
LlamaParse's chart parsing keeps the same per-page price but takes about three to six times longer.

\paragraph{Form extraction.}
Form features make smaller differences.
The default scores of ranked parsers range from \resMinParserForms{} to \resMaxParserForms{}.
LlamaParse's enriched forms add one to three points, including a change from \optAgenticFormsBase{} to \optAgenticFormsOpt{} on Agentic.
Azure's key--value feature lowers its Forms score from \optAzureFormsBase{} to \optAzureFormsOpt{}.

\paragraph{Estimated overall effect.}
Replacing only the Charts dimension score would raise LlamaParse Cost Effective's overall score from \optCeChartsVtwoBase{} to \optCeChartsVtwoOpt{}.
The same calculation raises Reducto r-1 from \optReductoChartsVtwoBase{} to \optReductoChartsVtwoOpt{}.
These are estimates because we tested each feature only on its target collections.
Enabling it on every page could also affect other dimensions and the total cost.

\begin{table}[!htbp]
\RCsetup\scriptsize\setlength{\tabcolsep}{2pt}
\begin{tabularx}{\linewidth}{>{\raggedright\arraybackslash}Xllccccccccc}
\toprule
& & & \multicolumn{3}{c}{\sffamily\bfseries Dimension} & & \multicolumn{2}{c}{\sffamily\bfseries \$/page} & \multicolumn{2}{c}{\sffamily\bfseries \pbtwo{} score} & \\
\cmidrule(lr){4-6}\cmidrule(lr){8-9}\cmidrule(lr){10-11}
\RChead{Base configuration} & \RChead{Feature} & \RChead{Dim.} & \RChead{Base} & \RChead{Feature} & \RChead{$\Delta$} & \RChead{Core forms} & \RChead{Base} & \RChead{Feat.} & \RChead{Base} & \RChead{w/ feat.} & \RChead{p50 (s)} \\
\midrule
\RCstart
LlamaParse (Agentic) & specialized charts & Charts & 77.4 & 87.1 & $+$9.7 & -- & 0.013 & 0.013 & 86.0 & 87.7 & 6.5 $\to$ 21.5\RCnext
LlamaParse (Cost Effective) & specialized charts & Charts & 73.1 & 87.2 & $+$14.2 & -- & 0.004 & 0.004 & 84.6 & 87.0 & 3.6 $\to$ 20.6\RCnext
Reducto (r-1) & chart agent & Charts & 25.2 & \incomplete{73.6} & \incomplete{$+$48.4} & -- & 0.010 & 0.064 & 76.8 & \incomplete{84.8} & 3.0 $\to$ 238.0\RCnext
Datalab (Accurate) & chart understanding & Charts & 67.8 & 61.8 & $-$6.1 & -- & 0.010 & 0.013 & 81.3 & 80.3 & 7.1 $\to$ 13.7\RCnext
LlamaParse (Agentic Plus) & enriched forms & Forms & 81.8 & 82.9 & $+$1.1 & 79.8 $\to$ 81.2 & 0.028 & 0.028 & 90.1 & 90.2 & 25.7 $\to$ 35.8\RCnext
LlamaParse (Agentic) & enriched forms & Forms & 80.9 & 82.7 & $+$1.8 & 78.9 $\to$ 81.0 & 0.013 & 0.013 & 86.0 & 86.3 & 6.5 $\to$ 25.7\RCnext
LlamaParse (Cost Effective) & enriched forms & Forms & 81.2 & 82.5 & $+$1.2 & 79.2 $\to$ 80.7 & 0.004 & 0.004 & 84.6 & 84.8 & 6.5 $\to$ 25.7\RCnext
Azure Document Intelligence (layout) & key--value pairs & Forms & 76.3 & 75.4 & $-$0.8 & 73.7 $\to$ 72.9 & 0.010 & 0.010 & 66.1 & 66.0 & 4.6 $\to$ 5.7\RClast

\bottomrule
\end{tabularx}
\caption{Effect of optional chart and form features. Base and Feature columns compare scores and costs, and $\Delta$ gives the dimension-score change. Core forms scores only the Forms core collection. Overall scores with a feature replace only its target dimension and are estimates. p50 gives median seconds per page before and after enabling the feature. \incompletenote}
\label{tab:options}
\end{table}


\begin{thebibliography}{74}
\providecommand{\natexlab}[1]{#1}
\providecommand{\url}[1]{\texttt{#1}}
\expandafter\ifx\csname urlstyle\endcsname\relax
  \providecommand{\doi}[1]{doi: #1}\else
  \providecommand{\doi}{doi: \begingroup \urlstyle{rm}\Url}\fi

\bibitem[Ahmed et~al.(2026)Ahmed, Auer, and D'Souza]{ahmed2026sciimageminer}
Fahad Ahmed, S{\"o}ren Auer, and Jennifer D'Souza.
\newblock {ICDAR 2026 Competition on Information Extraction from Atomic Layer Deposition/Etching (ALD/E) Scientific Figures}, 2026.
\newblock URL \url{https://arxiv.org/abs/2607.26848}.

\bibitem[{Allen Institute for AI}(2026)]{allenai2026olmocrbench}
{Allen Institute for AI}.
\newblock {olmOCR-bench}: Document parsing tests.
\newblock Software and documentation, 2026.
\newblock URL \url{https://github.com/allenai/olmocr/tree/main/olmocr/bench}.
\newblock Accessed: 2026-11-12.

\bibitem[Banerjee and Lavie(2005)]{banerjee2005meteor}
Satanjeev Banerjee and Alon Lavie.
\newblock {METEOR}: An automatic metric for {MT} evaluation with improved correlation with human judgments.
\newblock In \emph{Proceedings of the ACL Workshop on Intrinsic and Extrinsic Evaluation Measures for Machine Translation and/or Summarization}, pages 65--72. Association for Computational Linguistics, 2005.
\newblock URL \url{https://aclanthology.org/W05-0909/}.

\bibitem[Berkane and Majumder(2026)]{berkane2026epicurvebench}
Thomas Berkane and Maimuna~S. Majumder.
\newblock {EpiCurveBench}: Evaluating {VLMs} on epidemic curve digitization.
\newblock arXiv preprint arXiv:2605.27195, 2026.
\newblock URL \url{https://arxiv.org/abs/2605.27195}.

\bibitem[Bertillo et~al.(2026)Bertillo, Melchiorri, and Merialdo]{bertillo2026paperunplot}
Daniele Bertillo, Giorgio Melchiorri, and Paolo Merialdo.
\newblock {PaperUnPlot}: Benchmarking chart-to-table in the wild.
\newblock In \emph{Proceedings of the 4th International Workshop on Tabular Data Analysis (TaDA), VLDB Workshops}, 2026.
\newblock URL \url{https://tabular-data-analysis.github.io/tada2026/papers/TaDA26_13.pdf}.

\bibitem[Blecher et~al.(2024)Blecher, Cucurull, Scialom, and Stojnic]{blecher2023nougat}
Lukas Blecher, Guillem Cucurull, Thomas Scialom, and Robert Stojnic.
\newblock Nougat: Neural optical understanding for academic documents.
\newblock In \emph{International Conference on Learning Representations (ICLR)}, 2024.

\bibitem[{CHART-Infographics organizers}(2026)]{chartinfo2026metrics}
{CHART-Infographics organizers}.
\newblock {CHART-Infographics}: Task 6b evaluation code.
\newblock Software and documentation, 2026.
\newblock URL \url{https://github.com/chartinfo/chartinfo.github.io/blob/master/metrics/metric6b.py}.
\newblock Accessed: 2026-11-12.

\bibitem[Chen et~al.(2025)Chen, Liu, Shi, Chen, Zhao, and Ren]{chen2025mdeval}
Zhongpu Chen, Yinfeng Liu, Long Shi, Xingyan Chen, Yu~Zhao, and Fuji Ren.
\newblock {MDEval: Evaluating and Enhancing Markdown Awareness in Large Language Models}, 2025.
\newblock URL \url{https://arxiv.org/abs/2501.15000}.

\bibitem[Chi et~al.(2019)Chi, Huang, Xu, Yu, Yin, and Mao]{chi2019scitsr}
Zewen Chi, Heyan Huang, Heng-Da Xu, Houjin Yu, Wanxuan Yin, and Xian-Ling Mao.
\newblock {Complicated Table Structure Recognition}, 2019.
\newblock URL \url{https://arxiv.org/abs/1908.04729}.

\bibitem[Choi et~al.(2026)Choi, Lee, and Kim]{choi2026gap}
Hong-Jun Choi, Jongho Lee, and Jaeyoung Kim.
\newblock {Rethinking the Pointer Loss in Table Structure Recognition: Geometry-Aware Pointer Loss for Spatial Locality}, 2026.
\newblock URL \url{https://arxiv.org/abs/2606.18721}.

\bibitem[Clausner et~al.(2020)Clausner, Pletschacher, and Antonacopoulos]{clausner2020flexible}
Christian Clausner, Stefan Pletschacher, and Apostolos Antonacopoulos.
\newblock Flexible character accuracy measure for reading-order-independent evaluation.
\newblock \emph{Pattern Recognition Letters}, 131:\penalty0 390--397, 2020.
\newblock \doi{10.1016/j.patrec.2020.02.003}.
\newblock URL \url{https://www.primaresearch.org/publications/PRL_Clausner_FlexibleCharacterAccuracy}.

\bibitem[Cleveland and McGill(1984)]{cleveland1984graphical}
William~S. Cleveland and Robert McGill.
\newblock Graphical perception: Theory, experimentation, and application to the development of graphical methods.
\newblock \emph{Journal of the American Statistical Association}, 79\penalty0 (387):\penalty0 531--554, 1984.

\bibitem[Cliche et~al.(2017)Cliche, Rosenberg, Madeka, and Yee]{cliche2017scatteract}
Mathieu Cliche, David Rosenberg, Dhruv Madeka, and Connie Yee.
\newblock {Scatteract: Automated extraction of data from scatter plots}, 2017.
\newblock URL \url{https://arxiv.org/abs/1704.06687}.

\bibitem[{Datalab}(2026)]{marker2026benchmark}
{Datalab}.
\newblock {Marker}: Document conversion benchmark code.
\newblock Software and documentation, 2026.
\newblock URL \url{https://github.com/datalab-to/marker/tree/v1.6.0/benchmarks}.
\newblock Accessed: 2026-11-12.

\bibitem[{Docling project}(2026)]{docling2026eval}
{Docling project}.
\newblock {Docling} evaluation framework.
\newblock Software and documentation, 2026.
\newblock URL \url{https://github.com/docling-project/docling-eval}.
\newblock Accessed: 2026-11-12.

\bibitem[El~Maarouf et~al.(2021)El~Maarouf, Kang, Ait~Azzi, Bellato, Gan, and El-Haj]{maarouf2021fintoc}
Ismail El~Maarouf, Juyeon Kang, Abderrahim Ait~Azzi, Sandra Bellato, Mei Gan, and Mahmoud El-Haj.
\newblock The financial document structure extraction shared task ({FinTOC2021}).
\newblock In \emph{Proceedings of the 3rd Financial Narrative Processing Workshop}, pages 111--119. Association for Computational Linguistics, 2021.
\newblock URL \url{https://aclanthology.org/2021.fnp-1.21/}.

\bibitem[Freitag et~al.(2021)Freitag, Foster, Grangier, Ratnakar, Tan, and Macherey]{freitag2021experts}
Markus Freitag, George Foster, David Grangier, Viresh Ratnakar, Qijun Tan, and Wolfgang Macherey.
\newblock Experts, errors, and context: A large-scale study of human evaluation for machine translation.
\newblock \emph{Transactions of the Association for Computational Linguistics}, 9:\penalty0 1460--1474, 2021.

\bibitem[Fu et~al.(2025)Fu, Kuang, Song, Huang, Yang, Li, Zhu, Luo, Wang, Lu, Li, Tang, Shan, Lin, Liu, Wu, Feng, Liu, Huang, Tang, Chen, Jin, Liu, and Bai]{fu2025ocrbenchv2}
Ling Fu, Zhebin Kuang, Jiajun Song, Mingxin Huang, Biao Yang, Yuzhe Li, Linghao Zhu, Qidi Luo, Xinyu Wang, Hao Lu, Zhang Li, Guozhi Tang, Bin Shan, Chunhui Lin, Qi~Liu, Binghong Wu, Hao Feng, Hao Liu, Can Huang, Jingqun Tang, Wei Chen, Lianwen Jin, Yuliang Liu, and Xiang Bai.
\newblock {OCRBench} v2: An improved benchmark for evaluating large multimodal models on visual text localization and reasoning.
\newblock In \emph{Advances in Neural Information Processing Systems (NeurIPS), Datasets and Benchmarks Track}, 2025.

\bibitem[G{\"o}bel et~al.(2013)G{\"o}bel, Hassan, Oro, and Orsi]{gobel2013icdar}
Max G{\"o}bel, Tamir Hassan, Ermelinda Oro, and Giorgio Orsi.
\newblock {ICDAR} 2013 table competition.
\newblock In \emph{12th International Conference on Document Analysis and Recognition (ICDAR)}, pages 1449--1453, 2013.

\bibitem[He et~al.(2026)He, Ying, Cheng, Peng, Tian, Deng, and Wu]{he2026exchart}
Yuchen He, Peizhi Ying, Liqi Cheng, Kuilin Peng, Yuan Tian, Dazhen Deng, and Yingcai Wu.
\newblock Making multimodal {LLMs} reliable chart data extractors: A benchmark and training framework.
\newblock In \emph{Proceedings of the 2026 CHI Conference on Human Factors in Computing Systems}. Association for Computing Machinery, 2026.
\newblock \doi{10.1145/3772318.3790721}.
\newblock URL \url{https://doi.org/10.1145/3772318.3790721}.

\bibitem[Heakl et~al.(2025)Heakl, Sohail, Ranjan, Hossam, Ahmad, El-Geish, Maher, Shen, Khan, and Khan]{heakl2025kitab}
Ahmed Heakl, Abdullah Sohail, Mukul Ranjan, Rania Hossam, Ghazi~Shazan Ahmad, Mohamed El-Geish, Omar Maher, Zhiqiang Shen, Fahad Khan, and Salman Khan.
\newblock {KITAB-Bench: A Comprehensive Multi-Domain Benchmark for Arabic OCR and Document Understanding}, 2025.
\newblock URL \url{https://arxiv.org/abs/2502.14949}.

\bibitem[Heer and Bostock(2010)]{heer2010crowdsourcing}
Jeffrey Heer and Michael Bostock.
\newblock Crowdsourcing graphical perception: Using {Mechanical Turk} to assess visualization design.
\newblock In \emph{Proceedings of the SIGCHI Conference on Human Factors in Computing Systems (CHI)}, pages 203--212, 2010.

\bibitem[Horn and Keuper(2026)]{horn2026beyond}
Pius Horn and Janis Keuper.
\newblock Beyond string matching: Semantic evaluation of {PDF} table extraction.
\newblock arXiv preprint arXiv:2603.18652, 2026.

\bibitem[Jaume et~al.(2019)Jaume, Ekenel, and Thiran]{jaume2019funsd}
Guillaume Jaume, Hazim~Kemal Ekenel, and Jean-Philippe Thiran.
\newblock {FUNSD: A Dataset for Form Understanding in Noisy Scanned Documents}, 2019.
\newblock URL \url{https://arxiv.org/abs/1905.13538}.

\bibitem[{JiWER contributors}(2026)]{jiwer2026}
{JiWER contributors}.
\newblock {JiWER}: Word and character error rate evaluation.
\newblock Software and documentation, 2026.
\newblock URL \url{https://github.com/jitsi/jiwer}.
\newblock Accessed: 2026-11-12.

\bibitem[Kendall(1938)]{kendall1938tau}
Maurice~G. Kendall.
\newblock A new measure of rank correlation.
\newblock \emph{Biometrika}, 30\penalty0 (1/2):\penalty0 81--93, 1938.

\bibitem[Kuhn(1955)]{kuhn1955hungarian}
Harold~W. Kuhn.
\newblock The {Hungarian} method for the assignment problem.
\newblock \emph{Naval Research Logistics Quarterly}, 2\penalty0 (1--2):\penalty0 83--97, 1955.

\bibitem[Levenshtein(1966)]{levenshtein1966binary}
Vladimir~I. Levenshtein.
\newblock Binary codes capable of correcting deletions, insertions, and reversals.
\newblock \emph{Soviet Physics Doklady}, 10\penalty0 (8):\penalty0 707--710, 1966.

\bibitem[Li et~al.(2020)Li, Cui, Huang, Wei, Zhou, and Li]{li2020tablebank}
Minghao Li, Lei Cui, Shaohan Huang, Furu Wei, Ming Zhou, and Zhoujun Li.
\newblock {T}able{B}ank: Table benchmark for image-based table detection and recognition.
\newblock In \emph{Proceedings of the Twelfth Language Resources and Evaluation Conference}, pages 1918--1925. European Language Resources Association, 2020.
\newblock URL \url{https://aclanthology.org/2020.lrec-1.236/}.

\bibitem[Li et~al.(2025{\natexlab{a}})Li, Yepes, You, Pluci{\'n}ski, Operlejn, and Wolfe]{li2025score}
Renyu Li, Antonio~Jimeno Yepes, Yao You, Kamil Pluci{\'n}ski, Maximilian Operlejn, and Crag Wolfe.
\newblock {SCORE: A Semantic Evaluation Framework for Generative Document Parsing}, 2025{\natexlab{a}}.
\newblock URL \url{https://arxiv.org/abs/2509.19345}.

\bibitem[Li et~al.(2025{\natexlab{b}})Li, Zhou, Luo, Xiao, and Xu]{li2025chartanchor}
Xinhang Li, Jingbo Zhou, Pengfei Luo, Yixiong Xiao, and Tong Xu.
\newblock {ChartAnchor: Chart Grounding with Structural-Semantic Fidelity}, 2025{\natexlab{b}}.
\newblock URL \url{https://arxiv.org/abs/2512.01017}.

\bibitem[Li et~al.(2025{\natexlab{c}})Li, Abulaiti, Lu, Chen, Zheng, Lin, Han, Jiang, Dong, and Sun]{li2024readoc}
Zichao Li, Aizier Abulaiti, Yaojie Lu, Xuanang Chen, Jia Zheng, Hongyu Lin, Xianpei Han, Shanshan Jiang, Bin Dong, and Le~Sun.
\newblock {READoc}: A unified benchmark for realistic document structured extraction.
\newblock In \emph{Findings of the Association for Computational Linguistics: ACL 2025}, pages 21889--21905, 2025{\natexlab{c}}.

\bibitem[Lin et~al.(2014)Lin, Maire, Belongie, Hays, Perona, Ramanan, Doll{\'a}r, and Zitnick]{lin2014coco}
Tsung-Yi Lin, Michael Maire, Serge Belongie, James Hays, Pietro Perona, Deva Ramanan, Piotr Doll{\'a}r, and C.~Lawrence Zitnick.
\newblock Microsoft {COCO}: Common objects in context.
\newblock In \emph{European Conference on Computer Vision (ECCV)}, pages 740--755. Springer, 2014.

\bibitem[Liu et~al.(2023)Liu, Eisenschlos, Piccinno, Krichene, Pang, Lee, Joshi, Chen, Collier, and Altun]{liu2023deplot}
Fangyu Liu, Julian~Martin Eisenschlos, Francesco Piccinno, Syrine Krichene, Chenxi Pang, Kenton Lee, Mandar Joshi, Wenhu Chen, Nigel Collier, and Yasemin Altun.
\newblock {DePlot}: One-shot visual language reasoning by plot-to-table translation.
\newblock In \emph{Findings of the Association for Computational Linguistics: ACL 2023}, pages 10381--10399, 2023.

\bibitem[Luo et~al.(2021)Luo, Li, Wang, and Lin]{luo2021chartocr}
Junyu Luo, Zekun Li, Jinpeng Wang, and Chin-Yew Lin.
\newblock {ChartOCR}: Data extraction from charts images via a deep hybrid framework.
\newblock In \emph{Proceedings of the IEEE/CVF Winter Conference on Applications of Computer Vision (WACV)}, pages 1917--1925, 2021.
\newblock URL \url{https://openaccess.thecvf.com/content/WACV2021/html/Luo_ChartOCR_Data_Extraction_From_Charts_Images_via_a_Deep_Hybrid_WACV_2021_paper.html}.

\bibitem[Masry et~al.(2022)Masry, Long, Tan, Joty, and Hoque]{masry2022chartqa}
Ahmed Masry, Do~Xuan Long, Jia~Qing Tan, Shafiq Joty, and Enamul Hoque.
\newblock {ChartQA}: A benchmark for question answering about charts with visual and logical reasoning.
\newblock In \emph{Findings of the Association for Computational Linguistics: ACL 2022}, pages 2263--2279. Association for Computational Linguistics, 2022.

\bibitem[Mathew et~al.(2021)Mathew, Karatzas, and Jawahar]{mathew2021docvqa}
Minesh Mathew, Dimosthenis Karatzas, and C.~V. Jawahar.
\newblock {DocVQA}: A dataset for {VQA} on document images.
\newblock In \emph{Proceedings of the IEEE/CVF Winter Conference on Applications of Computer Vision (WACV)}, pages 2200--2209, 2021.

\bibitem[{ocreval contributors}(2026)]{ocreval2026}
{ocreval contributors}.
\newblock {ocreval}: {ISRI} analytic tools for {OCR} evaluation.
\newblock Software and documentation, 2026.
\newblock URL \url{https://eddieantonio.ca/ocreval/}.
\newblock Accessed: 2026-11-12.

\bibitem[{OmniAI}(2026)]{omniai2026benchmark}
{OmniAI}.
\newblock {OmniAI OCR} benchmark.
\newblock Software and documentation, 2026.
\newblock URL \url{https://github.com/getomni-ai/benchmark}.
\newblock Accessed: 2026-11-12.

\bibitem[{OpenDataLab}(2026)]{omnidocbench2026code}
{OpenDataLab}.
\newblock {OmniDocBench}: Evaluation code and matching procedures.
\newblock Software and documentation, 2026.
\newblock URL \url{https://github.com/opendatalab/OmniDocBench}.
\newblock Accessed: 2026-11-12.

\bibitem[{OpenDataLoader project}(2026)]{opendataloader2026bench}
{OpenDataLoader project}.
\newblock {OpenDataLoader} benchmark.
\newblock Software and documentation, 2026.
\newblock URL \url{https://github.com/opendataloader-project/opendataloader-bench}.
\newblock Accessed: 2026-11-12.

\bibitem[Ouyang et~al.(2025)Ouyang, Qu, Zhou, Zhu, Zhang, Lin, Wang, Zhao, Jiang, Zhao, Shi, Wu, Chu, Liu, Li, Xu, Zhang, Shi, Tu, and He]{ouyang2024omnidocbench}
Linke Ouyang, Yuan Qu, Hongbin Zhou, Jiawei Zhu, Rui Zhang, Qunshu Lin, Bin Wang, Zhiyuan Zhao, Man Jiang, Xiaomeng Zhao, Jin Shi, Fan Wu, Pei Chu, Minghao Liu, Zhenxiang Li, Chao Xu, Bo~Zhang, Botian Shi, Zhongying Tu, and Conghui He.
\newblock {OmniDocBench}: Benchmarking diverse {PDF} document parsing with comprehensive annotations.
\newblock In \emph{Proceedings of the IEEE/CVF Conference on Computer Vision and Pattern Recognition (CVPR)}, pages 24838--24848, 2025.

\bibitem[Pandey et~al.(2026)Pandey, Manchkanti, Adain, Hadi, and Sekhar]{pandey2026pulsebenchtab}
Ritvik Pandey, Sid Manchkanti, Mohammed~Wazir Adain, Mohammed Hadi, and Dushyanth Sekhar.
\newblock {PulseBench-Tab}: A multilingual benchmark for table extraction with graph-based evaluation.
\newblock arXiv preprint arXiv:2606.07534, 2026.
\newblock URL \url{https://arxiv.org/abs/2606.07534}.

\bibitem[Papineni et~al.(2002)Papineni, Roukos, Ward, and Zhu]{papineni2002bleu}
Kishore Papineni, Salim Roukos, Todd Ward, and Wei-Jing Zhu.
\newblock {BLEU}: a method for automatic evaluation of machine translation.
\newblock In \emph{Proceedings of the 40th Annual Meeting of the Association for Computational Linguistics (ACL)}, pages 311--318, 2002.

\bibitem[Pfitzmann et~al.(2022)Pfitzmann, Auer, Dolfi, Nassar, and Staar]{pfitzmann2022doclaynet}
Birgit Pfitzmann, Christoph Auer, Michele Dolfi, Ahmed~S. Nassar, and Peter Staar.
\newblock {DocLayNet}: A large human-annotated dataset for document-layout segmentation.
\newblock In \emph{Proceedings of the 28th ACM SIGKDD Conference on Knowledge Discovery and Data Mining}, pages 3743--3751. ACM, August 2022.

\bibitem[Popovi{\'c}(2015)]{popovic2015chrf}
Maja Popovi{\'c}.
\newblock chr{F}: character n-gram {F}-score for automatic {MT} evaluation.
\newblock In \emph{Proceedings of the Tenth Workshop on Statistical Machine Translation}, pages 392--395. Association for Computational Linguistics, 2015.
\newblock \doi{10.18653/v1/W15-3049}.
\newblock URL \url{https://aclanthology.org/W15-3049/}.

\bibitem[Poznanski et~al.(2025)Poznanski, Rangapur, Borchardt, Dunkelberger, Huff, Lin, Wilhelm, Lo, and Soldaini]{poznanski2025olmocr}
Jake Poznanski, Aman Rangapur, Jon Borchardt, Jason Dunkelberger, Regan Huff, Daniel Lin, Christopher Wilhelm, Kyle Lo, and Luca Soldaini.
\newblock {olmOCR}: Unlocking trillions of tokens in {PDFs} with vision language models.
\newblock arXiv preprint arXiv:2502.18443, 2025.

\bibitem[{Reducto}(2026{\natexlab{a}})]{reducto2026pair}
{Reducto}.
\newblock {Pairwise AI Ranking (PAIR)}: Open parse evaluation.
\newblock \url{https://reducto.ai/blog/pairwise-ai-ranking-parse-evaluation}, 2026{\natexlab{a}}.
\newblock Blog post, November 9, 2026. Accessed: 2026-11-10.

\bibitem[{Reducto}(2026{\natexlab{b}})]{reducto2026tablebench}
{Reducto}.
\newblock {RD-TableBench}: An open benchmark for {PDF} table extraction.
\newblock Software and documentation, 2026{\natexlab{b}}.
\newblock URL \url{https://github.com/reductoai/rd-tablebench}.
\newblock Accessed: 2026-11-12.

\bibitem[Rei et~al.(2020)Rei, Stewart, Farinha, and Lavie]{rei2020comet}
Ricardo Rei, Craig Stewart, Ana~C. Farinha, and Alon Lavie.
\newblock {COMET}: A neural framework for {MT} evaluation.
\newblock In \emph{Proceedings of the 2020 Conference on Empirical Methods in Natural Language Processing (EMNLP)}, pages 2685--2702, 2020.

\bibitem[Ribeiro et~al.(2020)Ribeiro, Wu, Guestrin, and Singh]{ribeiro2020checklist}
Marco~Tulio Ribeiro, Tongshuang Wu, Carlos Guestrin, and Sameer Singh.
\newblock Beyond accuracy: Behavioral testing of {NLP} models with {CheckList}.
\newblock In \emph{Proceedings of the 58th Annual Meeting of the Association for Computational Linguistics (ACL)}, pages 4902--4912, 2020.

\bibitem[Sai et~al.(2021)Sai, Dixit, Sheth, Mohan, and Khapra]{sai2021perturbation}
Ananya~B. Sai, Tanay Dixit, Dev~Yashpal Sheth, Sreyas Mohan, and Mitesh~M. Khapra.
\newblock Perturbation {CheckLists} for evaluating {NLG} evaluation metrics.
\newblock In \emph{Proceedings of the 2021 Conference on Empirical Methods in Natural Language Processing (EMNLP)}, pages 7219--7234, 2021.

\bibitem[Sellam et~al.(2020)Sellam, Das, and Parikh]{sellam2020bleurt}
Thibault Sellam, Dipanjan Das, and Ankur Parikh.
\newblock {BLEURT}: Learning robust metrics for text generation.
\newblock In \emph{Proceedings of the 58th Annual Meeting of the Association for Computational Linguistics (ACL)}, pages 7881--7892, 2020.

\bibitem[Smock et~al.(2022)Smock, Pesala, and Abraham]{smock2022pubtables}
Brandon Smock, Rohith Pesala, and Robin Abraham.
\newblock {PubTables-1M}: Towards comprehensive table extraction from unstructured documents.
\newblock In \emph{Proceedings of the IEEE/CVF Conference on Computer Vision and Pattern Recognition (CVPR)}, pages 4634--4642, 2022.

\bibitem[Smock et~al.(2023)Smock, Pesala, and Abraham]{smock2023gritsgridtablesimilarity}
Brandon Smock, Rohith Pesala, and Robin Abraham.
\newblock {GriTS}: Grid table similarity metric for table structure recognition.
\newblock In \emph{Document Analysis and Recognition -- ICDAR 2023}, volume 14188 of \emph{Lecture Notes in Computer Science}, pages 535--549. Springer, 2023.

\bibitem[Song et~al.(2026)Song, Efat, and Tavanapong]{song2026fairchart}
Seok~Hwan Song, Azher~Ahmed Efat, and Wallapak Tavanapong.
\newblock {Assessing Y-Axis Influence: Bias in Multimodal Language Models on Chart-to-Table Translation}, 2026.
\newblock URL \url{https://arxiv.org/abs/2604.24987}.

\bibitem[Soric et~al.(2026)Soric, Gracianne, Manolescu, and Senellart]{soric2026benchmarking}
Marijan Soric, C{\'e}cile Gracianne, Ioana Manolescu, and Pierre Senellart.
\newblock Benchmarking table extraction from heterogeneous scientific {PDF} documents.
\newblock In \emph{Proceedings of the ACM SIGKDD International Conference on Knowledge Discovery and Data Mining}, 2026.
\newblock URL \url{https://pierre.senellart.com/publications/soric2026benchmarking.pdf}.

\bibitem[Turski et~al.(2025)Turski, Chili{\'n}ski, and Borchmann]{turski2025checkboxqa}
Micha{\l} Turski, Mateusz Chili{\'n}ski, and {\L}ukasz Borchmann.
\newblock {Unchecked and Overlooked: Addressing the Checkbox Blind Spot in Large Language Models with CheckboxQA}, 2025.
\newblock URL \url{https://arxiv.org/abs/2504.10419}.

\bibitem[{Unstructured Technologies}(2026)]{unstructured2026eval}
{Unstructured Technologies}.
\newblock Unstructured evaluation metrics.
\newblock Software and documentation, 2026.
\newblock URL \url{https://github.com/Unstructured-IO/unstructured/tree/main/unstructured/metrics}.
\newblock Accessed: 2026-11-12.

\bibitem[{Upstage}(2026)]{upstage2026dpbench}
{Upstage}.
\newblock {DP-Bench}: Document parsing benchmark.
\newblock Software and documentation, 2026.
\newblock URL \url{https://huggingface.co/datasets/upstage/dp-bench}.
\newblock Accessed: 2026-11-12.

\bibitem[Wang et~al.(2026)Wang, Zhang, Xu, Lu, Li, Zhang, Huang, Xiong, Wang, Lu, Huang, Yang, Li, Li, Xu, and Xu]{wang2026finixdoc}
Hang Wang, Jin Zhang, Guoliang Xu, Pengyue Lu, Yao Li, Zijiao Zhang, Tianyu Huang, Weiqi Xiong, Yulong Wang, Chuqiao Lu, Wenkang Huang, Kai Yang, Yadong Li, Hui Li, Xingzhong Xu, and Xiao Xu.
\newblock {FinixDoc: Rethinking Financial Document Parsing Beyond Saturated Benchmarks}, 2026.
\newblock URL \url{https://arxiv.org/abs/2608.22842}.

\bibitem[Wise(1993)]{wise1993gst}
Michael~J. Wise.
\newblock String similarity via greedy string tiling and running {Karp-Rabin} matching.
\newblock Technical report, Department of Computer Science, University of Sydney, 1993.

\bibitem[Wu and Jin(2026)]{wu2026chartdensity}
Xinhe Wu and Yadong Jin.
\newblock {ChartDensity-Bench: Benchmarking MLLMs for Numerical Data Reconstruction under Visual Density}, 2026.
\newblock URL \url{https://arxiv.org/abs/2609.38781}.

\bibitem[Wu et~al.(2022)Wu, Zhang, and Li]{wu2022texttotable}
Xueqing Wu, Jiacheng Zhang, and Hang Li.
\newblock Text-to-table: A new way of information extraction.
\newblock In \emph{Proceedings of the 60th Annual Meeting of the {A}ssociation for {C}omputational {L}inguistics (Volume 1: Long Papers)}, pages 2518--2533. Association for Computational Linguistics, 2022.
\newblock \doi{10.18653/v1/2022.acl-long.180}.
\newblock URL \url{https://aclanthology.org/2022.acl-long.180/}.

\bibitem[Xia et~al.(2023)Xia, Peng, Ye, Li, Yan, Ye, Shi, Qiao, Yan, and Zhang]{xia2023structchart}
Renqiu Xia, Haoyang Peng, Hancheng Ye, Mingsheng Li, Xiangchao Yan, Peng Ye, Botian Shi, Yu~Qiao, Junchi Yan, and Bo~Zhang.
\newblock {StructChart: On the Schema, Metric, and Augmentation for Visual Chart Understanding}, 2023.
\newblock URL \url{https://arxiv.org/abs/2309.11268}.

\bibitem[Xia et~al.(2024)Xia, Zhang, Ye, Yan, Liu, Zhou, Chen, Ye, Dou, Shi, Yan, and Qiao]{xia2024chartx}
Renqiu Xia, Bo~Zhang, Hancheng Ye, Xiangchao Yan, Qi~Liu, Hongbin Zhou, Zijun Chen, Peng Ye, Min Dou, Botian Shi, Junchi Yan, and Yu~Qiao.
\newblock {ChartX \& ChartVLM: A Versatile Benchmark and Foundation Model for Complicated Chart Reasoning}, 2024.
\newblock URL \url{https://arxiv.org/abs/2402.12185}.

\bibitem[Xu et~al.(2022)Xu, Lv, Cui, Wang, Lu, Florencio, Zhang, and Wei]{xu2022xfund}
Yiheng Xu, Tengchao Lv, Lei Cui, Guoxin Wang, Yijuan Lu, Dinei Florencio, Cha Zhang, and Furu Wei.
\newblock {XFUND}: A benchmark dataset for multilingual visually rich form understanding.
\newblock In \emph{Findings of the {A}ssociation for {C}omputational {L}inguistics: {ACL} 2022}, pages 3214--3224. Association for Computational Linguistics, 2022.
\newblock \doi{10.18653/v1/2022.findings-acl.253}.
\newblock URL \url{https://aclanthology.org/2022.findings-acl.253/}.

\bibitem[Yang et~al.(2025)Yang, Tang, Li, Wang, Wan, Zhong, Liu, Yang, Wang, Bai, Jin, and Lin]{yang2024ccocr}
Zhibo Yang, Jun Tang, Zhaohai Li, Pengfei Wang, Jianqiang Wan, Humen Zhong, Xuejing Liu, Mingkun Yang, Peng Wang, Shuai Bai, Lianwen Jin, and Junyang Lin.
\newblock {CC-OCR}: A comprehensive and challenging {OCR} benchmark for evaluating large multimodal models in literacy.
\newblock In \emph{Proceedings of the IEEE/CVF International Conference on Computer Vision (ICCV)}, pages 21744--21754, 2025.

\bibitem[Zhang et~al.(2026)Zhang, Acosta, Carlson, Bron, Doulcet, Ospina, and Suo]{zhang2026parsebench}
Boyang Zhang, Sebasti{\'a}n~G. Acosta, Preston Carlson, Sacha Bron, Pierre-Lo{\"i}c Doulcet, Daniel~B. Ospina, and Simon Suo.
\newblock {ParseBench}: A document parsing benchmark for {AI} agents.
\newblock arXiv preprint arXiv:2604.08538, 2026.

\bibitem[Zheng et~al.(2023)Zheng, Chiang, Sheng, Zhuang, Wu, Zhuang, Lin, Li, Li, Xing, Zhang, Gonzalez, and Stoica]{zheng2023judging}
Lianmin Zheng, Wei-Lin Chiang, Ying Sheng, Siyuan Zhuang, Zhanghao Wu, Yonghao Zhuang, Zi~Lin, Zhuohan Li, Dacheng Li, Eric~P. Xing, Hao Zhang, Joseph~E. Gonzalez, and Ion Stoica.
\newblock Judging {LLM}-as-a-judge with {MT-Bench} and {Chatbot Arena}.
\newblock In \emph{Advances in Neural Information Processing Systems (NeurIPS), Datasets and Benchmarks Track}, 2023.

\bibitem[Zheng et~al.(2024)Zheng, Feng, Si, She, Lin, Jiang, and Wang]{zheng2024mmtab}
Mingyu Zheng, Xinwei Feng, Qingyi Si, Qiaoqiao She, Zheng Lin, Wenbin Jiang, and Weiping Wang.
\newblock Multimodal table understanding.
\newblock In \emph{Proceedings of the 62nd Annual Meeting of the {A}ssociation for {C}omputational {L}inguistics (Volume 1: Long Papers)}, pages 9102--9124. Association for Computational Linguistics, 2024.
\newblock \doi{10.18653/v1/2024.acl-long.493}.
\newblock URL \url{https://aclanthology.org/2024.acl-long.493/}.

\bibitem[Zheng et~al.(2021)Zheng, Burdick, Popa, Zhong, and Wang]{zheng2020gte}
Xinyi Zheng, Douglas Burdick, Lucian Popa, Xu~Zhong, and Nancy Xin~Ru Wang.
\newblock Global table extractor ({GTE}): A framework for joint table identification and cell structure recognition using visual context.
\newblock In \emph{Proceedings of the IEEE/CVF Winter Conference on Applications of Computer Vision (WACV)}, pages 697--706, 2021.

\bibitem[Zhong et~al.(2019)Zhong, Tang, and Jimeno~Yepes]{zhong2019publaynet}
Xu~Zhong, Jianbin Tang, and Antonio Jimeno~Yepes.
\newblock {PubLayNet}: Largest dataset ever for document layout analysis.
\newblock In \emph{2019 International Conference on Document Analysis and Recognition (ICDAR)}, pages 1015--1022. IEEE, 2019.

\bibitem[Zhong et~al.(2020)Zhong, ShafieiBavani, and Jimeno~Yepes]{zhong2020imagebasedtablerecognitiondata}
Xu~Zhong, Elaheh ShafieiBavani, and Antonio Jimeno~Yepes.
\newblock Image-based table recognition: Data, model, and evaluation.
\newblock In \emph{European Conference on Computer Vision (ECCV)}, pages 564--580. Springer, 2020.

\end{thebibliography}
\end{document}